# Semantic Measures for the Comparison of Units of Language, Concepts or Instances from Text and Knowledge Representation Analysis

## A Comprehensive Survey and a Technical Introduction to Knowledge-based Measures Using Semantic Graph Analysis

*Sébastien Harispe,*
*Sylvie Ranwez,*
*Stefan Janaqi,*
*Jacky Montmain*

*LGI2P/EMA Research Center, Parc scientifique*
*G. Besse 30035 Nîmes Cedex 1, France*                    firstname.name@mines-ales.fr

**From the authors :** **Please do not cite or refer to this work** – it is an outdated, mostly non-rigorous, working draft that has been used to prepare the book *Semantic Similarity from Natural Language and Ontology Analysis*. This draft has been made publicly available in the aim of sharing a primer attempt to provide an extended version of a state of the art on this broad topic. It should not be considered anymore as a valid reference. **Please refer to the following book for a more rigorous reference:**

**Harispe Sebastien, Sylvie Ranwez, Stefan Janaqi, and Jacky Montmain. "Semantic similarity from natural language and ontology analysis." Synthesis Lectures on Human Language Technologies 8, no. 1 (2015): 1-254.**

**http://doi.org/10.2200/S00639ED1V01Y201504HLT027**

## Abstract

Semantic measures are widely used today to estimate the strength of the semantic relationship between elements of various types: units of language (e.g., words, sentences, documents), concepts or even instances semantically characterized (e.g., diseases, genes, geographical locations). Semantic measures play an important role to compare such elements according to semantic proxies: texts and knowledge representations, which support their meaning or describe their nature. Semantic measures are therefore essential for designing intelligent agents which will for example take advantage of semantic analysis to mimic human ability to compare abstract or concrete *objects*.

This paper proposes a comprehensive survey of the broad notion of semantic measure for the comparison of units of language, concepts or instances based on semantic proxy analyses. Semantic measures generalize the well-known notions of semantic similarity, semantic relatedness and semantic distance, which have been extensively studied by various communities over the last decades (e.g., Cognitive Sciences, Linguistics, and Artificial Intelligence to mention a few). Definitions, related contributions in cognitive sciences, practical applications, and the several approaches used for the definitions of semantic measures are presented. In addition, protocols and benchmarks for their evaluations, as well as, software solutions dedicated to them are introduced.

The general presentation of the large diversity of existing semantic measures we propose is further completed by a detailed survey of a specific type of measures relying on knowledge representation analysis. This technical presentation mainly focuses on measures based on graph analyses. These measures are of particular interest for numerous communities and have recently gained a lot of attention in research and application, by taking advantage of several types of graph-based knowledge representations to compare words, concepts, or instances.

We conclude this work by highlighting some of the challenges offered to the communities involved in the study of semantic measures.

# Table of Contents

# 1 Introduction

*Semantic measures* (SMs) are widely used today to estimate the strength of the semantic relationship between elements such as units of language, concepts or even semantically characterized instances, according to information formally or implicitly supporting their meaning or describing their nature. They are based on the analysis of *semantic proxies* from which semantic evidences can be extracted. These evidences are expected to directly or indirectly characterize the meaning/nature of the compared elements. The semantic likeness of terms or concepts is sometimes better understood as the probability of a *mental activation* of one term/concept when another term/concept is discussed. Notice that the notion of SM is not framed in the rigorous mathematical definition of *measure*. It should instead be understood as any theoretical tool or function which enables the comparison of elements according to semantic evidences. SMs are therefore used to estimate the degree of the semantic relatedness[i] of elements through a numerical value.

Two broad types of semantic proxies can be used to extract semantic evidences. The first type corresponds to unstructured or semi-structured texts[ii] (e.g., plain texts, dictionaries). These texts contain informal evidences of the semantic relationship(s) between units of language. Intuitively, the more two words are related semantically, the more frequently they will co-occur in texts. For instance, the word *coffee* is more likely to co-occur with the word *sugar* than with the word *cat*, and, since it's common to drink coffee with sugar, most will agree that the pair of words *coffee/sugar* is more semantically coherent than the pair of words *coffee/cat*. It is therefore possible to use simple assumptions regarding the distribution of words to estimate the strength of the semantic relationship between two words based on the assumption that words semantically related tend to co-occur.

The other type of semantic proxy from which semantic evidences can be extracted is more general. It encompasses a large range of computer-readable and understandable resources, from structured vocabularies to highly formal knowledge representations (KRs). Contrary to the first type of semantic proxy (i.e., texts), proxies of this type are structured and explicitly model knowledge about the elements they define. As an example, in a knowledge representation defining the concepts *Coffee* and *Sugar*, a specific relationship will explicitly define that *Coffee - can be drink with - Sugar*. SMs based on knowledge analysis rely on techniques used to take advantage of semantic graphs (e.g., thesaurus, taxonomies, lightweight ontologies), or even highly formal KRs such as ontologies based on (description) logic.

A large diversity of measures exist to estimate the similarity or the dissimilarity between specific data structures (e.g., vectors, matrices, graphs) and data types (e.g., numbers, strings, dates). The specificity of SMs relies in the fact that they are based on the analysis of semantic proxies to take into account the *semantics* in the definition of the function which will be used to drive the comparison of elements. As an example, the measures used to compare two words according to their sequences of characters cannot be considered as SMs – only the characters of the words and their ordering is taken into account, not their meaning. Therefore, according to such measures, the two words *foal* and *horse* will be regarded as unrelated words.

From gene analysis to recommendation systems, SMs have recently found a broad field of applications and are today essential to leverage data mining, data analysis, classification, knowledge extraction, textual processing or even information retrieval based on text corpora or formal KRs. Due to their essential roles in numerous treatments requiring the meaning of compared elements (i.e., semantics) to be taken into ac-

---

[i] The broad notion of semantic relatedness will be rigorously introduced later, e.g., through the definition of semantic distance and similarity to cite a few.

[ii] Discourse is here considered as a text

count, the study of SMs has always been an interdisciplinary effort. Psychology, Cognitive Sciences, Linguists, Natural Language Processing, Semantic Web, and Biomedical informatics are among the most active communities which contribute to the study of SMs (2013). Due to this interdisciplinary nature of SMs, last decades have been very prolific in contributions related to the notion of semantic relatedness, semantic similarity or semantic distance, to mention a few. Before defining the technical terminology required to further introduce SMs, let's focus on their large diversity of applications.

## 1.1 Semantic Measures in Action

SMs are used to solve problems in a broad range of applications and domains. They enable to take advantage of the knowledge encompassed in unstructured/semi-structured texts corpora and KRs to compare *things*. They are therefore essential tools for the design of numerous algorithms and treatments in which semantics matters. Diverse practical applications which involve SMs are presented in this section. Three domains of applications are considered in particular: (i) Natural Language Processing, (ii) Knowledge Engineering/Semantic Web and Linked Data, and (iii) Biomedical informatics and Bioinformatics. Additional applications related to information retrieval and clustering are also briefly considered. The list of usages of SMs presented in this section is far from being exhaustive and only gives an overview of the large diversity of perspectives they open. Therefore, as a supplement to this list, an extensive classification of contributions related to SMs is proposed in appendix 1. This classification underlines the broad range of applications of SMs and highlights the large number of communities involved - it can thus be used to gain more insight on their usages in numerous contexts.

### 1.1.1 Natural Language Processing

Linguists have, quite naturally, been among the first to study SMs in the aim of comparing units of language (e.g., words, sentences, paragraphs, documents). The estimation of words/concepts relatedness plays an important role to detect paraphrase, e.g., duplicate content and plagiarism (Fernando & Stevenson 2008), to generate thesaurus or texts (Iordanskaja et al. 1991), to summarize texts (Kozima 1993), to identify discourse structure, and to design question answering systems (Bulskov et al. 2002; Freitas et al. 2011; C. Wang et al. 2012) to mention a few. The effectiveness of SMs to resolve both syntactic and semantic ambiguities have also been demonstrated multiple times, e.g., (Sussna 1993; Resnik 1999; Patwardhan et al. 2003).

Several surveys relative to usages of SMs and to the techniques used for their design for natural language processing can be found in (Curran 2004; S. M. Mohammad & Hirst 2012).

### 1.1.2 Knowledge Engineering, Semantic Web and Linked Data

Communities associated to Knowledge Engineering, Semantic Web and Linked Data play an import role in the definition of methodologies and standards to formally express machine-understandable KRs. They extensively study the problematic associated to the expression of structured and controlled vocabularies, as well as ontologies, i.e., formal and explicit specification of a shared conceptualisation defining a set of concepts, their relationships and axioms to model a domain[i] (Gruber 1993). These models rely on structured KRs in which the semantics of the concepts (classes) and relationships (properties) are rigor-ously and formally defined in an unambiguous way. Such KRs are therefore proxies of choice to compare the concepts and the instances of the domain they model. As we will see, a taxonomy of concepts, which is the backbone of most if not all KR, is particularly useful to estimate the degree of similarity of two concepts.

---

[i] More about ontologies: e.g., (Gruber 1995; Guarino et al. 2009; Fernandez-Lopez & Corcho 2010)

SMs are essential to integrate heterogeneous KRs and more generally for data integration. They play an important role to find correspondences between ontologies (ontology alignment[i]), in which similar concepts defined in different ontologies must be found (Euzenat & Shvaiko 2007). SMs are also used for the task of instance matching, in the aim of finding duplicate instances across data sources. Applications to provide inexact search capabilities based on KR analysis have also been proposed, e.g., (Hliaoutakis 2005; Varelas et al. 2005; Hliaoutakis et al. 2006; Kiefer et al. 2007; Sy et al. 2012; Pirró 2012). SMs have also been successfully applied for learning tasks using Semantic Web technologies (D'Amato 2007). Their benefits to take advantage of the Linked Data paradigm in the definition of recommendation systems have also been stressed in (Passant 2010; Harispe, Ranwez, et al. 2013a).

### 1.1.3 Biomedical Informatics & Bioinformatics

A large number of SMs have been defined for biomedical or bioinformatics studies. Indeed, in these domains, SMs are commonly used to take advantage of biomedical ontologies to study various types of instances (genes, proteins, drugs, diseases, phenotypes) which have been semantically characterized through a KR, e.g., ontologies or controlled vocabularies[ii]. Several surveys relative to usages of SMs in the biomedical domain can be found; we orient the reader to (Pedersen et al. 2007; Pesquita, Faria, et al. 2009; Guzzi et al. 2012).

The Gene Ontology (GO) (Ashburner et al. 2000) is the example of choice to highlight the large success encountered by ontologies in biology[iii]. Indeed, the GO is extensively used to conceptually annotate gene products on the basis of experimental observations or automatic inferences. These annotations are used to formally characterize gene products regarding their molecular functions, the biological processes they are involved in or even their cellular location. Thus, using SMs, these annotations make possible the automatic comparisons of genes' products not on the basis of particular gene properties (e.g. sequence, structural similarity, gene expression) but rather on the analysis of biological aspects formalized by the GO. Therefore, genes can further be analysed by considering their representation in a multi-dimensional semantic space expressing our current understanding of particular aspects of biology. In such cases, conceptual annotations bridge the gap between global knowledge of biology (e.g., organisation of molecular functions or cellular component) and fine-grained understanding of specific instance (e.g., the specific role of a gene at molecular level). SMs enable to take advantage of this knowledge to analyse instances, here genes and, open interesting perspectives to infer new knowledge about them.

Various studies have highlighted the relevance of SMs for assessing the functional similarity of genes (Wang et al. 2007; Z. Du et al. 2009), building gene clusters (Sheehan et al. 2008), validating and studying protein-protein interactions (Xu et al. 2008), analysing gene expression (Xu et al. 2009), evaluating gene sets' coherence (Diaz-Diaz & Aguilar-Ruiz 2011) or recommending gene annotations (Couto et al. 2006), to mention a few. A survey dedicated to SMs applied to the GO can be found in (Guzzi et al. 2012).

---

[i] The reader interested to ontology alignment may also consider the related problematic of schema matching and mapping (Bellahsène et al. 2011). The classification of the elementary matching approaches proposed by (Euzenat & Shvaiko 2007) is also an interesting starting point for a broad overview of the large diversity of measures and approaches proposed for alignment tasks.

[ii] Biology and biomedicine are heavy users of ontologies and controlled vocabularies, e.g. BioPortal, a portal dedicated to ontologies related to biology and the biomedical domain, references hundreds of ontologies (Whetzel et al. 2011).

[iii] More than 11k citations between 2000 and 2013!

### 1.1.4 Other Applications

*1.1.4.1 Information Retrieval*

SMs are used to overcome limitations of information retrieval techniques based on plain lexicographic term matching, i.e., simple models consider that a document is relevant according to a query, only if the terms specified in the query are used in the document. SMs can be used to take into account the meaning of words by going over syntactic search, and can therefore be used to refine models, e.g., synonyms will not be considered as words totally different anymore. SMs have successfully been used in the design of ontology-based information retrieval systems and for query expansion, e.g., (Hliaoutakis 2005; Varelas et al. 2005; Hliaoutakis et al. 2006; Baziz et al. 2007; Saruladha, Aghila & Raj 2010b; Sy et al. 2012).

SMs based on KRs also open interesting perspectives for the field of information retrieval as they enable to analyse and to query non-textual resources, e.g. genes annotated by concepts (Sy et al. 2012).

*1.1.4.2 GeoInformatics*

*GeoInformatics* actively contributes to the study of SMs. In this domain, SMs have, for instance, been used to compute the similarity between locations according to semantic characterizations of their geographic features (Janowicz et al. 2011), e.g. estimating the semantic similarity of tags defined in the OpenStreetMap Semantic Network (Ballatore et al. 2012). Readers interested in the applications of SMs in this field may also refer to the various references proposed in Appendix 1, e.g. (Akoka et al. 2005; Rodríguez et al. 2005; Formica & Pourabbas 2008; Janowicz et al. 2008).

## 1.2 Organization of this Survey

This contribution proposes both a general introduction to SMs and a technical survey regarding a specific type of measures based on KR analysis. It is organized as follows:

Section 2 introduces general notions related to SMs. Several cognitive models defined to better understand human cognition regarding his appreciation of similarity are briefly presented. As we will see, these cognitive models play an essential role for the design of SMs and are critical to deeply understand technical aspects of the measures. Several mathematical notions related to the notions of distance and similarity are also introduced. They are needed to formally define SMs in mathematical terms by taking into consideration key mathematical contributions related to distance and similarity. In this section, the reader is also introduced to the commonly adopted terminology associated to SMs; notions of semantic similarity, dissimilarity, distance, relatedness, or even taxonomical distance will be defined.

Based on the introduction of the broad notion of SMs, section 3 presents a classification of the large diversity of strategies proposed for the definition of SMs. The proposed classification relies on the analysis of:

- The type of compared elements (units of language, concepts/classes, instances semantically characterized).
- The canonical form used to represent these elements.

- The semantic proxy which is used to extract the semantics associated to the compared elements, i.e.., corpora of texts, KRs.

According to the type of semantic proxy on which is based the comparison, three families of SMs are further distinguished:

- *Distributional measures* which mainly analyse corpora of texts.
- *Knowledge-based measures* which take advantage of structured knowledge to extract the semantics on which the SMs rely.
- *Hybrid measures* which take advantage of both, text corpora and KRs.

Section 4 is dedicated to the practical computation and evaluation of SMs. Several software solutions for the computation and the analysis of measures are presented. We also discuss the protocols and methodologies commonly used to assess the accuracy and the performance of measures in specific usage contexts.

Section 5 is dedicated to a technical and in-depth presentation of a specific type of SMs based on KR analysis. In this section we focus on SMs based on graph analysis, a highly popular approach used to compare structured terms, concepts, groups of concepts or even instances defined in KRs such as ontologies.

At the light of this study, section 0distinguishes some of the challenges faced by SMs designers and scientific communities contributing to the topic. A general conclusion ends this article.

# 2 General Notions and Definitions

SMs have been studied through various notions and not always in rigorous terms. Some definitions are even still subject to debate and not all communities agree on the semantics carried by the terminology they use. Thus, the literature related to the topic manipulates notions of semantic similarity, relatedness, distance, taxonomic distance and dissimilarity (I let your creativity speak); these notions deserve to be rigorously defined. This reflects the difficulty to limit the semantic similarity, as detected by humans, inside formal (and partial) logical mathematical models.

This section first introduces generalities related to the domain and a more precise definition of the notion of SM is proposed. The main models of similarity defined in cognitive sciences are next introduced. As we will see, they play an important role to understand the (diversity of) approaches adopted to design SMs. Several mathematical definitions and properties related to distance and similarity are next presented. These definitions will be used to distinguish mathematical properties of interest for the characterization and the study of SMs.

## 2.1 Semantic Measures: Generalities

### 2.1.1 Semantic Measures: Definition

Human cognitive system is sensitive to similarity, which explains that the capacity to estimate the similarity of things is essential in numerous treatments. It is indeed a key element to initiate the process of learning in which the capacity to recognize similar situations[i], for instance, helps us to build our experience, to activate mental traces, to make decisions, to innovate applying experience gained in previously solved problem to similar problems[ii] (Holyoak & Koh 1987; Ross 1987; Novick 1988; Ross 1989; Vosniadou & Ortony 1989; Gentner & Markman 1997). According to the theories of transfer, the process of learning is also subject to similarity since new skills are expected to be easier to learn if they are similar to skills already learned (Markman & Gentner 1993). Similarity is therefore a central component of memory retrieval, categorization, pattern recognition, problem solving, reasoning, as well as social judgment, e.g., refer to (Markman & Gentner 1993; Hahn et al. 2003; Goldstone & Son 2004) for associated references.

In this context, the goal of SMs is easy to understand – they aim to capture the strength of the semantic interaction between elements (e.g., words, concepts) regarding their meaning. Are the words *car* and *auto* more semantically related than the words *car* and *mountain*? Most people will agree to say yes. This has been proved in multiple experiments, inter-human agreement on semantic similarity ratings is high, e.g. (Rubenstein & Goodenough 1965; Miller & Charles 1991; Pakhomov et al. 2010)[iii].

Appreciation of similarity is obviously subject to multiple factors. Our personal background is an example of such a factor, e.g., elderly persons and teenagers will probability not associate the same score of semantic similarity between the two concepts *Phone* and *Computer*[iv]. However, most of the time, a con-

---

[i] Cognitive models based on categorization consider that human classify things, e.g., experience of life, according to their similarity to some prototype, abstraction or previous examples (Markman & Gentner 1993).

[ii] The similarity is here associated to the notion of generalization and is measured in terms of probability of inter-stimulus-confusion errors (Nosofsky 1992).

[iii] As an example, considering three benchmarks, (Schwartz & Gomez 2011) observed 73% to 89% human inter-agreement between scores of semantic similarity associated to pairs of words.

[iv] Smartphone are today kinds of computers and very different from the first communication device patented in 1876 by Bell.

sensus regarding the estimation of the strength of the semantic link between elements can be reached - this is what makes the notion of SMs intuitive and meaningful[i].

The majority of SMs try to mimic human capacity to assess the degree of *relatedness* between things according to semantic evidences. However, strictly speaking, SMs evaluate the strength of the semantic interactions between things according to the analysis of semantic proxies (texts, KRs), that's it. Therefore, not all measures aim at mimicking human appreciation of similarity. Indeed, in some cases, SMs' designers only aim to compare elements according to the information defined in a semantic proxy, no matter if the results produced by the measure correlate with human appreciation of semantic similarity/relatedness. This is, for instance, often the case in the design of SMs based on KRs. In these cases, the KR can be associated to our brain and the SM can be regarded as our capacity to take advantage of our knowledge to compare *things*. The aim is therefore to be coherent with the knowledge expressed in the semantic proxy considered, without regard on the coherence of the knowledge modelled. As an example, a SM based on a KR built by animal experts will not consider *Sloth* and *Monkey* to be similar, even if most people think sloths are monkeys.

Given that SMs aim at comparing things according to their meaning captured from semantic evidences, it's difficult to further define the notion of SMs without defining the concepts of *Meaning* and *Semantics*.

Taking the risk to disappoint the reader, this section will not face the challenge of the demystification of the notion of *Meaning*. As stressed by (Sahlgren 2006) "*Some 2000 years of philosophical controversy should warn us to steer well clear of such pursuits*". The reader can refer to the various theories proposed by linguists and philosophes. In this contribution, we only consider that we are dealing with the notion of semantic meaning proposed by linguists: *how meaning is conveyed through signs or language*. Regarding the notion of *semantics*, it can be defined as the meaning or interpretation of any lexical units, linguistic expressions or instances semantically characterized according to a specific context.

**Definition**: *Semantic Measures* are mathematical tools used to estimate quantitatively or qualitatively the strength of the semantic relationship between units of language, concepts or instances, through a numerical or symbolic description obtained according to the comparison of information formally or implicitly supporting their meaning or describing their nature.

It is important to stress the diversity of the domain (in a mathematical sense) on which SMs can be used. They can be used to drive word-to-word, concept-to-concept, text-to-text or even instance-to-instance comparison. In this paper we will therefore, as much as possible, refer to any element of the domain of measures through the generic term *element*. An element can therefore be any unit of language (e.g. word, text), a concept/class, an (abstract) instance semantically characterized in a KR (e.g., gene products, ideas, locations, persons)

---

[i] Despite some hesitations and interrogations regarding the notion of similarity, it's commonly admitted that semantic measure design is meaningful. Examples of authors questioning the relevance of notions such as similarity are numerous, e.g. "*Similarity, ever ready to solve philosophical problems and overcome obstacles, is a pretender, an impostor, a quack.*" (Goodman 1972) or "*More studies need to performed with human subjects in order to discover whether semantic distance actually has any meaning independent of a particular person, and how to use semantic distance in a meaningful way*"(Delugach 1993), see also (Murphy & Medin 1985; Robert L Goldstone 1994; Hahn & Ramscar 2001).

We formally define a SM as a function:

$$\sigma_k \colon E_k \times E_k \to \mathcal{R}$$

with $E_k$ the set of elements of type $k \in K$ and $K$, the various types of elements which can be compared regarding their semantics, e.g., $K = \{$words, concepts, sentences, texts, web pages, instances annotated by concepts, $\dots\}$, and $\mathcal{R} = \{[0,1], \ \mathbb{R}^+, \{a, b, c \dots\}\}$.

This expression can be generalized to take into account the comparison of elements of different types. This could be interesting to evaluate entailment of texts or to compare words and concepts to mention a few. However, in this paper, we restrict our study to the comparison of pairs of elements of the same nature (which is already a vast subject of research). We stress that SMs must implicitly or explicitly take advantage of semantic evidences. As an example, as we said in the introduction, measures comparing words through their syntactical similarity cannot be considered to be SMs; recall that semantics refers to evidences regarding the meaning or the nature of compared elements.

The distinction between approaches that can and cannot be assimilated to SMs is sometime narrow; there is no clear border distinguishing non-semantics to semantic-augmented approaches, but rather a range of approaches. Some explanations can be found in the difficulty to clearly characterize the notion of *Meaning*. For instance, someone can say that measures used to evaluate lexical distance between words, such as edit distances, capture semantic evidences regarding the meaning of words. Indeed, the sequence of characters associated to a word derives from its etymology which is sometimes related to its meaning, e.g. words created through morphology derivation such as *subset* from *set*.

Therefore, the notion of SM is sometimes difficult to distinguish from measures used to compare specific data structures. This fine line can also be explained by the fact that some SMs compare elements which are represented, in order to be processed by computer algorithms, through canonical forms corresponding to specific data structures for which specific (non-semantic) similarity measures have been defined. As an example, pure graph similarity measures can be used to compare instances semantically characterized through semantic graphs.

In some cases, the semantics of the measure is therefore not captured by the measure used to compare the canonical forms of the compared elements. It's rather the process of mapping an element (e.g., word, concept) from a semantic space (text, KR) to a specific data structure (e.g., vector, set), which makes the comparison semantically enhanced. This is however an interesting paradox, the definition of the rigorous semantics of the notion of SM is hard to define – this is one of the challenges this contribution tries to face.

### 2.1.2 Semantic Relatedness and Semantic Similarity

Among the various notions associated to SMs, this section defines the two central notions of *semantic relatedness* and *semantic similarity*. They are among the most used in the literature related to SMs. Several authors already distinguished them in different communities, e.g., (Resnik 1999; Pedersen et al. 2007). Based on these works, we propose the following definitions.

**Definition** *Semantic relatedness*: strength of the semantic interactions between two elements without restriction regarding the types of semantic links considered.

**Definition** *Semantic similarity*: specialises the notion of semantic relatedness, by only considering taxonomical relationships in the evaluation of the semantic strength between two elements.

In other words, semantic similarity measures compare elements regarding the properties they share and the properties which are specific to them. The two concepts `Tea` and `Cup` are therefore highly *related* despite the fact that they are not *similar*: the concept `Tea` refers to a `Drink` and the concept `Cup` refers to a `Vessel`. Thus, the two concepts only share few of their constitutive properties. This highlights a potential interpretation of the notion of similarity, which can be understood in term of substitution, i.e., evaluating the implication to substitute the compared elements: `Tea` by `Coffee` or `Tea` by `Cup`.

In some specific cases, communities such as Linguists will consider a more complex definition of the notion of semantic similarity for words. Indeed, word-to-word semantic similarity is sometimes evaluated not only considering (near-) synonymy, or the lexical relations which can be considered as equivalent to the taxonomical relationships for words, e.g., hyponymy and hypernymy or even troponymy for verbs. Indeed, in some contributions, linguists also consider that the estimation of the semantic similarity of two words must also take into account other lexical relationships such as antonymy (S. M. Mohammad & Hirst 2012).

In other cases, the notion of semantic similarity refers to comparison of the elements, not the semantics associated to the results of the measure. As an example, designer of SMs relying on KRs sometimes use the term semantic similarity to denote measures based on a specific type of semantic relatedness which only considers meronymy, e.g., partial ordering of concepts defined by *part–whole* relationships. The semantics associated to the scores of relatedness computed from such restrictions differ from semantic similarity. Nevertheless, technically speaking, as we will see, most approaches defined to compute semantic similarities based on KR can be used on any restriction of semantic relatedness considering a type of relationship which is transitive[i], reflexive[ii] and antisymmetric[iii] (e.g., part-whole relationships). In this paper, for the sake of clarity, we consider that only taxonomical relationships are used to estimate the semantic similarity of compared elements.

Older contributions relative to SMs do not stress the difference between the notions of similarity and relatedness. The reader should be warned that in the literature, authors sometimes introduce semantic similarity measures which estimate semantic relatedness and *vice versa*. In addition, despite the fact that the distinction between the two notions is commonly admitted by most communities, it is still common to observe improper use of both notions.

A large terminology refers to the notion of SMs and contributions related to the domain often refer to the notions of semantic distance, closeness, nearness or taxonomical distance, etc. The following subsection attempts to clarify the semantics associated to the terminology commonly used in the literature.

---

[i] A binary relation $R$ over a set $X$ is transitive if for $a, b, c \in X$, $R(a, b) \wedge R(b, c) \Rightarrow R(a, c)$.

[ii] A binary relation $R$ on a set $X$ is reflexive if $\forall a \in X$, $R(a, a)$.

[iii] A binary relation $R$ on a set $X$ is antisymmetric if for $a, b \in X$, $R(a, b) \wedge R(b, a) \Rightarrow a \equiv b$.

### 2.1.3 The Diversity of Types of Semantic Measures

We have so far introduced the broad notion of SMs. We also distinguished the two notions of semantic relatedness and semantic similarity. A large terminology has been used in the literature to refer to the notion of SM. We define the meaning of the terms commonly used (the list may not be exhaustive):

- *Semantic relatedness*, sometimes called *proximity*, *closeness* or *nearness*, refers to the notion introduced above.
- *Semantic similarity* has also already been defined. In some cases, the term *taxonomical semantic similarity* is used to stress the fact that only taxonomical relationships are used to estimate the similarity.
- *Semantic distance*: Generally considered as the *inverse* of the semantic relatedness, all semantic interactions between the compared elements are considered. These measures respect (most of the time) the mathematical properties of distances. These properties will be introduced in subsection 2.3.1. Distance is also sometimes denoted *farness*.
- *Semantic Dissimilarity* is understood as the inverse of semantic similarity.
- *Taxonomical distance* also corresponds to the semantics associated to the notion of dissimilarity. However, these measures are expected to respect the properties of distances.

**Figure 1** presents a chart in which the various notions related to SMs are structured through semantic relationships. Most of the time, the notion considered to be the inverse of semantic relatedness is denoted semantic distance, without regard if the measure respects the mathematical properties characterizing a distance. Therefore, we introduce the term *semantic unrelatedness* to denote the set of measures which have a semantics which is the inverse to the one carried by semantic relatedness measures, without necessary respecting the properties of a distance. This is, to our knowledge, a notion which has never been used in the literature[i].

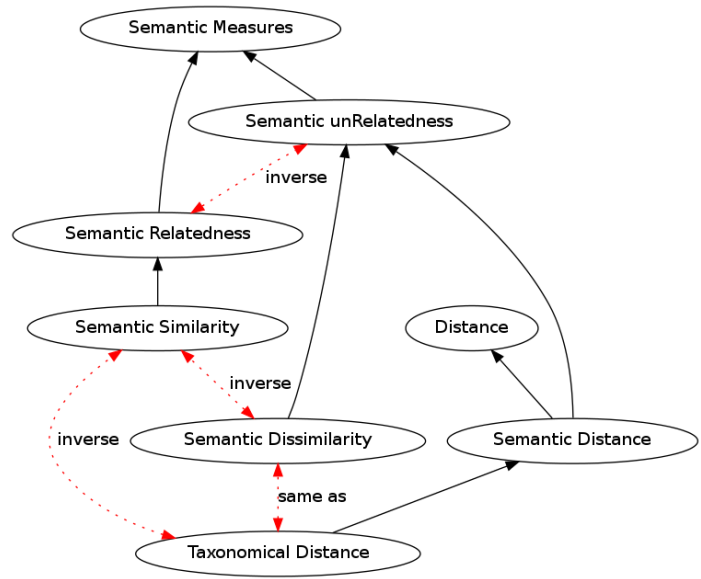

**Figure 1**: Semantic graph defining the relationships between the various types of semantics which can be associated to SMs in the literature. Black (plain) relationships correspond to taxonomical relationships (i.e., subclass-of), inverse relationships refer to the semantic interpretation associated to the score of the measure, e.g., semantic similarity and dissimilarity measures have inverse semantic interpretations.

---

[i] Our aim is not here to make the terminology heavier but rather to be rigorous in the characterization of measures.

## 2.2 Cognitive Models of Similarity

In this subsection, we provide a brief overview of the psychological theories of similarity by introducing the main models proposed by cognitive sciences to study and explain (human) appreciation of similarity. Here we do not discuss the notion of semantic similarity; the process of similarity assessment should be understood in a broad sense, i.e., as a way to compare objects, stimuli.

Cognitive models of similarity generally aim to study the way human evaluate the similarity of two mental representations according to some kind of psychological space. They are based on assumptions regarding the KR from which the similarity will be estimated. Indeed, as stated by several authors, the notion of similarity, *per se*, can be criticized as a notion purely artificial. In (Goodman 1972), the notion of similarity is defined as "*an imposture, a quack*" as objectively, everything is equally similar to everything else. This can be disturbing but, conceptually, two random objects have infinitively properties in common and infinitively different properties[i], e.g. a flower and a computer are both smaller than 10m, 9.99m, 9.98m…. An important notion to understand, which have been underlined by cognitive sciences, is that differential similarities emerge only when some predicates are selected or *weighted* more than others. This observation doesn't mean that similarity is not an explanatory notion but rather that the notion of similarity is heavily framed in psychology. Similarity assessment must therefore not be understood as an attempt to compare object realizations through the evaluation of their properties, but rather as a process aiming to compare objects as they are understood by the *agent* which rule the estimation of the similarity (e.g., a person). The notion of similarity, therefore, only makes sense according to the consideration of a partial mental representation from which the estimation of objects will be based on.

Contrary to real objects, representations of objects do not contain infinitesimal properties. As an example, our mental representations of things only capture a limited number of dimensions of the object which is represented. Therefore, the philosophical worries regarding the soundness of similarity vanish considering that similarity aim at comparing partial representations of objects, e.g., human mental representation of objects (Hahn 2011). The similarity is thus estimated between mental representations. Considering that these representations are defined by a human agent, the notion of similarity may thus be understood as how similar objects appear to us. Considering the existential requirement of representations to compare things and to consider similarity as a meaningful notion, much of the history of research on similarity in cognitive sciences focuses on the definition of models of mental representation of objects.

The central role of cognitive sciences regarding the study of similarity relies in the design of cognitive models of both, mental representations and similarity. These models are further used to study how humans store their knowledge and interact with it to compare objects represented as pieces of knowledge. They next test these models according to our understanding of human appreciation of similarity. Indeed, evaluations of human appreciation of similarity help us to distinguish constraints/expectations on the properties an accurate model should have, which is essential to reject hypothesis and improve the models. As an example, studies have demonstrated that appreciation of similarity is sometimes asymmetric: the similarity between a painter and his portrait is commonly expected to be greater than the inverse - isn't it? Therefore, the expectation of asymmetric estimation of similarity is incompatible with the mathematical properties of a distance, which is symmetric by definition. Models based on distance axioms have therefore to be re-vised or to be used with moderation. In this context, the introduction of cognitive models of similarity will be particularly useful to understand the foundations of some approaches adopted for the definition of SMs.

---

[i] This statement also stands if we restrict the comparison of objects to finite set of properties. The reader may refer to Andersen's famous story of the *Ugly Duckling*. Proved by (Watanabe & Donovan 1969), the Ugly Duckling theorem highlights the intrinsic bias associate to classification, showing that all things are equal and therefore that an ugly duckling is similar to a swan as two swan are to each other. The important teaching is that inductive biases are required to make a judgment and to classify, i.e., to prefer certain categories over others.

Cognitive models of similarity are commonly organized in four different approaches; (i) Spatial models, (ii) Feature models, (iii) Structural Models and (iv) Transformational models. We briefly introduce these four models; a more detailed introduction can be found in (Goldstone & Son 2004) and (Schwering 2008). A captivating talk introducing to cognition and similarity, on which is based this introduction, can be also be found in (Hahn 2011).

### 2.2.1 Spatial Models

The spatial models, also named geometric models, rely on one of the most influencal theory of similarity in cognitive sciences. They are based on the notion of psychological distance and consider objects (here perceptual effects of stimuli or concepts for instance) as points in a multi-dimensional metric space.

Spatial models consider similarity as a function of the distance between the mental representations of the compared objects. These models derive from Shepard's spatial model of similarity, in which objects are represented in a multi-dimensional space. The locations of the objects are defined by their dimensional differences (Shepard 1962).

In his seminal and highly influencal work on generalization, (Shepard 1987) provides a statistical technique in the form of Multi-Dimensional Scaling (MDS) to derive the locations of objects represented in a multi-dimensional space. Indeed, MDS can be used to derive some potential spatial representations of objects from proximity data (similarity between pairs of objects). Based on these spatial representations of objects, Shepard derived the *universal law of generalization* which demonstrates that various kinds of stimuli (e.g., Morse code signals, shapes, sounds) have the same lawful relationship between distance (in an underlined MDS) and perceive similarity measures (in term of confusability) - the similarity between two stimuli was defined as an exponentially decaying function of their distance[i].

By demonstrating a negative exponential relationship between similarity and generalization Shepard established the first sounded model of mental representation on which cognitive sciences will base their studies on similarity[ii]. The similarity is in this case assumed to be inversely proportional to the distance which separates the perceptual representations of the compared stimuli (Ashby & Perrin 1988). The similarity defined as a function of distance is therefore constrained to the axiomatic properties of distance (properties which will be detailed in the following section).

A large number of geometric models have been proposed. They have long been among the most popular in cognitive sciences. However, despite their intuitive nature and large adoptions, geometric models have been subject to intense criticisms due to the constraints defined by the distance axioms. Indeed, several empirical analysis have questioned and challenged the validity of the geometric framework (i.e., both the model and the notion of psychological distance), by underlying inconsistencies with human appreciation

---

[i] The similarity between two stimuli is here understood as the probability that a response to a stimulus will be generalized to the other (Shepard 1987). With $sim(A, B)$ the similarity between two stimuli $A, B$ and $dist(A, B)$ their distance, we obtain the tion $sim(A, B) = e^{-dist(A,B)}$, that is $dist(A, B) = -\log \, sim(A, B)$, a form of entropy.

[ii] A reason which also explains the success encountered by spatial models is to find in their central role in another highly successful formal model provided by psychology studies: the (Nosofsky 1986) *generalized context model of classification*. Nosofsky showed that a classification task, i.e., the probability that an item belongs to a category, can be explained as a function of the sum of the similarities between the item to categorize and all known items of a category normalized by the sum of the similarities between the item to categorize and all the other items (Hahn 2011).

of similarity, e.g., violation of the symmetry, triangle inequality and identity of the indiscernibles, e.g., (Tversky 1977; Tversky & Gati 1982)[i].

### 2.2.2 Feature Models

This approach, introduced by (Tversky 1977) to answer the limitation of the geometric models, proposes to characterize an object through a set of features, considering that a feature "*describes any property, characteristic, or aspect of objects that are relevant to the task under study*" (Tversky & Gati 1982). Feature models evaluate the similarity of two stimuli according to a feature-matching function $F$ which makes use of their common and distinct features:

$$sim(u, v) = F(U \cap V, U \backslash V, V \backslash U)$$

The function $F$ is expected to be non-decreasing, i.e., the similarity increases when common (distinct) features are added (removed). The feature model is therefore based on the assumption that $F$ is monotone and that common and distinct features of compared objects are sufficient for their comparison. In addition, an important aspect is that the feature-matching process is expressed in term of matching function as defined in set theory (i.e., binary evaluation).

The similarity is further derived as a parameterized function of the common and distinct features of the compared objects. Two models, the *contrast model* ($sim_{CM}$) and the *ratio model* ($sim_{RM}$) have initially been proposed by Tversky. They can be used to compare two objects $u$ and $v$ represented through sets of features $U$ and $V$:

$$sim_{CM}(u, v) = \gamma f(U \cap V) - \alpha\, f(U \backslash V) - \beta\, f(V \backslash U)$$

$$sim_{RM}(u, v) = \frac{f(U \cap V)}{\alpha\, f(U \backslash V) + \beta\, f(V \backslash U) + f(U \cap V)}$$

The symmetry of the measures produced by the two models can be tuned according to the parameters $\alpha$ and β. This enables the design of asymmetric measures. In addition, one of the major constructs of the feature model is the function $f$ which is used to capture the *salience* of a (set of) feature(s).

The salience of a feature is defined as a notion of specificity: "*the salience of a stimulus includes intensity, frequency, familiarity, good form, and informational content*" (Tversky 1977). Therefore, the operators ∪,∩ and \ are based on feature matching and the function $f$ evaluates the contribution of the common or distinct features (distinguished by previous operators) to estimate the similarity. Notice that the concept of the salience of features implicitly defines the possibility to design measures which do not respect the identity of the indiscernibles, i.e. which enable non-maximal self-similarity.

---

[i] Note that recent contributions propose to answer these inconsistencies by generalizing the classical geometric framework through quantum probability. Compared objects are represented in a quantum model in which they are not seen as points or distributions of points, but entire subspaces of potentially very high dimensionality, or probability distributions of these spaces (Pothos et al. 2013).

### 2.2.3 Structural Alignment Models

Structural Models are based on the assumption that objects are represented by structured representations. Indeed, a strong critic of the feature model was that (features of) compared objects are considered to be unstructured, contrary to evidences suggesting that perceptual representations are well characterized by hierarchical systems of relations, e.g., (Markman & Gentner 1993; Gentner & Markman 1994).

Structural alignment models are structure mapping models in which the similarity is estimated using matching functions which will evaluate the correspondences between the compared elements (Markman & Gentner 1993; Gentner & Markman 1994). The process of similarity assessment is here expected to involve a structural alignment between two mental representations in order to distinguish matches - the more the number of correspondences, more similar the objects will be considered. In some cases, the similarity is estimated in an equivalent manner to analogical mapping (Markman & Gentner 1990) and similarity is expected to involve mapping between both features and relations.

Another example of structural model was proposed by (R.L. Goldstone 1994; Goldstone 1996). The authors proposed to model similarity as an interactive activation and mapping model using a connectionism activation networks based on mappings between representations.

### 2.2.4 Transformational Models

Transformational models assumes that similarity is defined by the transformational distance between mental representations (Hahn et al. 2003). The similarity is framed in *Representational Distortion* (Chater & Hahn 1997) and is expected to be assessed based on the analysis of the *modifications* required to transform one representation to another. The similarity, which can be explained in terms of Kolmogorov complexity theory (Li & Vitányi 1993), is therefore regarded as a decreasing function of transformational complexity (Hahn et al. 2003).

### 2.2.5 Unification of Cognitive Models of Similarity

Several studies highlighted correspondences and deep parallels between the various cognitive models. (Tenenbaum & Griffiths 2001) propose a unification of spatial, feature-based and structure-based models through a framework relying on generalization of Bayesian inference (see (Gentner 2001) for critics). Alternatively (Hahn 2011) propose an interpretation of the models in which the transformational model is presented as a generalization of the spatial, feature and structure-based models.

In this section, we have presented several cognitive models proposing to explain and study (human) appreciation of similarity. These models are characterized by particular interpretations and assumptions on the way knowledge is characterized, mentally represented, and processed. Despite several meaningful initiatives for the unification of the cognitive models in order to develop framework generalizing existing models, we have stressed that one of the fundamental differences between the models rely on their mathematical properties, e.g., symmetry, triangle inequality... The next section proposes an overview of mathematical notions required to rigorously manipulate the notions of similarity and distance. Several mathematical properties which can be used to characterize semantic measures are next introduced.

## 2.3 From Distance Metrics and Similarities to Semantic Measures

Are SMs mathematical measures? What are the specific properties of a distance or a similarity measure? Do semantic similarity measures correspond to similarity measures in the way mathematicians understand them? As we have seen in section 2.1, contributions related to SMs most of the time do not rely on formal definitions of the notion of measures or distance. Indeed, generally, contributions related to SMs rely on the commonly admitted and intuitive expectations regarding these notions, e.g. similarity (resp. distance) must be higher (resp. lower) the more (resp. less) the two compared elements share commonness[i]. However, the notions of measure and distance have been rigorously defined in mathematics through specific axioms from which particular properties derive. These notions have been expressed for weel defined objects (element domain). Several contributions rely on these axiomatic definitions and interesting results have been demonstrated according to them. This section briefly introduces the mathematical background relative to the notions of distance and similarity. It will help us to rigorously define and better characterize SMs in mathematical terms; this is a prerequisite to clarify the fuzzy terminology commonly used in studies related to SMs.

For more information on the definition of measures, distance and similarity, the reader can refer to: (i) the seminal work of (Deza & Deza 2013) – Encyclopedia of Distances, (ii) the work of (Hagedoorn & others 2000) – chapter 2, a theory of similarity measures, and (iii) the definitions proposed by (D'Amato 2007). Most of the definitions proposed in this section have been formulated based on these contributions, and more particularly based on (D'Amato 2007). Therefore, for convenience, we will not systematically refer to them. In addition, contrary to most of the definitions presented in these works, we here focus on highlighting the semantics of the various definitions according to the terminology introduced in section 2.1.

### 2.3.1 Mathematical Definitions and Properties of Distance and Similarity

For the definitions presented hereafter, we consider a set $D$ which defines the elements of the domain we want to compare and a totally ordered set $(V, \preccurlyeq)$, with $min_V$ the element of $V$ such as $\forall v \in V : min_V \preccurlyeq v$ and $max_V \in V$ such as $\forall v \in V : v \preccurlyeq max_V$.

**Definition** *Distance*: a function $dist : D \times D \to V$ is a distance on $D$ if, $\forall x, y \in D$, the function is:

- Non-negative, $dist(x, y) \succcurlyeq min_V$ and $min_V = 0$.
- Symmetric, $dist(x, y) = dist(y, x)$.
- Reflexive $dist(x, x) = min_V$ and $\forall y \in D \wedge y \neq x : dist(x, x) < dist(x, y)$.

To be considered as a *distance metric*, i.e., a distance in a metric space, the distance must additionally respect two properties:

- The *identity of indiscernibles* also known as strictness property, minimality or self-identity, that is $dist(x, y) = min_V$ if and only if $x = y$.
- The *triangle inequality*, the distance between two points must be the shortest distance along any path: $dist(x, y) \leq dist(x, z) + dist(z, y)$.

---

[i] The works of (D'Amato 2007; Blanchard et al. 2008) are among the exceptions.

Despite the fact that some formal definitions of similarity have been proposed, e.g., (Hagedoorn & others 2000; Deza & Deza 2013), contrary to the notion of distance, there is no axiomatic definition of similarity that sets the standard in mathematics. The notion of similarity appears in different fields of mathematics, e.g., figures with the same shape are denoted similar (in geometry), similar matrices are expected to have the same eigenvalues, etc. In this paper, we consider the following definition.

**Definition** *Similarity*: a function $sim: D \times D \to V$ is a similarity on $D$ if, for all $x, y \in D$, the function $sim$ is non-negative ($min_V = 0$), symmetric and reflexive, i.e., $sim(x, x) = max_V$ and $\forall y \in D \wedge y \neq x : sim(x, x) > sim(x, y)$.

**Definition** *Normalized function*: Any function $f$ on $D$ with values in $V$ (e.g. similarity, distance) is said to be normalized if: $\forall x, y \in D : 0 \leq f(x, y) \leq 1$, i.e., $min_V = 0$ and $max_V = 1$.

Notice that a normalized similarity $sim$ can be transformed to a distance $dist$ considering multiple approaches (Deza & Deza 2013). Inversely, a normalized distance can also be converted to a similarity. Some of the approaches used for the transformations are presented in appendix 5.

As we have seen, distance and similarity measures are formally defined in mathematics as functions with specific properties. They are most of the time defined considering $min_V = 0$. They are extensively used to demonstrate results and develop proofs. However, the benefits to fulfil some of these properties, e.g., *triangle inequality* for distance metric, have been subject to debate among researchers. As an example, (Jain et al. 1999) stress that the mutual neighbour distance used in clustering tasks do not satisfy the triangle inequality but perform well in practice - to conclude by "*This observation supports the viewpoint that the dissimilarity does not need to be a metric*".

A large number of properties not presented in this section have been distinguished to further characterize distance or similarity functions, e.g. (Deza & Deza 2013). These properties are important as specific theoretical proofs require studied functions to fulfil particular properties. However, as we have seen, the definition of SMs proposed in the literature is not framed in the mathematical axiomatic definitions of distance or similarity. In some case, such a distortion among the terminology creates difficulties to bridge the gap between the various communities. As an example, in the encyclopaedia of distances, (Deza & Deza 2013) do not distinguish the notions of distance and dissimilarity, which is the case in the literature related to SMs (refer to section 2.1.3). In this context, the following section defines the terminology commonly adopted in the study of SMs, with regard to the mathematical properties already introduced.

### 2.3.2 Flexibility of Semantic Measures Regarding Mathematical Properties

Notice that we didn't introduced the precise and technical mathematical definition of a *measure* proposed by measure theory. This can be disturbing considering that this paper extensively refers to the notion of SM. The notion of measure we use is indeed not framed in the rigorous mathematical definition of the mathematical concept of *measure*. It refers to any "*measuring instruments*" which can be used to "*assess the importance, effect, or value of (something)*" (Oxford dictionary 2013) – in our case, any functions answering the definitions of semantic distance/relatedness/similarity… proposed in section 2.1.

Various communities have used the concepts of similarity or distance without considering the rigorous axiomatic definitions proposed in mathematics but rather using their broad intuitive meanings[i]. To be in accordance with most contributions related to SMs, and to facilitate the reading of this paper, we will not limit ourselves to the mathematical definitions of distance and similarity.

The literature related to SMs generally refers to a semantic distance as any (non-negative) function, designed to capture the inverse of the strength of the semantic interactions linking two elements. Such functions must respect: the higher the strength of the semantic interactions between two elements is, the lower their distance. The axiomatic definition of a distance (metric) may not be respected. A semantic distance is most the time what we define as a function estimating *semantic unrelatedness* (please refer to the organization of the measures proposed in section 2.1.3). However, to be in accordance with the literature, we will use the term semantic distance to refer to any function designed to capture semantic unrelatedness. We will explicitly precise that the function respects (or not) the axiomatic definition of a distance (metric) when required.

Semantic relatedness measures are functions which are associated to an inverse semantics to the one associated to semantic unrelatedness: the higher the strength of the semantic interactions between two elements is, the higher the function will estimate their semantic relatedness.

In this paper, the terminology we use (distance, relatedness, similarity), refers to the definitions presented in sections 2.1.2 and 2.1.3. To be clear, the terminology refers to the semantics of the functions, not their mathematical properties. However, we further consider that SMs must be characterized through mathematical properties. **Table 1** and **Table 2** summarize some of the properties which can be used to formally characterize any function designed in order to capture the intuitive notions of semantic distance and relatedness/similarity. These properties will be used in the next sections to characterize some of the measures we will consider. They are essential to further understand the semantics associated to the measures and to distinguish SMs which are adapted to specific contexts and usages.

| Properties | Definitions |
|---|---|
| Non-negative | $dist(x, y) \geq min_V$ and $min_V = 0$ |
| Symmetric | $dist(x, y) = dist(y, x)$ |
| Reflexive | $dist(x, x) = min_V$ |
| Normalized | $0 \leq dist(x, y) \leq 1$, i.e. $min_V = 0$ and $max_V = 1$ |
| identity of indiscernibles | $dist(x, y) = min_V$ if and only if $x = y$ |
| Triangle inequality | $dist(x, y) \leq dist(x, z) + dist(z, y)$ |

**Table 1**: Properties which can be used to characterize any function which aims to estimate the notion of distance between two elements.

---

[i] As we have seen, researchers in cognitive science have demonstrated that human expectations regarding (semantic) distance challenges the mathematical axiomatic definition of distance. Thus, the communities involved in the definition of SMs mainly consider a common vision of these notions without always clearly defining their mathematical properties.

| Properties | Definitions |
|---|---|
| Non-negative | $sim(x,y) \geq min_V$ and $min_V = 0$ |
| Symmetric | $sim(x,y) = sim(y,x)$ |
| Reflexive | $sim(x,x) = max_V$ |
| Normalized | $0 \leq sim(x,y) \leq 1$ i.e. $min_V = 0$ and $max_V = 1$ |
| identity of indiscernibles | $sim(x,y) = max_V$ if and only if $x = y$ |
| Integrity | $sim(x,y) \leq sim(x,x)$ |

**Table 2**: Properties which can be used to characterize any function which aims to estimate the notion of similarity/relatedness between two elements.

We have so far introduced the cognitive models used to study the notions of similarity as well as the formal mathematical definitions of the notion of distance and similarity. Several mathematical properties which can be used to characterize SMs have also been presented. Before we look at the classification of SMs we will first introduce the notion of knowledge representation (KR).

## 2.4 A Brief Introduction to Knowledge Representations

Here we use the term *computational* or *formal knowledge representation* (KR in short), to refer to any computational model or computational artefact used to formally express knowledge in a machine understandable form[i]. This section doesn't aim to introduce the reader to knowledge engineering or to present the various computational models which can be used to formally express knowledge in a machine understandable form. We only aim to give an overview of the notion of KRs in order to introduce SMs which take advantage of them. For more information, the reader can refer to some of the seminal contributions related to the topic, e.g., (Minsky 1974; Sowa 1984; Davis et al. 1993; Gruber 1993; Borst 1997; Studer et al. 1998; Baader 2003; Guarino et al. 2009; Robinson & Bauer 2011; Hitzler et al. 2011).

### 2.4.1 Generalities

From simple taxonomies and terminologies to complex KRs based on logics, a large spectrum of approaches has been proposed to express knowledge. **Figure 2** presents several approaches which can be used to express KRs going from weak semantic descriptions of terms and linguistic relationships, to more refined and complex *conceptualization* associated to strong semantics.

---

[i] The notion of Computational/formal knowledge representation refers to the notion introduced by (Davis et al. 1993) in which they refer to *knowledge representation technologies*. As stress in (Guarino et al. 2009), *'For AI Systems, what exists is that which can be represented'*

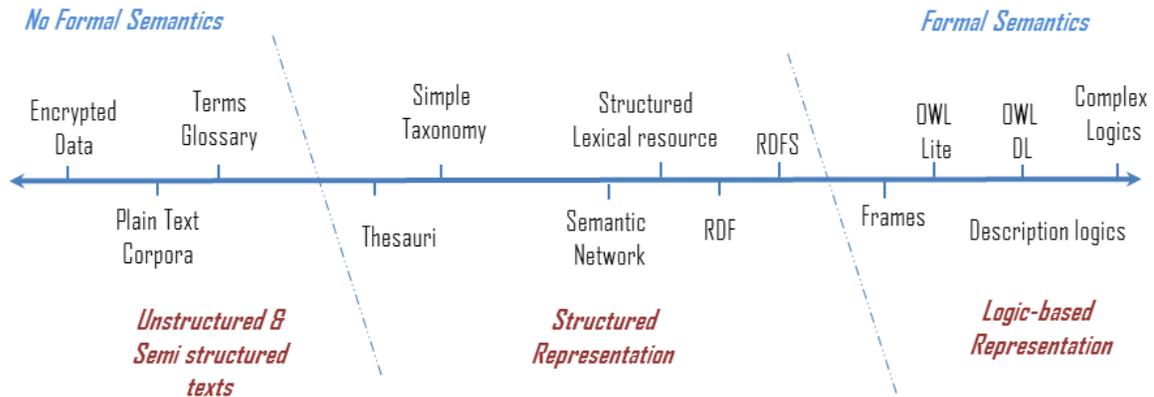

**Figure 2**: From non-formal semantics to formal knowledge representations, adapted from (Jimeno-Yepes et al. 2009).

The challenging problematic tackled by the field of knowledge engineering, *how to formally express knowledge*, have gained a lot of attention in artificial intelligence. Such a success is naturally explained by the large implications of formal expressions of knowledge in computer science, it gives computers and algorithms an access door to our knowledge. Therefore, over the last decades, contributions relative to computational KRs have been numerous. Ontology is probably the most famous and mysterious word of this domain. It has been overused to the point of becoming a propaganda tool, and to be honest, as well as reinsuring those which have been *lost in the translation*, it is today difficult to find two knowledge engineers which will give the same definition of an ontology – no offense to the seminal contributions which focused on the demystification of the notion of ontology, e.g. (Guarino et al. 2009), it's today a concrete reality in both academic and industrial fields. Indeed, based on (Gruber 1993; Borst 1997), an ontology is often defined in highly abstract terms, as *"a formal, explicit specification of a shared conceptualization"* (Studer et al. 1998). However, despite its popularity, this definition relies on the informal and rarely questioned definition of *conceptualization* (Guarino et al. 2009).

As we will see, a conceptualization can partially be seen as a formal expression of a set of concepts and instances within a domain, as well as the relationships between these concepts/instances. For others, a conceptualization of a domain relies on the definition of its vocabulary. In this section, based on (Guarino et al. 2009), we will adopt a specific definition of the notions of ontology and conceptualization; these notion are (commonly) admitted in knowledge engineering but may differ to usages in other communities.

As we already said, we will use the general term KR to refer to all formal and machine understandable representation of knowledge, e.g. all the range of structured and logic-based approaches which appear in **Figure 2**. This choice has been motivated by the difficulty to consider several formal KRs as ontologies (and by the will to differentiate them from conceptualizations), e.g., lexical databases or term-oriented models of thesaurus structure which use terms as primitive elements and not concepts (thesaurus, classification scheme, etc.). However, to ease the presentation of the various KRs, we denote any abstract view of a set of common things through the generic term *class* or *concept*.

### 2.4.2 Knowledge Representations as Conceptualization

Regardless of the particularities of some domain-specific KRs and regardless of the language considered for the modelling, all approaches used to represent knowledge share common components:

- *Classes* or *Concepts* used to denoted set of things sharing common properties, e.g., `Human`.
- *Instances*, i.e. members of classes, e.g., `alan` (an instance of the class `Human`).
- *Predicates,* the types of relationships which define the semantic relationships which can be established between instances or classes, e.g., taxonomic relationships.
- *Relationship*s, concrete links between classes and instances which carry a specific semantics, e.g., `(alan, isA, Human)`, `(alan, worksAt, Bletchley Park)`
- *Attributes*, properties of instances or classes, e.g.; `(Alan, hasName, "Turing")`
- *Axioms,* for instance defined through properties of the predicates, e.g. *"taxonomical relationships are transitive"*, or constraints on properties and attributes, e.g., *"Any Human has exactly 2 legs"*.

A simple knowledge representation (KR) can therefore be formally defined by $O : \{C, R, E, A^O\}$, with:

- $C$ the set of classes/concepts and $R$ the set of predicates which can be used to link two classes or two predicates, e.g., `{subClassOf, partOf, subPredicateOf}`. A predicate $r \in R$ is also named a type of relationship. The set of classes and predicates are expected to be disjoint, i.e. $C \cap R = \emptyset$.
- $E \subseteq C \times R \times C \cup R \times R \times R$, the set of oriented relationships of a specific type $r \in R$ which link a pair of classes or a pair of predicates. Any relationship is therefore characterized by a triplet $(u, t, v)$ with $u, v \in C$ or $u, v \in R$ and $t \in R$. Note that a triplet is also called a statement.
- $A^O$ a set of axioms defining the interpretations of each class and predicat.

Only considering $O : \{C, R, E\}$ leads to a labelled graph structuring classes and relationships through labelled oriented edges. A vocabulary can be associated to any class and predicate. In addition, a lexical reference (didactical device) is generally considered to be used to refer, in an unambiguous manner, to a specific class/predicate, e.g., the string `Mammal` refers to a specific clade of animals. In practice, the unique identifier is generally an Internationalized Resource Identifier (IRI) [i].

The semantics of a class/predicate is so far implicitly defined by the definition of unambiguous (lexical) references. As we will see, the formal semantics of the KR expressed in the graph is specified by the set of axioms $A^O$.

The set of axioms $A^O$ defines that $O : \{C, R, E\}$ is not a simple graph data structure. They will be used to define the interpretations of the classes and predicates. The set of axioms for instance defines the properties associated to the predicates. Among the numerous properties which can be used to characterize predicates, we can distinguish the transitivity, reflexivity and anti-symmetry. These specific properties characterize taxonomical relationships used to define a partial ordering among classes and predicates. We name `subClassOf` the taxonomical relationship which specifies that a class is subsumed by another[ii] and `sub-`

---

[i] Which is a generalization of the Uniform Resource Identifier (URI) in which specific characters (from ISO 10646) can be used.

[ii] In some contributions, the `isA` relationship commonly refers to the taxonomical relationship. However, the relationships characterized by the predicate `subClassOf` define taxonomies of concepts/classes and the those corresponding to `isA` relationship are used to type instances, i.e. to define that an instance is of a specific type (see `rdf:type` in RDF specification).

`PredicateOf` the taxonomical relationship defining that a predicate inherits from another predicate[i]. As an example, it can be defined that:

> Mammal subClassOf Animal
> isFatherOf subPredicateOf isParentOf

The semantics associated to the predicates through the axiomatic characterization of their properties leads to the definition of both taxonomies of classes and predicates. These taxonomies are expressed by the relationships defined in $E$. We distinguish:

- $\preccurlyeq_C \subseteq E$, the taxonomy of classes which defines a partial ordering of classes. We note $u \prec v$ if $u$ is subsumed by $v$ (or $v$ subsumes $u$). As an example, the statement `Mammal` $\prec$ `Animal` formally means that all mammals are animals. The taxonomical relationship between classes is the type of relationship the most represented in ontologies. Therefore, for convenience, we use the term taxonomical relationship to refer to taxonomical relationship between classes. The taxonomy of classes forms a Directed, Acyclic, and connected Graph (DAG). Indeed, a unique class, denoted the *root*, is generally considered to subsume all the classes. A fictive root is considered if none is explicitly specified (e.g., `Thing`).
- $\preccurlyeq_R \subseteq E$ the taxonomic structuration of the predicates, `isFatherOf` $\prec$ `isParentOf`. Like for $\preccurlyeq_C$, a root predicate is expected to be defined and the graph defined by $\preccurlyeq_R$ is also a directed, acyclic, and connected.

$A^O$ can further describe constraints on interpretations associated to predicates by defining a domain and a range (co-domain) of any predicate of $R$. They can be used to define a specific interpretation of a type of relationship, e.g., the domain and the range of the predicate `isParentOf` are defined to be the class `Person`, which means that for all statement `(x,isParentOf,y)` we can infer that `x` and `y` are members of the class `Person`.

The axiomatic definition of classes and predicates can be based on a large variety of logical constructs (e.g., negation, conjunction, and disjunction). They are used to further constraint the interpretation of classes and predicates, and enable more complex expressions of knowledge. The presentation of the various logical constructors which can be used is out of the scope of this paper; please refer to (Baader 2003) for an introduction to logic-based KR. Hereafter, we only briefly present some knowledge expressions which can be based on logical constructs:

- The classes `Man` and `Women` are disjoint as the sets of instances of the two classes are expected to be disjoint, i.e., $Man \sqcap Woman \equiv \bot$, that is to say, with $I(Man)$, the set of instances of the class $Man$, we have $I(Man) \cap I(Woman) = \emptyset$.
- The class `Man` refers to instances which are both `Male` and `Person`, i.e., $Man \equiv Male \wedge Person$.
- The class `Man` refers to instances which are `Person` and not `Women`, i.e., $Man \equiv Person \sqcap \neg\, Woman$.
- The predicate `subClassOf` is transitive, i.e., `Man` $\prec$ `Person` $\Rightarrow I(Man) \subseteq I(Person)$.

---

[i] Generally named `subPropertyOf`, e.g, in RDFS.

The literature generally distinguishes lightweight ontologies from highly formal ontologies depending on the degree of complexity of $A^O$ (e.g., predicate properties, logical constructs).

### 2.4.3 More Refined Knowledge Representations

In knowledge modelling, the general abstract KR $O: \{C, R, E, A^O\}$, (i.e., classes and predicates definitions, class relationships and axioms) is generally distinguished from the knowledge relative to the instances of the considered domain. The conceptualization of the general abstract knowledge is named the TBox (Terminological Box). The statement *'All Mammals are Animals'*, i.e. (`Mammal`, `subClassOf`, `Animal`), is an example of statement found in the TBox. However, KR is not only about conceptualization. In some cases you may also want to express knowledge about specific instances of your domain, i.e., specific realization of the classes defined in $O$.

Knowledge relative to instances of a domain are expressed in the form of statements, e.g., (`bob`, `isA`, `Man`). These statements must be compliant with the TBox. As an example, if it's defined that `Man` and `Women` are two disjoint classes, the conceptualization is violated by the definition of both statements (`bob`, `isA`, `Man`) and (`bob`, `isA`, `Women`). The set of statements related to instances are defined by the term ABox (Assertional box).

To be compliant with the introduction of information relative to instances, the formal definition of a KR introduced above must be revisited. To this end, we introduce a set of instance $I$ and we authorize both the definition of statements between instances and between both instances and class, $E \subseteq C \times R \times C \cup R \times R \times R \cup I \times R \times C \cup I \times R \times I$. We denote $I(X)$ the set of instances of the class $X$.

Note that in some cases, data values of specific data type can also be used to further characterize classes or instances, e.g. to specify the age of a person (`bob`, `hasAge`, `52`). Example of models introducing attributes of specific data types to classes can be found in (Ehrig et al. 2004). In this case, a set of datatypes and their structuration $\leqslant_D$ can be defined. A set of attributes of a specific type can therefore be associated to a data type. An attribute can be represented as a specific predicate (type of relationship). A data value can therefore be considered as an instance of a specific data type[i] and a specific semantic relationship between a concept and a data value can be used to represent the value of an attribute which is associated to a class.

Formally, to enable data value to be used considering the model introduced so far, it is needed to:

- Distinguish a set of data values associated to specific data type (which can also be structured in some cases),
- Further extends $E$ such as statements enabling data value to be used are possible.

However, to facilitate the reading we will not introduce further notations.

---

[i] This is a figment of imagination, RDF graph do not enable Literals to be used as subject of a triple, it is therefore not possible to express statement such as ("London", rdf:type, String). More information at http://www.w3.org/TR/2011/WD-rdf11-concepts-20110830

### 2.4.4 Knowledge Representations as Semantic Graphs

In this paper, we consider a *semantic graph* as any declarative KR in which unambiguous resources are represented through nodes interconnected through semantic relationships associated to a controlled semantics. We therefore consider that any synsets/concepts/classes structured through a semantic graph can be considered in an equivalent manner[i]. A semantic graph is therefore a specific type of KR in which the axiomatic definitions do not relies on complex logical constructs such as negation, conjunction, disjunction and so far (e.g., lightweight ontologies).

A semantic graph is composed of a set of statements, each of them composed of a triplet subject-predicate-object, e.g. (`Human, subClassOf, Mammal`). The name of a class will be used to refer to both the conceptualization and the corresponding node in the graph. Any node which refers to a realization of a class will be named an *instance*. Notice that we always use the term *relationship* to refer to a binary relationship between a subject and an object even if we consider that all relationships are associated to a predicates.

**Figure 3** presents a basic example of a simple KR represented as a semantic graph. The graph structures few classes through taxonomical relationships (plain black relationships) and relationships carrying a specific meaning (e.g., *hunts*).

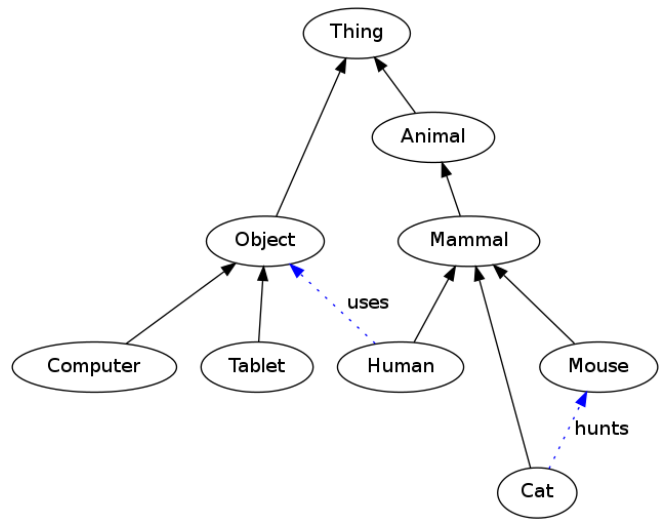

**Figure 3**: Basic example of a semantic graph defining a set of classes, their taxonomical structuration (plain black relationships), and other relationships (dotted), e.g. (`Cat, hunts, Mouse`).

As we have seen, some semantic graphs represent a KR which not only contains TBox statements but also knowledge relative to the instances of the domain. As an example, in WordNet, instances are distinguished from the classes. **Figure 4** presents a more complex representation of a semantic graph which corresponds to a KR involving classes, predicates, instances and data values. Various classes defined in the graph are taxonomically structured in the layer *C*. Several types of instances are also defined in layer *I*, e.g. music bands, music genres. These instances can be characterized according to specific classes, e.g., (`rollingStones, isA, MusicBand`) and can be interconnected through specific predicate, e.g., (`rollingStones, hasGenre, rock`). In addition, specific data values (layer *D*) can be used to specify information relative to both classes and instances, e.g., (`rollingStones, haveBeenFormedIn, 1962`). All relationships linking the various nodes of the graph are directed and semantically characterized, i.e., they carry an unambiguous and controlled semantics. Notice that extra information such as the taxonomy of predicates or axiomatic definitions of predicate properties are not represented in this figure.

---

[i] We therefore consider hyponymy (hyperonymy) as a taxonomical relationship of specialization (generalization, equivalent of '*subclass of*' for terms).

We consider that a semantic graph doesn't rely on complex logic-based semantics, i.e., logical constructors (such as disjunction) are not required to understand the semantics associated to the KR. Specific mapping techniques can be used to reduce any KR to a semantic graph. In addition, the knowledge defined in highly formal KR might not be explicit and a reasoner can be required to deduce implicit knowledge, e.g. applying entailment rules or inference mechanisms, prior to applying mapping techniques. We will not broad this technical subject in this section, detailed information relative to the construction of a semantic graph from several types of KR is presented in section 5.2.

A semantic graph is therefore considered as any declarative KR which can be expressed through a graph and which carries a specific semantics, e.g., a semantic network/net, a conceptual graph, a lexical database (WordNet), an RDF(S) graph[i], a lightweight ontology, to mention a few.

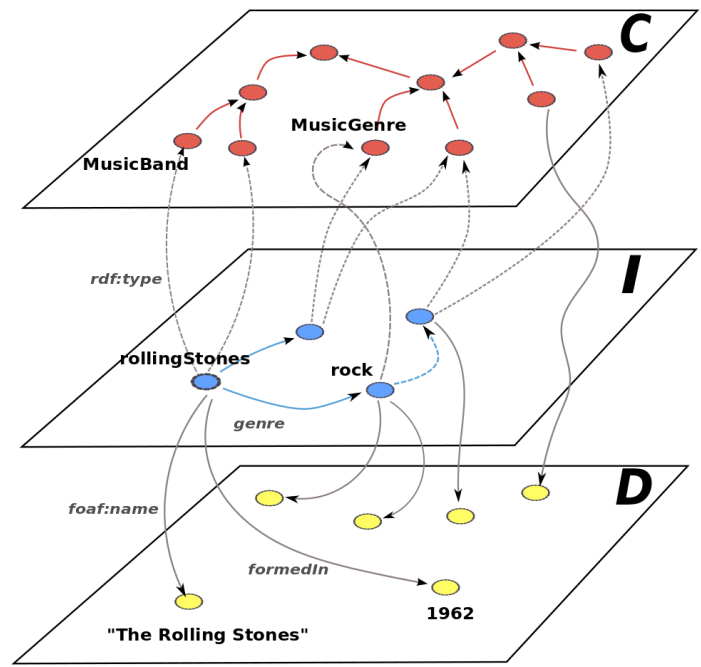

**Figure 4**: Example of a semantic graph involving classes, instances and data values (Harispe, Ranwez, et al. 2013a).

---

[i] RDF graph are in some cases expected to be entailed, i.e. the semantics requires to be taken into account in order to materialize specific relationships implicitly defined in the graph, e.g. using RDFS. A discussion related to the subject is provided in appendix 4.

### 2.4.5 Conceptual annotations as a Semantic Graph

In some cases, a collection of instances is annotated by concepts structured in a KR. In those cases, the knowledge base is expected to be composed of a collection of annotations and a KR. To ease the reading, we consider that an instance which is characterized through a set of conceptual annotations can also be represented in a semantic graph, i.e., the instance can be represented by a node which establishes semantic relationships to the concepts associated to its annotations.

As an example, if a document is annotated to a specific concept/class, let's say the document `docA` is annotated to a set of concepts structured in a KR `{Physics, Gravitation}`, we can create the relationship `(docA, isAnnotatedBy, Physics)` and `(docA, isAnnotatedBy, Gravitation)` to model this knowledge through a unique KR.

### 2.4.6 Examples of Knowledge Representations Commonly Processed as Semantic Graphs

Several KR considered as semantic graphs, have been used to design and evaluate SMs. Among the most used we distinguish:

- *WordNet*[i] (Miller 1998; Fellbaum 2010) is widely used in natural language processing and computational linguistics. It models the lexical knowledge of native English speakers in a lexical database structured through a semantic network composed of synsets/concepts linked by semantic and lexical relations. In WordNet, concepts are associated to a specific meaning defined in a gloss. They are also characterized by a set of cognitive synonyms (synset) which can be composed of nouns, verbs, adjectives, adverbs. According to the official documentation, WordNet 3.0 is composed of 117 000 concepts (synsets), which are linked together by different types of semantic relationships, e.g., hyperonymy, hyponymy. SENSUS (Swartout et al. 1996) is another semantic graph derived from WordNet.

- *Cyc*[ii] (*OpenCyc*), an ontology defining concepts (called constants), instances, and relationships between the concepts and instances. Other constructs are also provided

- *The Gene Ontology*[iii] and *gene product annotations*. The Gene Ontology (GO) defines a structured vocabulary related to molecular biology. It can be used to characterize various aspects of gene products (molecular function, biological processes, and cellular component).

- *MeSH*[iv] – Medical Subject Headings, structured controlled vocabulary defining a hierarchy of biological and medical terms. The MeSH is provided by the U.S. National Library of Medicine and is used to index PubMed articles.

---

[i] http://wordnet.princeton.edu/
[ii] http://www.cyc.com/
[iii] http://www.geneontology.org/
[iv] http://www.ncbi.nlm.nih.gov/mesh

# 3 Classification of Semantic Measures

We have seen that various mathematical properties can be used to characterize technical aspects of SMs. This section distinguishes other general aspects which may be interesting to classify SMs. They will be used to introduce the large diversity of approaches proposed in the literature. We first present some of the general aspects of SMs which can be relevant for their classification. We next introduce two general classes of measures.

## 3.1 How to Classify Semantic Measures

The classification of SMs can be made according to several aspects; we propose to discuss four of them:
- The type of elements the measures aim to compare.
- The semantic proxies used to extract the semantics required by the measure.
- The semantic evidences and assumptions considered during the comparison.
- The canonical form adopted to represent an element and to handle it.

### 3.1.1 Types of Elements to Compare: Words, Concepts, Sentences…

SMs can be used to compare various types of elements:
- Units of language: words, sentences, paragraphs, documents.
- Concepts/Classes, groups of concepts.
- Instances semantically characterized.

We considered the notion of concept through a broad sense, i.e., class of instances which can be of any kind (abstract/concrete, elementary/composite, real/fictive) (Smith 2004). We also consider that a concept can also be represented through a synset, i.e., group of data elements considered semantically equivalent. A concept can therefore be represented as any set of words referring to the same notion, e.g., the terms dog, Canis lupus familiaris refer to the concept Dog. Note that we use both the notions of *concept* and *class* interchangeably. However, we will, as much as possible, favour the use of the term *class* as specifications used to express KRs generally refer to it, e.g. RDF(S)[i].

The notion of instance semantically characterized encompasses several situations in which an object is described through information from which semantic analyses can be performed. A semantic characterization could be the RDF description of the corresponding instance, a set of conceptual annotations associated to it, a set of tags, or even a subgraph of an ontology, to mention a few.

SMs can therefore be classified according to the type of elements they aim to compare.

---

[i] Note that in OWL a class is denoted a concept. The introduction of RDF (Resource Description Framework), RDFS (RDF-Schema) and OWL (Web Ontology Language) as languages for the expression of knowledge representations is considered out-of-the-scope of this paper. Please refer to the official documentation proposed by the W3C. The book of (Hitzler et al. 2011) proposes an easy-to-read comprehensive introduction to semantic web technologies.

### 3.1.2 Semantic Proxies from which Semantics is Distilled

A semantic proxy is considered as any source of information from which the semantics of the compared elements, which will be used by a SM, can be extracted. Two broad types of semantic proxies can be distinguished:

- Unstructured or semi-structured texts: Text corpora, controlled vocabularies, dictionaries.
- Structured: thesaurus, structured vocabularies, and ontologies.

### 3.1.3 Semantic Evidences and the Assumptions Considered

Depending on the semantic proxy used to support the comparison of elements, various types of semantic evidences can be considered. The nature of these evidences conditions the assumptions associated to the measure.

### 3.1.4 Canonical Forms Used to Represent Compared Elements

The canonical form, i.e. representation, chosen to process a specific element can also be used to distinguish the measures defined for the comparison of a specific type of elements. Since the representation adopted to process an element corresponds to a specific reduction of the element, the degree of granularity with which the element is represented may highly vary depending on it. The selected canonical form is of major importance since it influences the semantics associated to the score produced by a measure, that is to say, how a score must be understood. This particular aspect is essential when inferences must be driven from the scores produced by SMs.

A SM is defined to process a given type of element represented through a specific canonical form.

## 3.2 Landscape of Semantic Measures

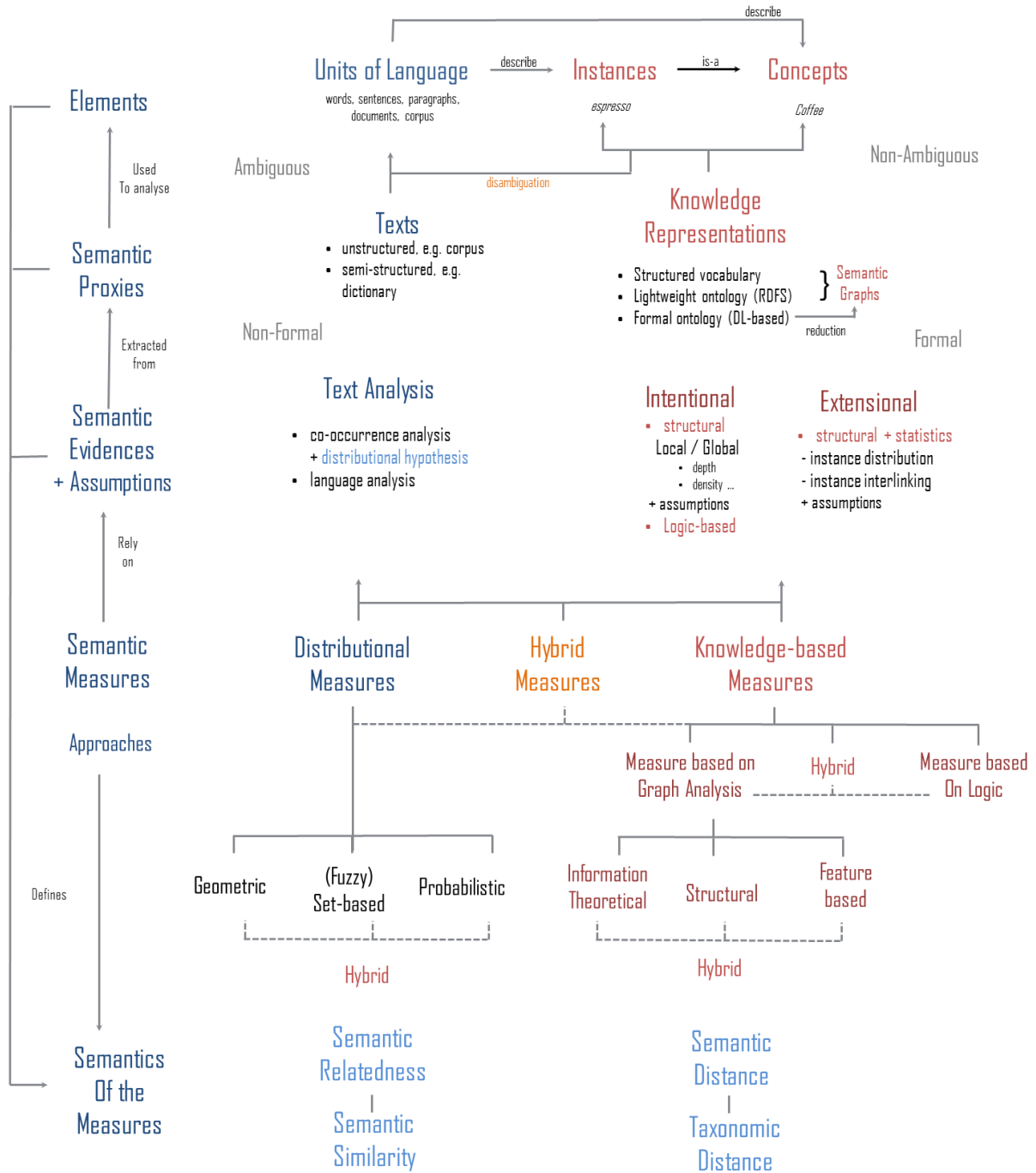

**Figure 5**: Partial Overview of the landscape of the types of semantic measures which can be used to compare various types of elements (words, concepts, instances …).

## 3.3 Distributional Measures

- *Semantic Proxy*: unstructured/semi-structured texts.
- *Type of elements compared*: units of language, i.e., words, sentences, paragraphs, documents.

Distributional measures enable the comparison of units of language through the analysis of unstructured texts. They are mainly used to compare words, sentences or even documents studying the repartition of words in texts (number of occurrences, location in texts)[i]. An introduction to this type of measures for the comparison of pair of words can be found in (Curran 2004; S. M. Mohammad & Hirst 2012).

Several contributions have been proposed to tackle the comparison of pairs of sentences or documents (text-to-text measures). Some of these measures derive from word-to-word SMs; others rely on specific strategies based on lexical/ngram overlap analysis, Latent Semantic Analysis extensions (Lintean et al. 2010), or even topic model using Latent Dirichlet Allocation (Blei et al. 2003). Text-to-text SMs are not presented in this section; we here focus on distributional measures which can be used to compare the semantics of two words regarding a collection of texts. Additional information and pointers regarding text-to-text measures are provided in appendix 2.

Distributional measures rely on the *distributional hypothesis* which considers that words occurring in similar *contexts* tend to be semantically close (Harris 1981). This hypothesis is one of the main tenets of statistical semantics. It was made popular through the idea of (Firth 1957): "*a word is characterized by the company it keeps*"[ii]. Considering that the context associated to a word can be characterized by the words surrounding it, the distributional hypothesis states that words occurring in similar contexts, i.e., often surrounded by the same words, are likely to be semantically similar as "*similar things are being said about both of them*" (S. M. Mohammad & Hirst 2012). It is therefore possible to build a distributional profile of a word according to the contexts in which it occurs.

A word is classically represented through the vector space model: a geometric metaphor of meaning in which a word is represented as a point in a multidimensional space modelling the diversity of the vocabulary in use (Sahlgren 2006). This model is used to characterize words through their distributional properties in a specific corpus of texts[iii]. To this end, words are generally represented through a matrix of co-occurrence – it can either be a word-word matrix or more generally a word-context matrix in which the context is any lexical unit (surrounding words, sentences, paragraphs or even documents). Such a characterization of a word regarding a specific corpus, sometimes denoted as *word space model* (Sahlgren 2006), is analogue to the vector space model widely known in Information Retrieval (Salton & McGill 1986).

Generally, the design of a SM for the comparison of words corresponds to the definition of a function which will assess the similarity of two context vectors. The various distributional measures are therefore mainly distinguished by the:

---

[i] In the literature, distributional measures are sometimes defined as a specific type of a more general type of measures denoted *corpus-based measures* (Panchenko & Morozova 2012). In this survey we consider distribution measures as any measure which relies on location and number of occurrences of words in text and there is therefore no need to distinguish them from corpus-based measures.

[ii] Also implicitly discussed in (Weaver 1955) originally written in 1949 (source: wiki of the Association for Computational Linguistics http://aclweb.org/aclwiki accessed 09/13)

[iii] According to (Sahlgren 2006), numerous limitations in the design of semantic measures to compare words are a consequence of the distributional methodology adopted as a discovery procedure.

- *Type of context* used to build the co-occurrence matrix.
- *Frequency weighting* (optional). The function used to transform the raw counts associated to each context in order to incorporate frequency and informativeness of the context (Curran 2004).
- *Dimension reduction technique* (optional) used to reduce the co-occurrence matrix. This aspect defines the type of co-occurrences which is taken into account (e.g. first order, second order, etc.).
- *Vector measure* used to assess the similarity/distance of the vectors which represent the words in the co-occurrence matrix. In some cases, vectors will be regarded as (fuzzy) sets.

Several distributional measures have been proposed. Here, we only briefly introduce the three main types of approaches; *Spatial/Geometric*, which evaluate the relative positions of the two words in the semantic space defined by the context vectors, and the *Set-based* and *Probabilistic* approaches which are based on the analysis of the overlap of the contexts in which the words occurs, e.g. (Ruge 1992).

### 3.3.1 Geometric Approach

The geometric approach is based on the assumption that compared elements are defined in a semantic space corresponding to the intuitive spatial model of similarity proposed by cognitive sciences (see section 2.2). A word is for instance considered as a point in a multi-dimensional space representing the diversity of the vocabulary in use. Two words are therefore compared regarding their location in this multi-dimensional space. The dimensions considered to represent the semantic space are defined by the context used to build the co-occurrence matrix. Words are represented through their corresponding vector in the matrix and are therefore compared through measures used to compare vectors. Among the most used, we can distinguish:

- Scalar product or measures from the $L_p$ Minkowsky family – $L_1$ Manhattan distance, $L_2$ Euclidian distance.
- Cosine similarity, the cosine of the angle between the vectors of the compared words (the smaller is the angle, the stronger the likeness will be considered). Measures of correlation can also be used in some cases (Ganesan et al. 2003).

### 3.3.2 (Fuzzy) Set-based Approach

Words are compared regarding the number of contexts in which they occur which are common and different (Curran 2004). The comparison can be made using classical set-based measures (e.g., Dice index, Jaccard coefficient). Several set-based operators have for instance been used to compare words (Terra & Clarke 2003; Bollegala 2007b). Extensions have also been proposed in order to take into account a weighting scheme through fuzzy sets, e.g. (Grefenstette 1994). Set-based measures relying on information theory metrics have also been proposed, they are introduced in the following subsection which presents the measures based on probabilistic approaches.

### 3.3.3 Probabilistic Approach

The distributional hypothesis enables to express the semantic relatedness of words in term of probability of co-occurrence, i.e. regarding both, the contexts in which compared words appear and the contexts in which the two words co-occur. These two evidences can intuitively be used to estimate the strength of association between two words. This strength of association can also be seen as the mutual information of two words. The mutual information can be expressed regarding the probability the two words occur in the corpus, as well as the probability the two words co-occur in the same context. Once again a large diversity of measures have been proposed, only those which are frequently used are considered (Dagan et al. 1999; S. Mohammad & Hirst 2012):

- *Maximum likelihood estimate* (MLE).
- *Pointwise Mutual Information* (PMI) (Fano 1961). The PMI was first adapted for the comparison of words by (Church & Hanks 1990). It is based on the analysis of the number of co-occurrences and individual occurrences of words, e.g., in paragraphs or sentences – examples of use are discussed in (Lemaire & Denhière 2008; S. M. Mohammad & Hirst 2012). To overcome the fact that PMI is biased towards infrequent words, various adaptations and correction factors have been proposed (Pantel & Lin 2002; S. Mohammad & Hirst 2012). Other related measures are Mutual Information and the Lexicographers Mutual Information, to mention a few.
- *Confusion probability.*

The vectors obtained from the co-occurrence matrix can also be seen as distribution functions corresponding to distribution profiles. Notice that these vectors can also correspond to vectors of strength of association if one of the metrics presented above have been used to convert the initial co-occurrence matrix. As an example, co-occurrence vectors can be transformed to mutual information vectors modifying the co-occurrence matrix using the PMI function. In both cases, the comparison relies on the comparison of two distribution functions/vectors. Therefore, despite their conceptual differences, the probabilistic approaches generally rely on the mathematical tools used by the geometrical approaches. The functions commonly used are:

- *Kullback-Leibler divergence* (information gain or relative entropy) is a classic measure used to compare two probability distributions and is often characterized as the loss of information when a probability distribution is approximated by another (Cover & Thomas 2012).
- *Jensen-Shannon divergence*. This function also measures the similarity between two probability distributions. It is based on the Kullback-Leibler divergence with the interesting properties of being symmetric and to always produce a finite value.
- *Skew Divergence* (Lee n.d.), denoted $\alpha$ SD ($\alpha$ stands for asymmetrical).
- Measures presented in the geometric approaches can also be used: L-Norm, cosine similarity.

An excerpt of the similarity functions which can be used to compare probability distributions can be found in (Pantel & Lin 2002); a comprehensive survey presenting a large collection of measures is also proposed in (Cha 2007)[i].

Several combinations can therefore be used to mix both the strength of association (weighting scheme, e.g., PMI) and the measures used to compare the probability functions/vectors. Fuzzy Metrics can also be used to compare words according to their strength of association, refer to (S. Mohammad & Hirst 2012) for detailed examples.

---

[i] An interesting correlation analysis between measures is also provided.

### 3.3.4 Capturing Deeper Co-occurrences

The probabilistic approach presented so far can be used to estimate the similarity of words regarding their *first order co-occurrence*, i.e., the similarity is assessed regarding if the two words occur in the same context. However, a strong limitation of first order co-occurrence studies is that *similar* words may not co-occur. As an example, some studies of large corpus have observed that the words *road* and *street* almost never co-occurs, although they can be considered as synonyms in most cases (Lemaire & Denhière 2008). Specific techniques have therefore been proposed to highlight deeper relationships between words, e.g., second order co-occurrences. These techniques will transform the co-occurrence matrix to enable evidences of deeper co-occurrence to be captured. The measures presented to compare words through the vector space model (see above) will be used after the matrix transformation.

Statistical analysis can be used to distinguish valuable patterns in order to highlight deeper co-occurrences between words. These patterns, which represent the relationships between words, can be identified using several techniques; among them we distinguish:

- Latent semantic analysis (LSA).
- Hyperspace Analogue to Language (HLA).
- Syntax or dependency-based model.
- Random indexing…

Latent Semantic Analysis/Indexing (LSA) use singular value decomposition (SVD) (Landauer et al. 1998) to capture the relationships between two words occurring with a third one (but not necessarily occurring together). To this end, the co-occurrence matrix is reduced by the SVD algorithm. SVD is a linear algebra operation which can be used to emphasize correlations among rows and columns – it reduces the number of dimensions with the interesting property to highlight second-order co-occurrences. The comparison of words is finally made comparing their corresponding rows in the matrix through vector similarity measures. LSA is often presented as an answer to the drawbacks of the standard vector space model such as sparseness and high dimensionality.

Hyperspace Analogue to Language (Lund & Burgess 1996) uses a word-to-word co-occurrence matrix according to a specific word window. The weight of a co-occurrence is defined according to the position of the two words (before/after) in the context window. An asymmetric co-occurrence matrix is therefore obtained (directional word-to-word matrix). Low entropy columns can therefore be dropped from the matrix prior to the comparison. Words are generally compared based on the concatenation of their respective row and column.

### 3.3.5 Advantages and Limits of Distributional Measures

We list some of the advantages and limits of distributional SMs.

#### *3.2.2.1. Advantages of Distributional Measures*

- Unsupervised, they can be used to compare the relatedness of words expressed in corpora without prior knowledge regarding their meaning or usage.

### 3.2.2.2. Limits of Distributional Measures

- The words to compare must occur at least few times.
- Highly depend on the corpus used. This specific point can also be considered as an advantage as the measure is context-dependent.
- Sense-tagged corpora are most of the time not available (Resnik 1999; Sánchez & Batet 2011). The construction of a representative corpus of text can be challenging in some usage context, e.g., biomedical studies.
- Difficulties to estimate the relatedness between concepts or instances due to the disambiguation process required prior to the comparison. Distributional measures are mainly designed for the comparison of words. However, some pre-processing and disambiguation techniques can be used to enable concepts or instances comparison from text analysis. Nevertheless, their computational complexity is most of the time a drawback making such approaches impracticable to be used with large corpora analysis.
- Difficulty to estimate the semantic similarity. Nevertheless, different observations are provided in the literature. It is commonly said that distributional measure can only be used to compare words regarding relatedness as co-occurrence can only be seen as an evidence of relatedness, e.g., (Batet 2011a). However, (S. Mohammad & Hirst 2012) specifies that similarity can be captured performing specific pre-processing. Nevertheless, capturing the similarity between words from text analysis requires elaborated techniques which are not tractable for large corpora analysis.
- Difficulty to explain and to trace the semantics of the relatedness. The interpretation of the score is almost driven by the distributional hypothesis; it is however difficult to deeply understand the semantics associated to the co-occurrences.

## 3.4 Knowledge-based Measures

- *Semantic Proxy*: knowledge representations, e.g., thesaurus, taxonomies, ontologies.
- *Type of elements compared*: words/terms, concepts, groups of concepts, instances semantically characterized.

Knowledge-based measures rely on any form of knowledge representation (KR), e.g., structured vocabularies, semantic graphs or ontologies, from which the semantics associated to the compared elements will be extracted. These measures are therefore based on the analysis of semantic graphs or expressive KRs defined using logic-based semantics.

A large diversity of measures have been defined to compare both concepts and instances defined in a KR. Two main types of measures can be distinguished considering the type of the KR which is taken into account:

- *Measures based on graph analysis* or framed in the relational setting. They consider KR as semantic graphs. They rely on the analysis of the structural properties of the semantic graph. Elements are compared studying their interconnections, in some case, by explicitly taking advantage of the semantics carried by the relationships.

• *Measures relying on logic-based semantics,* such as description logics. These measures use a higher degree of semantic expressivity; they can take into account logical constructors, and can be used to compare richer descriptions of knowledge.

Most of SMs have been defined to compare elements defined in a single KR some SMs have also been proposed to compare elements defined in different KRs. In this section, we mainly consider the measures defined for a single KR. Measures taking advantage of multiple KRs are briefly presented in section 3.4.3.

### 3.4.1 Semantic Measures Based on Graph Analysis

 SMs based on graph analysis do not take into account logical constructors which can sometimes be used to define the semantics of a KR. These measures only consider the semantics carried by the semantic relationships (relational setting), e.g., specific treatments can be performed regarding the type of relationship processed. Some properties associated to the relationships defined in the graph can be considered by the measures. The transitivity of the taxonomic relationship will for instance be implicitly or explicitly used in the design of the measures. In other cases, the taxonomy of predicates (the types of semantic relationships) can also be taken into account.

A large number of approaches have been proposed to express SMs using this approach. Section 5 is dedicated to them. We invite the reader willing to have more technical information to refer to it; here, we only present a non-technical overview of these measures focusing on those used to compare a pair of classes.

SMs based on graph analysis are commonly classified in four approaches: (i) Structure-based, (ii) Feature-based, (iii) Information-Theory and (iv) Hybrid.

#### 3.4.1.1 The Structural Approach

SMs based on the structural approach compare the elements defined in the graph through the study of the structure of the graph induced by its relationships. The measures are generally expressed as a function of the strength of the interconnections of the compared elements in the semantic graph. The structural approach corresponds, in some sense, to the design of SMs according to the structural model defined in cognitive sciences (refer to section 2.2). The graph corresponds to a structured space in which the compared elements are described.

The first measures based on the structural approach proposed to compare two classes regarding the shortest path linking them in the graph: the shorter the path, stronger the strength of their semantic relatedness. The types of relationships considered define the semantics of the measures, e.g., only the taxonomical relationships will be considered to estimate the semantic similarity. The similarity/relatedness of the compared elements is therefore estimated according to their distance in the graph.

As an example, considering **Figure 6**, the length of the shortest path between the classes `Computer` and `Tablet` is 2. Only considering taxonomical relationship, the length of the shortest path between the classes `Computer` and `Mouse` is 5. As expected, the classes `Computer` will therefore be considered to be more similar to the class `Tablet` than to the class `Mouse`.

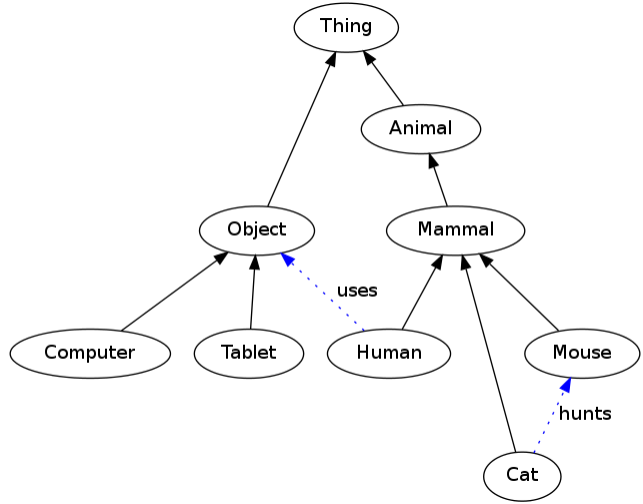

**Figure 6**: Example of simple semantic graph

A large diversity of structural measures have been proposed to compare elements structured in a graph as a function of the strength of the interconnection linking them (e.g., random-walk approaches). More refined measures take advantage of intrinsic factors analyses to better estimate the similarity, e.g., by considering non-uniform weights of relationships.

### 3.4.1.2 The Feature-based Approach

This approach can be associated to the feature model defined by Tversky. Measures estimate the semantic similarity or relatedness considering specific properties of the elements during the comparison.

A central element of the measures based on this approach is the function which characterizes the features of the elements on which will be based their comparison. Various strategies have been proposed. As an example, the features used to characterize a concept can be the senses it encompasses, its ancestors in the graph, i.e., all concepts which subsumes the concept according to the partial ordering defined by the taxonomy. Adopting this strategy we will consider the following feature-based representation of concepts `Computer`, `Tablet` and `Mouse`:

- `Computer = { Computer , Object, Thing}`
- `Tablet = { Tablet, Object, Thing}`
- `Mouse = { Mouse, Mammal, Animal, Thing}`

The comparison of two elements can therefore be made evaluating the number of features they share according to a feature-matching function. In this case, the pair `Computer / Tablet` will be estimated as more similar than the pair `Computer / Mouse` as the former pair share more senses than the latter (respectively 2 and 1).

### 3.4.1.1 The Information Theoretical Approach

This approach is based on Shannon's Information Theory (Shannon 1948). The elements, generally classes, are compared according to the amount of information they share and to the amount of information which is distinct between the two classes. These measures extensively rely on estimators of the information content of classes.

### 3.4.1.2 The Hybrid Approach

*Hybrid measures* are defined taking into account some of the specificities of the approaches briefly introduced above.

## 3.4.2 Measures Relying on Logic-based Semantics

SMs based on the relational setting cannot be used to compare complex descriptions of classes or instances, relying on logic-based semantics, e.g. description logics (DLs). To this end, SMs capable of taking into account logic-based semantics have been proposed. These measures are for example used to compare knowledge models expressed in OWL[i].

Among the diversity of proposals, measures based on simple description logics, e.g., only allowing concept conjunction (logic *A*), have initially been proposed (Borgida et al. 2005). More refined SMs have also been designed to exploit high expressiveness of DLs, e.g. ALC, ALN, SHI description logics (D'Amato et al. 2005a; Fanizzi & D'Amato 2006; Janowicz 2006; Hall 2006; Araújo & Pinto 2007; D'Amato et al. 2008a; D'Amato et al. 2009; Janowicz & Wilkes 2009; Stuckenschmidt 2009b). These measures rely, most of the time, on extensions of the feature model proposed by Tversky.

The thesis of (D'Amato 2007) proposes an in-depth presentation of the measures based on description logics.

## 3.4.3 Semantic Measures based on Multiple Knowledge Representations

Several approaches have been designed to estimate the relatedness of classes or instances using multiple KRs. These approaches are generally named cross ontology semantic similarity/relatedness measures in the literature. Their aim is double: (i) to enable the comparison of elements which have not been defined in the same KR given that the KRs in which they have been defined model a subset of equivalent elements, (ii) to refine the comparison of elements incorporating a larger amount of information during the process.

These measures are in some senses related to those commonly used for the task of ontology alignment/mapping and instance matching. Therefore, prior to their introduction we will first highlight the relation between these measures and those designed for the aforementioned processes.

---

[i] The book of (Hitzler et al. 2011) proposes a comprehensive introduction to OWL and Description Logics. (D'Amato 2007) also propose a digested introduction to description logic.

### 3.4.3.1 Comparison with Ontology Alignment/Mapping and Instance Matching

The task of ontology mapping aims at finding correspondences between the classes and predicates defined in a collection of KRs (e.g., ontologies). These mappings are further used to build an alignment between KRs. Instance Matching focuses in finding similar instances defined in a collection of KRs. These approaches generally rely on multiple matchers which will be aggregated to evaluate the similarity of the compared elements (Shvaiko & Euzenat 2013). The matchers commonly distinguished are:

- Terminological – based on string comparison of the labels or definitions of the elements.
- Structural – mainly based on the structuration of classes and predicates.
- Extensional – based on instance analysis.
- Logic-based – rely on logical constructs used to define the elements of the KRs.

The score produced by these matchers are generally aggregated. A threshold is next used to estimate if two (groups of) elements are similar enough to define a mapping between them. In some cases, the mapping will be defined between an element and a set of elements, e.g., depending on the difference of granularity of the compared KRs, a class can be mapped to a set of classes.

The problematic of ontology alignment and instance mapping is a field of study in itself. The techniques used for this purpose involve semantic similarity measures for the design of specific matchers – structural, extensional and logic-based (terminological matchers are not semantic). However, their aim being to find exact matches, they are generally not suited for the comparison of non-equivalent elements defined in different KRs. Indeed, techniques used for ontology alignment are for instance not suited to answer questions such as "To which degree the two concepts `Coffee` and `Cup` are related?".

Technically speaking, nothing prevents the use of matching techniques to estimate the similarity or relatedness of elements defined in different KRs since they have been designed to this specific purpose. Nevertheless, in applications, compared to approaches used for ontology alignment and instance matching, semantic measures based on multiple KRs:

- Can be used to estimate the semantic relatedness and not only the similarity.
- Sometimes rely on strong assumptions and approximations which cannot be considered to derive alignments, e.g., measures based on shortest-path techniques.
- Focus on the design of techniques for the comparison of elements defined in different KRs which generally consider a set of existing mappings between KRs.

In short, ontology alignment and instance matching are complex processes which use specific types of (semantic) similarity measures and which can be used to support the design of semantic measures involving multiple KRs.

### 3.4.3.2 Main approaches for the definition of semantic measures using multiple KRs

The design of semantic measures for the comparison of elements defined in different KRs have gained less attention than classical semantic measures designed for single KRs. They have been successfully used to support data integration (Rodríguez & Egenhofer 2003; M.C. Lange, D.G. Lemay 2007), clustering (Batet, Valls, et al. 2010), or information retrieval tasks (Xiao & Cruz 2005) to cite a few. In this context, several contributions have focused in the design of cross-KRs semantic measures without focusing on a specific application context (Rodríguez & Egenhofer 2003; Xiao & Cruz 2005; Petrakis et al. 2006; M.C.

Lange, D.G. Lemay 2007; Al-mubaid & Nguyen 2009; Coates et al. 2010; Batet, Valls, et al. 2010; Saruladha & Aghila 2011; Sánchez, Solé-Ribalta, et al. 2012; Batet et al. 2013).

The measures proposed in the literature can be distinguished according to the approach they adopt:

- Structural approach (Al-mubaid & Nguyen 2009).
- Feature-based approach (Petrakis et al. 2006; Batet, Valls, et al. 2010; Sánchez, Solé-Ribalta, et al. 2012; Batet et al. 2013; Solé-Ribalta et al. 2013)
- Information Theoretical approach (Saruladha, Aghila & Bhuvaneswary 2010; Saruladha 2011; Sánchez & Batet 2013).
- Hybrid approach (Rodríguez & Egenhofer 2003).

### 3.4.4 Advantages and Limits of Knowledge-based Measures

#### *3.3.2.1. Advantages of Knowledge-based Measures*

- They can be used to compare all types of elements defined in a KR, i.e., terms/classes, instances. Thus, these measures can be used to compare elements which cannot be compared using text analysis, e.g., comparison of gene products according to conceptual annotations corresponding to their molecular functions.
- Fine control on the semantic relationships taken into account to compare the elements. This aspect is important to understand the semantics associated to a score of SMs, e.g., semantic similarity/relatedness.
- Generally easier and less complex to compute than distributional measures.

#### *3.3.2.2. Limits of Knowledge-based Measures*

- Require a knowledge representation describing the element to compare.
- The use of logic-based measures can be challenging to compare elements defined in large knowledge bases (high computational complexity).
- Measures based on graph analysis most of the time require the knowledge to be modelled in a specific manner in the graph and are not designed to take into account of non-binary relationships. Such relationships are used in specific KRs and play an important role to define specific properties to relationships, e.g., a simple triplet cannot be used to model that a user has sent an email to another user at a specific date. Reification techniques are used to express such knowledge by defining a ternary relationship. The (binary) relationship is expressed by a node of the graph; specific triplets will be used to represent the sender (subject), the receiver (object), the type of relationship (predicate) and additional information associated to the binary relationship, such as the date in the given example. Despite the fact that some supervised approaches can be used to take advantage of such a type of knowledge expression, e.g. (Harispe, Ranwez, et al. 2013a), most measures based on graph analysis are not adapted to this case. This aspect is relative to the mapping of a KR to a semantic graph; a more detailed discussion of this specific aspect is proposed in appendix 4.

## 3.5 Mixing Knowledge-based and Distributional Approaches

Hybrid measures have been proposed to take advantage of both texts and KRs to estimate the similarity or relatedness of elements: units of language, concepts/classes and instances. They most of the time combine several single SMs (Panchenko & Morozova 2012).

Among the various mixing strategies, we can distinguish two broad types:

- Measures which take advantage of both corpus and KR analysis. This strategy has been used to estimate the specificity of concepts or terms structured in a (taxonomic) graph. As an example, (Resnik 1995) proposed to estimate the amount of information carried by a concept as the inverse of the probability the concept occurs in texts. Others propose to mix text analysis and structure-based measures. The extended gloss overlap measure introduced by (Banerjee & Pedersen 2002), and the two measures based on context vectors proposed by (Patwardhan 2003) are good examples. Interested readers may also consider (Patwardhan et al. 2003; Banerjee & Pedersen 2003; Patwardhan & Pedersen 2006).

- Simple aggregation of multiple measures. Considering several measures, the scores are aggregated according to the average, min, max, median or any aggregation function which can be designed to aggregate matrix of scores[i].

Several studies have demonstrated the gain of performance mixing knowledge-based and distributional approaches (Panchenko & Morozova 2012). See also the work of (Petrakis et al. 2006).

---

[i] Several aggregations have been proposed to compare groups of concepts. Please refer to section 5.6.2.2 for more information relative to aggregation functions.

# 4 Computation and Evaluation of Semantic Measures

This section introduces information relative to the comparison and the evaluation of SMs. Software solutions which can be used for the computation and analysis of SMs are first presented. In a second step, protocols, methodologies and benchmarks commonly used to assess and compare performances and accuracies of measures are presented.

## 4.1 Software Solutions for the Computation of Semantic Measures

This subsection presents the main software solutions dedicated to SM computation which are available to date (2013); new software solutions or more recent versions of those presented herein may be available. Please notice also the potential conflict of interest as the authors are involved in the Semantic Measures Library project (Harispe, Ranwez, et al. 2013b) which is related to the development of a software solution dedicated to SMs. Tools are presented in their alphabetical order.

### 4.1.1 Software Solutions Dedicated to Distributional Measures

List of existing software solutions for the computation of distributional measures:

- **_Disco_**: Java library dedicated to the computation of the semantic similarity between arbitrary words. A command-line tool is also available. Reference: (Kolb 2008), License: Apache 2.0, Last version: 2013, Website: http://www.linguatools.de/disco/

- **_Semilar_**: a toolkit and software environment dedicated to distributional SMs. It can be used to estimate the similarity between texts. It implements simple lexical overlap method for the comparison of texts, as well as word-to-word based measures. More sophisticated methods based on LSA/LDA are also provided. Semilar is available as an API and as a Java application with a graphical user interface. It also provides a framework for the systematic comparison of various SMs. Reference: (Rus et al. 2013), License: Unknown, Last Version: 2013.
  Website: http://www.cs.memphis.edu/~vrus/SEMILAR/

- **_SemSim_**: Java library dedicated to the computation of semantic similarity between words. Last Version: 2013, Website: http://www.marekrei.com/projects/semsim/

- Other tools:
  - Semantic similarity of sentences: http://sourceforge.net/projects/semantics/

### 4.1.2 Software Solutions Dedicated to Knowledge-based Measures

This subsection presents a list of existing software solutions for the computation of knowledge-based measures.

#### 4.1.2.1 Generic software solutions

*Generic software solutions* can be used to compare concepts/classes, groups of concepts or instances according to their descriptions in KRs. These software solutions can be used with a large variety of KRs (i.e. thesaurus, ontologies, and semantic graphs). The proposed solutions are listed alphabetically.

- *OWL Sim*: Java library dedicated to the comparison of classes defined in OWL. Reference: (Washington et al. 2009), License: Not specified, Last version: 2013.
  Website: http://code.google.com/p/owltools/wiki/OwlSim

- *The Semantic Measures Library* (SML): Java library and command line software dedicated to the computation of knowledge-based semantic similarity/relatedness. It can be used to compare a pair of concepts/classes, groups of concepts and instances defined in a semantic graph. The SML can be used as both, a Java API and command line toolkit. Various types of formats and languages used to express KRs are supported, e.g., RDF, OBO, OWL. Several Domain specific formats and semantic graphs are also supported, e.g., to load the Gene Ontology, WordNet or biomedical semantic graphs expressed using specific formats, such as MeSH or SNOMED-CT.
  Reference: (Harispe, Ranwez, et al. 2013b), Licence: Cecill (GPL compatible), Last version 2013.
  Website: http://www.semantic-measures-library.org

- *SemMF*: a Java library dedicated to the comparison of objects represented through RDF graphs. Reference: (Oldakowski & Bizer 2005). Licence: LGPL, Last version: 2008.
  Website: http://semmf.ag-nbi.de/

- *Similarity Library*: Java library which can be used to compute the semantic similarity of pairs of concepts defined in an ontology. Strictly speaking, the library is not generic but it can be used to exploit KRs such as WordNet, MeSH or the Gene Ontology. Reference: (Pirró & Euzenat 2010a), License: None – Download under request, Last version: unknown.
  Website: http://simlibrary.wordpress.com/

- *SimPack*: Java library which can be used to compute similarity between concepts defined a KR. SimPack has been used to include inexact search to SPARQL. Reference: (Bernstein et al. 2005). Licence: LGPL, Last version: 2008.
  Website: https://files.ifi.uzh.ch/ddis/oldweb/ddis/research/simpack/

- **SSonde**: Java framework and command-line toolkit used to estimate the similarity of instances defined in RDF graphs. Reference: (Albertoni & Martino 2012), Licence: GPL v3, Last version: 2012.Website: https://code.google.com/p/ssonde/

#### 4.1.2.2 WordNet

Software solutions dedicated to the comparison of words or synsets using WordNet.

- **Generic software solutions**: Software solutions such as the Semantic Measures Library or the Similarity Library support WordNet. See above for more information.

- **JWNL** and **JAWS**: Java libraries for the computation of word similarity. License: JWNL - BSD, JAWS - none, Last version: JWNL 2009 – JAWS 2009.
  JWNL: http://sourceforge.net/apps/mediawiki/jwordnet/
  JAWS: http://sourceforge.net/projects/jwordnet

- **WordNet::Similarity**: Perl module which implements a variety of SMs (e.g., Resnik, Lin, Jiang and Conrath). Reference: (Pedersen et al. 2004), License: GPL v2, Last version: 2008, Website: http://wn-similarity.sourceforge.net/

- **WS4J**: Java library for the computation of similarity between synsets. An alternative Java implementation of the WordNet::Similarity can also be found at http://www.sussex.ac.uk/Users/drh21/ License: GNU GPL v2, Last version: 2013. Website: https://code.google.com/p/ws4j/

#### 4.1.2.3 Gene Ontology

List of the software solutions which can be used to compute semantic similarity/relatedness between Gene Ontology (GO) terms and gene products annotated by GO terms.

- **Generic software solutions**: The Semantic Measures Library and the Similarity Library support SM computation using the GO. They can be used to compare GO terms and gene product annotations. In addition, the SML-Toolkit, a command-line toolkit associated to the Semantic Measures Library, can also be used by non-developers to compute SMs using the GO. See above for more information.

- **FastSemSim**: Python library and command line toolkit which can be used to compare GO terms and gene products. License: GPL, Last version 2012.
  Website: http://sourceforge.net/p/fastsemsim

- **GOSemSim**: R library dedicated to semantic similarity computation using the GO. Reference: (Yu et al. 2010), License: GPL-2, Last version: 2012.
  Website: http://www.bioconductor.org/packages/2.11/bioc/html/GOSemSim.html

- **GOSim**: R library dedicated to semantic similarity computation using the GO. Reference: (Fröhlich et al. 2007), License: GPL, Last version: 2012.
  Website: http://www.dkfz.de/mga2/gosim/

Notice that several other software solutions have been developed for the Gene Ontology alone. Only the most important have been presented. Refer to (GO Tools 2013) for a more complete list.

*4.1.2.4  MeSH*

List of software solutions which can be used to compute semantic similarity/relatedness between MeSH descriptors.

- **Generic software solutions**: The Semantic Measures Library and the Similarity Library support SM computation using the MeSH. The Semantic Measures Library also supports the comparison of groups of MesH descriptors and can be used by non-developers through a command-line interface. See above for more information.

*4.1.2.5 Disease Ontology*

- **Generic software solutions**: The Semantic Measures Library can be used to compare pairs of (group of) terms defined in the Disease Ontology. See above for more information.

- **DOSim package**: R library dedicated to semantic similarity computation using the Disease Ontolgy. Reference: (J. Li et al. 2011), License: GPL-2, Last version: 2010.
  Website: http://210.46.85.150/platform/dosim/

*4.1.2.6 Unified Medical Language System (UMLS)*

- **Generic software solutions**: The Semantic Measures Library and the Similarity Library support SM computation using UMLS. See above for more information.

- **UMLS::Similarity**: Perl module dedicated to semantic similarity and relatedness computation using UMLS. Reference: (McInnes et al. 2009), License: unknown, Last version: 2013.
  Website: http://umls-similarity.sourceforge.net/

*4.1.2.7 Wikipedia*

- **Wp-semantic-similarity**: Java library for semantic similarity measures derived from Wikipedia. Last version: 2013, Website: https://github.com/shilad/wp-semantic-similarity

*4.1.2.8 Other Semantic Graphs*

Generic solutions can be used to handle semantic graphs for which no dedicated software solutions have been developed.

## 4.2 Evaluation of Semantic Measures

Evaluation protocols and benchmarks are essential to discuss the benefits and drawbacks of existing or newly proposed SMs. They are of major importance to objectively evaluate new contributions and to guide SMs' users in the selection of the best suited measures according to their needs. Nevertheless, despite the large literature related to the field, only few contributions focus on this specific topic.

Generally, any evaluation aims to distinguish the benefits and drawbacks of the compared alternatives according to specific criteria. Such comparisons are most of the time used to rank the *goodness* of measures regarding the selected criteria. Therefore, to be compared, three important questions deserve to be answered:

    (1) What are the criteria which can be used to compare SMs?
    (2) How to evaluate the goodness of a measure regarding a specific set of criteria?
    (3) Which criteria must be considered to evaluate measures for a specific application context?

### 4.2.1 Criteria for the Evaluation of Measures

Several criteria can be used to evaluate measures. Among them, we distinguish:

- The *accuracy and precision* of a measure.
- The *computational complexity*, i.e., algorithmic complexity.
- The *mathematical properties* of the measure.
- The *semantics* carried by the measure.

As we will discuss, these criteria can be used to discuss several aspects of measures.

#### 4.2.1.1 Accuracy and Precision

The accuracy of a measure can only be discussed according to predefined expectations regarding the results produced by the measure. Indeed, as defined in metrology, the science of measurement, the accuracy of a measurement must be understood as the closeness of the measurement of a quantity regarding the true value of that quantity (BIPM et al. 2008).

The precision of a measure (system of measurement) corresponds to the degree of reproducibility or repeatability of the score produced by the measure under unchanged conditions. Since most SMs are based on deterministic algorithms, i.e., they produce the same result given a specific input, here we focus on the notion of accuracy. We will further discuss the precision of a measure as a mathematical property.

The notion of accuracy of a measure is compulsory tight to a context, e.g., semantic proxy (specific corpus, KR, etc.), tuning of the parameters of the measure (if any). Indeed, there is no guaranty that a measure, which has been proved accurate in a specific context, will be accurate in all contexts. As we will see, SMs' accuracy is therefore evaluated according to expected results.

### 4.2.1.2 Computational Complexity

The computational complexity or algorithmic complexity of SMs is of major importance in most applications. Considering the growing volumes of datasets processed in semantic analysis (large corpus of texts and KRs), the algorithmic complexity of measures plays an important role for the adoption of SMs.

Considering equivalent accuracy in a specific context, most SMs' users will prefer to make concessions on accuracy for a significant reduction of computational time. However, the literature relative to SMs is very limited on this subject. It is therefore difficult to discuss algorithmic implications of current proposals, which hampers the non-empirical evaluation of measures and burden the selection of measures.

It is however difficult to blame SMs' designer for not providing detailed algorithmic analyses of their proposal. Indeed, computational complexity analyses of measures are both technical and difficult to make. In addition, most of the time, they depend on a specific type of data structure used to represent the semantic proxy taken into account by the measures, which sometimes create a gap between theoretical possibilities and practical implementations.

Despite its major importance, evaluation of SMs regarding their computational complexity is today difficult.

### 4.2.1.3 Mathematical Properties

Several properties of interest of measures have been distinguished in section 2.3, e.g., symmetry, identity of the indiscernibles, precision (for non-deterministic measures), and normalization. These mathematical properties are of particular importance for the selection of SMs. They are for instance essential to apply specific optimization techniques. They also play an important role to better understand the semantics carried by the measures, i.e., the meaning carried by the results produced by the measures.

Mathematical properties are central for the comparison of measures since they are, most of the time, required to ensure the coherency of treatments relying on SMs, this is for instance the case when inferences have to be made based on the scores produced by the measures. As an example, the implication of the lack of respect of the identity of the indiscernibles can be strong, It can be conceptually disturbing that comparing a class to itself can produces non-maximal or even low similarity scores, it is however the case using some measures in specific contexts[i].

As we will see, mathematical properties analyses are required to deeply understand measures and therefore evaluate their relevance for domain-specific application.

---

[i] As an example using Resnik's measure based on the notion of information content of concepts (introduced in section 5.5.3), the similarity of a general concept (near to the root) to itself will be low.

#### *4.2.1.4 Semantics*

The meaning (semantics) of SMs' results deserve to be understood by end-users of measures. This semantics is defined by the assumptions on which relies the algorithmic design of the measures. Some of these assumptions can be understood through the mathematical properties of the measures.

The semantics of a measure is also defined by the cognitive model on which relies the measure, the semantic proxy in use and the semantic evidences analysed. As we have seen in section 2.1.3, the semantic evidences taken into account by the measure define its type and its general semantics (e.g., the measure evaluates semantic similarity, relatedness…).

It is difficult to compare measures regarding the semantics they carry. It is however essential for SMs users to understand that measure selection may in some case strongly impact the conclusions which can be supported by the measurement (e.g. semantic similarity, relatedness, etc.).

### 4.2.2 Existing Protocols and Benchmarks for Accuracy Evaluation

Accuracy of SMs is considered as the *de-facto* metric to evaluate the performance of measures. SM's accuracy can be evaluated using a direct or an indirect approach. In most cases, measures are evaluated[i] using a direct approach, i.e., based on expected scores of measurement (e.g., similarity, relatedness) of pairs of elements. In all cases, evaluation of SMs' accuracy is performed regarding specific expectations/assumptions:

- *Direct evaluation*: based on the correlation of SMs with expected scores or other metrics. Measures are for instance evaluated regarding their capacity to mimic human rating of semantic similarity/relatedness. In this case, the accuracy of measures is discussed based on their correlations with gold standard benchmarks composed of pairs of terms/concepts associated to expected ratings. For domain-specific studies, a set of experts is used to assess the expected scores which will compose the benchmark (e.g., physicians in biomedical studies). The measure can also be evaluated regarding their capacity to produce scores highly correlated to metrics which summarize our knowledge of compared elements. As an example, in bioinformatics, evaluations of measures designed to compare gene products according to their conceptual annotations, is sometimes supported by studying their correlation with other measures aiming to compare genes (e.g., sequence similarity) (Lord 2003).

- *Indirect Evaluation*: The evaluation of the measures is based on the analysis of the performance of applications or algorithms which rely on SMs. The treatment considered is domain-specific dependant, e.g., accuracy of term disambiguation techniques, performance of a classifier or clustering relying on a SM to mention a few.

Thereafter, we present the benchmarks which can be used to compare SMs according to human ratings of similarity/relatedness. We next introduce other approaches which have been used to evaluate measures in specific domains.

---

[i] Please, in this subsection, understand evaluation as evaluation of accuracy.

### 4.2.2.1 Evaluation of measures based on human ratings

Benchmarks based on human-ratings are composed of pairs of elements for which humans have been asked to assign scores of similarity or relatedness. Existing benchmarks are mainly composed of pairs of terms. They have been built using a set of humans (experts). The distinction between the notions of similarity and relatedness has been introduced and the subjects have been trained prior to the experiment.

The measures are most of the time evaluated regarding their correlation (Pearson or Spearman) with averaged scores. In some cases, cleaning techniques are used to exclude abnormal ratings. In some cases, word-based benchmarks are conceptualized, i.e., words are mapped to concepts, in order to evaluate knowledge-based approaches[i], i.e., terms are manually mapped to concepts/synsets.

We distinguish the general benchmarks, dealing with common words, and domain-specific benchmarks involving a specific and technical terminology.

### 4.2.2.1.1 General Benchmarks

- (Rubenstein & Goodenough 1965) – semantic similarity - n=65

This benchmark is composed of 65 pairs of nouns (ordinary English words), e.g. *glass/magician*. Subjects were paid college undergraduates (n=51). They were asked to evaluate the (semantic) similarity using a 0-4 scale. The notion of semantic similarity was defined as the '*amount of similarity of meaning*' in the experiment.

The study focused on synonymy evaluation. The intra-subject reliability (n=15) on 36 pairs was r=0.85 using Pearson's correlation. Inter-subject correlation is not communicated but mean judgment of two different groups was impressively high (r=0.99).

- (Miller & Charles 1991) – semantic similarity

Subset of Rubenstein and Goodenough's benchmark composed of 30 pairs of nouns (n=38).

- *WordSim353* - (Finkelstein et al. 2001) – semantic relatedness

Two sets of English word pairs along with human rating of semantic relatedness. The first set contains 153 word pairs (n=13), which includes the 30 nouns pairs contained in (Miller & Charles 1991) bench-mark. The second set contains 200 pairs (n=16). Row results of the two sets are provided. A concatenation of the two sets with averaged scores is also given (which explains the name).

---

[i] In most cases, the concepts associated to the terms are not communicated in contributions. In some cases, words are mapped to multiple concepts and the best score is considered for the evaluation. Nevertheless, these particular cases are poorly documented in the literature.

- *Mturk-771* - (Halawi et al. 2012) – semantic relatedness

Benchmark composed of 771 pairs of English words along with their semantic relatedness (n>=20 for each pair). Inter-correlation of the results was assessed to r~0.89.

- (Ziegler et al. 2012) – semantic relatedness

Two sets of concept/instance pairs denoted by English words. The first contains 25 pairs, the second 30. The scores of relatedness were assessed based on an online survey. Inter-subject correlation based on Pearson's correlation is ~0.70 for the two sets.

### 4.2.2.1.2 *Domain specific Benchmarks*

- Biomedical domain

- (Pedersen et al. 2007) – semantic relatedness

  A set of 101 medical concepts rated for semantic relatedness. A subset of this set, composed of 29 pairs with higher inter-agreement, is generally considered.

- (Pakhomov et al. 2010) – semantic similarity and relatedness

  Two sets of UMLS concepts pairs. The first set contains 566 pairs of concepts and is dedicated to semantic similarity. The second set is composed of 587 pairs of concepts rated for semantic relatedness.

- Other benchmarks

The Semantic textual similarity campaign also proposes benchmark campaign for the comparison of texts (See http://ixa2.si.ehu.es/sts). Other datasets can be adapted to evaluate semantic similarity or relatedness, e.g. benchmarks used to evaluate distributional semantic models (Baroni & Lenci 2011).

### 4.2.2.2  Other domain-specific evaluations

#### 4.2.2.2.1 <u>Bioinformatics</u>

Please refer to (Pesquita, Faria, et al. 2009; Pesquita, Pessoa, et al. 2009; Guzzi et al. 2012).

# 5 Semantic Measures Based on Graph Analysis

This section is dedicated to a detailed and technical introduction to knowledge-based SMs which rely on graph analysis. These measures have been briefly presented in section 3.4.1, they are also denoted *graph-based SMs* or measures framed in the *relational setting* in the literature (D'Amato 2007). They differ from knowledge-based measures taking advantage of logic-based semantics analysis as they are designed to take advantage of KRs which do not rely on semantics based on logical constructors and formal grammar.

For convenience, SMs based on graph analysis will be denoted SMs in this section. They can be used to compare classes and instances and by extension groups of classes and instances. This section is structured as follows.

The first part discusses the importance of graph-based SMs and explains why they have gained a lot of attention in the last decades.

The second part extends the introduction to KR presented in section 2.4 and introduces a formal introduction of a semantic graph. The notations which will be considered to discuss technical aspects of the measures are also given (e.g., graph notations). In this part, we also discuss the construction of a semantic graph from other KRs. Specific treatments of semantic graphs which are required to ensure coherencies of (some) measures are also introduced.

The third part discusses the important notion of semantic evidence and present several metrics which can be used to extract semantics from a semantic graph. The pivotal notion of class specificity and strength of connotation, as well as strategies which have been proposed for their estimation are presented.

Part 4 presents the narrow link between the semantics associated to a measure (e.g., relatedness, similarity) and the information of the graph which is taken into account by the measures. Several approaches for the design of SMs are next introduced.

Part 5 to 8 are dedicated to the presentation of graph-based SMs which have been designed for the comparison of pair of classes:

- Part 5, estimation of the semantic similarity of pair of concepts.
- Part 6, estimation of the semantic similarity of comparison of groups of concepts.
- Part 7, discusses the unification of the measures defined for the estimation of the semantic similarity of (sets of) classes.
- Part 8, estimation of the semantic relatedness of two classes.

Finally, part 9, is dedicated to SMs which have been proposed for the estimation of the semantic relatedness of two instances.

## 5.1 Importance of Graph-based Semantic Measures

As we have seen, two main families of SMs can be distinguished: distributional measures, which take advantage of unstructured or semi-structured texts, and knowledge-based measures which rely on KRs.

Distributional measures are essential to compare units of languages such as words, or even concepts, when there is no formal expression of knowledge available to drive the comparison. As we stressed, these measures rely on algorithms governed by assumptions to capture the implicit semantics of the elements they compare (i.e., mainly the distributional hypothesis). On the contrary, knowledge-based SMs rely on formal expressions of knowledge explicitly defining how must be understood the compared elements. Thus, they are not constrained to the comparison of units of language and can be used to drive the comparison of any piece of knowledge formally describing a large diversity of elements, e.g., concepts, genes, person, music bands, etc.

We have underlined the limitation of knowledge-based measures, which mainly rely on their strong dependence on the availability of a KR - an expression of knowledge which can be difficult to obtain and may therefore not be available for all domain of studies. However, in the last decades, we have observed, both in numerous scientific communities and industrial fields, the growing adoption of knowledge-enhanced approaches based on KRs. As an example the Open Biological and Biomedical Ontology (OBO) foundry gives access to hundreds of KRs related to biology and biomedicine. Therefore, thanks to the large efforts made to standardize the technology stack which can be used to define and take advantage of KRs (e.g., RDF(S), OWL, SPARQL - triple stores implementations) and thanks to the increasing adoption of the Linked Data and Semantic Web paradigms, a large number of experts and initiatives give access to KRs in numerous domains (e.g., biology, geography, cooking, sports).

Even large corporations adopt KRs to support their large-scale worldwide systems. The most significant example of the recent years is the adoption of the *Knowledge Graph* by Google, a graph built from a large collection of billions of non-ambiguous subject-predicate-object statements used to formally describe general or domain-specific pieces of knowledge. This KR is used to enhance their search engine capabilities and millions of users benefit from it daily. Several examples of such large KRs are today available, some of them for free: DBpedia, Freebase[i], Wikidata, Yago.

Another significant fact of the increasing adoption of KRs is the joint effort made by the major search engines companies, e.g., Bing (Microsoft), Google, Yahoo! and Yandex[ii], to design Schema.org, a set of structured schemas defining a vocabulary which can be used to characterize in an unambiguous manner the content of web pages.

An interesting aspect of the last years is also the growing adoption of graph databases (e.g., Neo4J, OrientDB, Titan). These databases rely on a graph structure to describe information in a NoSQL fashion. They actively contribute to the growing adoption of the graph property model to describe information (Robinson et al. 2013).

In this context, a lot of attention has been given to KRs, which in numerous cases merely correspond to semantic graphs – characterized elements, concepts, classes, instances and relationships are defined in an unambiguous manner and the KR relies only on simple semantics expressions which do not take into account complex logical constructs. Such semantic graphs have the interesting properties to be easily expressed and maintained while ensuring a good ratio between semantic expressivity and effectiveness, for instance in term of computational complexity of the treatments which rely on them. This justifies the large

---

number of contributions related to the design of SMs dedicated to semantic graphs – a diversity of measures this section is dedicated to.

## 5.2 From Knowledge Representations to Semantic Graphs

For the sake of clarity, let's remind that we consider any machine understandable expression of knowledge through the generic term *knowledge representation* (KR), the language used to express this representation is called a KR language (e.g., RDF, OWL, OBO). Graph-based SMs can be used to take advantage of any KR which corresponds or can be embedded to a semantic graph.

In the literature, most knowledge-based SMs have been defined for a specific KR (language), e.g., WordNet, Gene Ontology, Linked Data represented as RDF graphs, ontologies taking advantage of specific description logic (languages). Since the aim of this section is to introduce the diversity of graph-based SMs, we must consider a generic formalism which can be used to refer to the expression of several KRs (e.g., RDFS graphs, OWL, simple taxonomy). Some of these KRs are not fully representable through a semantic graph[i]; we will therefore detail the general process explaining how complex models are generally reduced to semantic graph.

### 5.2.1 Formal Definitions

We first extend the formal model of a KR introduced in section 2.4 and we introduce several notations relative to semantic graphs.

#### 5.2.1.1 A Formal Model of a Knowledge Representation

A KR can be formally defined by $O: \{C, R, I, V, D, E, A^O\}$, with:

- $C$ the set of classes.
- $R$ the set of predicates.
- $I$ the set of instances.
- $V$ the set of data value
- $D$ the set of data types.

- $E$ the set of oriented relationships of a specific predicate $r \in R$:
  $E \subseteq E_{CC} \cup E_{RR} \cup E_{IC} \cup E_{CV} \cup E_{RV} \cup E_{IV}$ with:
  - $E_{CC} \subseteq C \times R \times C$
  - $E_{RR} \subseteq R \times R \times R$
  - $E_{IC} \subseteq I \times R \times I$
  - $E_{CV} \subseteq C \times R \times V$
  - $E_{RC} \subseteq R \times R \times V$
  - $E_{IV} \subseteq I \times R \times V$

- $A^O$ the set of axioms defining the interpretations of classes and predicates.

---

[i] Even if they rely on a graph-based language such as RDF, e.g., complex OWL ontologies can be expressed in RDF but are not semantic graphs.

The set of classes ($C$), predicates ($R$), instances ($I$), values ($V$) and data types ($D$) are expected to be mutually disjoint, e.g. $C \cap I = \emptyset$. Models enabling the use of data type to characterize specific attribute(s) (values) of classes and instances can also defined. These attributes can also be represented as semantic relationships (e.g., **Figure 4** page 29).

We consider that each instance is member of at least a class and that all classes are expected to be connected, e.g., subsumed by a general class denoted the *root*. The graph containing both the instances and the classes is therefore connected.

The membership of an instance $i$ to a class $X$ is asserted by a triplet (`i, isA, X`). Depending on the language used to express the KR the `isA` relationships may change, e.g., RDF(S) uses `rdf:type`.

We denote $I(X)$ the set of instances which are members of the class $X$ considering the transitivity of $\preccurlyeq_C$, with:

$$X \prec Y \implies I(X) \subseteq I(Y)$$

We also define $I^-(X)$ as the set of instances which are directly associated to a class, i.e., $I^-$ evaluates class membership without considering $\preccurlyeq_C$. $I^-(X)$ corresponds to the set of instances which are directly linked to a class by an `isA` relationship (considering that redundant relationships have been removed):

$$I^-(X) = \{i \mid \exists\, (i, isA, X)\}$$

As an example, consider the following set of statements:

```
Book subClassOf Document.
Magasine subClassOf Document.

book_1 isA Book .
book_2 isA Book .
magasine_1 isA Magasine .
letter_1 isA Document .
book_2 isA Document .
```

We obtain:

$I(Document) =$ { `book_1, book_2, magasine_1 ,letter_1` }
$I^-(Document) =$ { `letter_1` }
$I(Book) =$ { `book_1, book_2` }

Notice that, in some cases, annotated instances are (indirectly) characterized by classes without being members of them. For example, books can be annotated to specific topics corresponding to classes:

```
book_1 hasTitle "On the Origin of Species" .
book_1 hasAuthor charles_Darwin .
book_1 hasTopic EvolutionaryBiologyTopic .
EvolutionaryBiologyTopic subClassOf BiologyTopic
```

In this example, it cannot be considered that the instance `book_1` is an instance of the class `Evolutionary­BiologyTopic`. However, in numerous cases, *artificial instances* of classes will be considered by SMs, e.g., the set of instances characterizing `EvolutionaryBiologyTopic` will contain `book_1`. The relevance of the function $I$ is to be in accordance with the partial ordering of the classes, i.e., $X \prec Y \Rightarrow I(X) \subseteq I(Y)$.

Notice also that in other specific cases, a function $\mathcal{J}$ can be used to characterized instances. Indeed, under specific conditions, SM dedicated to the comparison of classes will be used to compare instances structured in a partial ordered set. As an example, let's consider that the topics used to annotate the books have been defined as instance of a class `Topic` and have next be structured in a poset through relationships of predicate `subTopicOf`.

Formally we can define the mapping $\mathcal{J}$ by:

$$\mathcal{J} : C \cup I \rightarrow P(I)$$

A KR $O$ with no axiomatic definition, $A^O = \emptyset$, is a graph with specific types of nodes. Therefore KRs which rely on simple axiomatic definitions of the properties of the predicates can be mapped to semantic graphs without loss of semantics. Some of these KRs are sometimes denoted *lightweight ontologies* in the literature. On the contrary, KRs based on complex axioms and constraints (generally called *heavyweight* ontologies) can only be partially represented by semantic graphs; the mapping to a semantic graph implies a reduction and a loss of knowledge initially expressed in the KR (Corcho 2003).

### 5.2.1.2 Semantic Graphs, Relationships and Paths

We further introduce the notations used to refer to particular constitutive elements of a semantic graph.

#### 5.2.1.2.1 <u>Relationships/ Statements /Triplets</u>

The relationships of a semantic graph are distinguished according to their predicate and to the pair of elements they link. The triplet $(u, t, v)$ corresponds to the unique relationship of type $t \in R$ which links the elements $u,v$. In the triplet $(u, t, v)$, $u$ is named the subject, $t$ the predicate and $v$ the object. Relationships are central elements of semantic graphs and will be used to define algorithms and to characterize paths graph.

Since the relationships are oriented, we denote $t^-$ the type of relationship carrying the inverse semantic of $t$. We therefore consider that any relationship $(u, t, v)$ implicitly implies $(v, t^-, u)$, even if the type of relationship $t^-$ and the relationship $(v, t^-, u)$ are not explicitly defined in the graph. As an example, the relationship (`Human, subClassOf, Mammal`) implies the inverse relationship (`Mammal, super­ClassOf, Human`), considering that $subClassOf^- \equiv superClassOf$. The notion of inverse type will be considered to discuss detailed paths. In some KR languages, inverse types are explicitly defined by specific construct, e.g., `owl:inverseOf`.

##### 5.2.1.2.2 _Graph traversals_

Graph traversals are often represented through path in a graph, i.e., sequence of relationships linking two nodes. To express such graph paths, we adopt the following notations[i].

> _Path_: Sequence of relationships $[(c_{i-1}, t_i, c_i), (c_i, t_{i+1}, c_{i+1}), \dots]$. To lighten the formalism, if a single predicate is used, the path is denoted $[c_{i-1}, c_i, c_{i+1}, \dots]^t$.

> _Path Pattern_: We denote $\pi = <t_1, \dots, t_i>$ with $t_i \in R$, a _path pattern_ which correspond to a list of predicates[ii]. Therefore, any path which is a sequence of relationships is an instance of a specific path pattern $\pi$.

We extend the use of the path pattern notation to express concise expressions of paths:

- $<t_*>$ corresponds to the set of paths of any length composed only of relationships having for predicate $t$.
- $<t_*^*>$ corresponds to the set of paths of any length composed of relationships associated to the predicate $t$ or $t^{-1}$. As an example, $\{Human, <subClassOf_*>, Animal\}$ refers to all paths linking the concepts $Human$ and $Animal$ which are only composed of relationships _subclass-of_ (and do not contain relationships of type $subClassOf^-$).

We also use mixing of the notations to characterize set of paths between specific elements. As an example, $\{u, <t, subClassOf_*>, v\}$ represents the set of paths linking the elements $u$ and $v$ which starts by a relationship of predicate $t$ and which finishes by a path (possibly empty) of _subclass-of_ relationships. As an example the class membership function $I$ which has been introduced above to characterize the instances of a specific class can formally be redefined by:

$$I(X) = \{i \mid \{i, <is-a, subClassOf_*>, X\} \neq \emptyset\}$$

Since the set of paths $\{u, <p_1>, v\}$ corresponds to a singleton $\{(u, p, v)\}$, or an empty set if the relationship $(u, p, v)$ doesn't exists in the graph, we consider that $\{u, <p_*>, v\}$ can be shorten by $\{u, p, v\}$.

#### 5.2.1.3 Notations for Ordered Sets and Taxonomies

A strict partial order is a binary relation $\prec$ over a set $C$ which is:

- Irreflexive $\forall c \in C : \neg (c \prec c)$
- Transitive $\forall u, v, w \in C : (u \prec v \land v \prec w) \Rightarrow u \prec w$

A non-strict total order is a binary relation $\leqslant$ over a set $C$ which is:

---

[i] This notation is based on an adaptation of the notation used by (Lao 2012)
[ii] In SPARQL 1.1, such paths are denoted using path properties $t_1/t_2/$ .

- Antisymmetric $\forall u, v \in C : (u \preccurlyeq v \land v \preccurlyeq u) \Rightarrow u = v$
- Transitive $\forall u, v, w \in C : (u \preccurlyeq v \land v \preccurlyeq w) \Rightarrow u \preccurlyeq w$
- Total $\forall u, v \in C : u \preccurlyeq v \lor v \preccurlyeq u$

A non-strict partial order is a binary relation $\preccurlyeq$ over a set $C$ which is:

- Reflexive $\forall c \in C : c \preccurlyeq c$
- Antisymmetric
- Transitive

The set $C$ with a partial order is named a partially ordered set (*poset*).

The taxonomy $G_T$ is the non-strict partial order defined by the taxonomical relationship over the set of classes $C$. Below, we introduce the notations used to characterize a taxonomy $G_T$, as well as its classes. Some of the notations have already been introduced and are repeated for clarity.

- $C(G_T)$, shorten by $C$ refers to the set of classes defined in $G_T$.

- $E(G_T)$, shorten by $E_T$ refers to the set of relationships defined in $G_T$ which link two classes, i.e. the set of relationships named $E_{CC}$ in the general introduction of a semantic graph.

- A class $v$ subsumes another class $u$ if $u \preccurlyeq v$, i.e., $\{u, <subClassOf_*>, v\} \neq \emptyset$ . We also say that the class $u$ is subsumed by $v$.

- $A(u)$ the set of classes which subsumes $u$, also named the ancestors of $u$ or its superclasses, i.e., $A(u) = \{c \mid \{u, <subClassOf_*>, c\} \neq \emptyset\} \cup \{u\}$. We also denote $A^-(u) = A(u)\backslash u$, the exclusive set of ancestors of $u$.

- $parent(u)$ the minimal subset of $A^-(u)$ from which $A^-(u)$ can be inferred according to the taxonomy $G_T$, i.e., if $G_T$ doesn't contain taxonomical redundancies, we obtain: $parent(u) = \{c \mid \exists (u, subClassOf, c)\}$.

- $D(u)$ the set of classes which are subsumed by $u$, also named the descendants of $u$, or its subclasses, i.e., $D(u) = \{c \mid \{c, <subClassOf_*>, u\} \neq \emptyset\} \cup \{u\}$.
  We also denote $D^-(u) = D(u)\backslash u$, the exclusive set of descendants of $u$.

- $children(u)$ the minimal subset of $D^-(u)$ from which $D^-(u)$ can be inferred according to the taxonomy $G_T$, i.e., if $G_T$ doesn't contain taxonomical redundancies, we obtain: $children(u) = \{c \mid \exists (c, subClassOf, u)\}$.

- $roots(G)$, shorten by $roots$, the set of classes $\{c \mid A(c) = \{c\}\}$. We call the $root$ the unique class, if any, which subsumes all classes, i.e., $\forall c \in C, c \preccurlyeq root$.

- $leaves(G)$, shorten by $leaves$, the set of classes without descendants, i.e.
  $leaves = \{c \mid D(c) = \{c\}\}$.

- $G_T^+(u)$ the graph composed of $A(u)$ and the set of relationship linking two classes in $A(u)$.

- $G_T^-(u)$ the graph composed of $D(u)$ and the set of relationship linking two classes in $D(u)$.

- $G_T(u) = G_T^+(u) \cup G_T^-(u)$ the graph induced by both the set of ancestors and descendants of $u$.

- $\Omega(u, v)$, the set of Non Comparable Common Ancestors (NOCA) of the classes $u, v$ and formally defined by $\forall (x, y) \in \Omega(u, v), x, y \in A(u) \cap A(v) \wedge x \notin A(y) \wedge y \notin A(x)$. The NCCA are also called the Disjoint Common Ancestors (DCAs) in some contributions.

Note that, despite the fact that we here use the taxonomy of classes and a specific semantics to the notations, they can be used to characterize any partially ordered set.

A taxonomical tree is a special case of $G_T$ in which:

$$\forall\, c \in C(G_T) : |parent(c)| < 2$$

### 5.2.2 Building a Semantic Graph from a Knowledge Representation

This section discusses the reduction of a KR to a semantic graph.

#### 5.2.2.1 The main steps

A generic process can be used to model the main steps which can be applied to obtain a semantic graph from any KR.

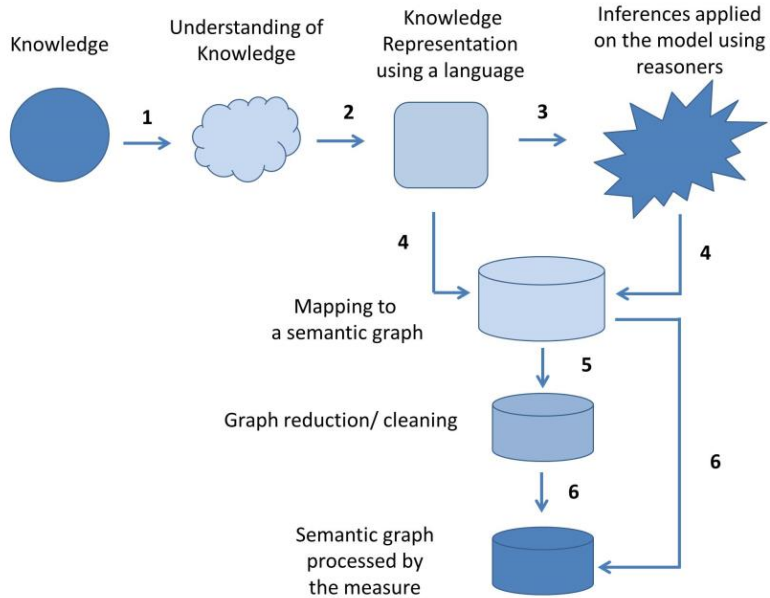

**Figure 7**: Scheme of the main steps used to build a semantic graph from a knowledge representation.

**Figure 7** presents these main steps:

- **Knowledge modelling**: Steps 1 and 2 represent the modelling of a piece of knowledge to a machine understandable and computational representation. Step 2 defines the expression of a KR in a specific language, e.g., OWL, RDF(S). The language which is used conditions the expressivity of the language constructs.

- **Knowledge inference:** Step 3 represents the optional use of a reasoner to infer knowledge implicitly defined in the KR. As an example, in a KR expressed in RDF(S), this step corresponds to the entailment of the RDF graph according to the semantics defined by RDFS, i.e., the use of a reasoner to infer triplets according to RDFS entailment rules.

- **Mapping to a graph representation:** Step 4 is of major importance. It corresponds to the mapping of the knowledge base to the corresponding graph representation which can be processed by certain measures. In some cases, this step is implicit since the knowledge is already expressed through a graph, e.g., taxonomies, WordNet lexical database. Depending on the language used to express the KR, this phase may imply a loss of knowledge and must therefore be carefully considered.

- **Graph reduction / cleaning**: Step 5 corresponds to the removal of some relationships or classes defined in the graph. It may be required to ensure the coherency of SMs.

The phase of knowledge modelling (steps 1 and 2) will not be further discussed in this section. Step 3 briefly mentions knowledge inferences. We mainly focus on the mapping phase used to create a semantic graph from a KR.

### 5.2.2.2 Knowledge Inferences

As we have seen, implicit knowledge can be associated to a KR. Such knowledge may, in some cases, be included into the semantic graph built from the reduction of the KR. As an example, the definition of the domain and range of a specific predicates is important information which can be used to class instances. To ensure that implicit knowledge is inferred according to the semantics associated to the model, inference engines (reasoners) are expected to be used prior to the construction of the resulting semantic graph, e.g., to apply RDFS entailment rules on a RDF graph.

### 5.2.2.3 Mapping To a Graph Representation

In some cases, the mapping of a KR into a semantic graph is straightforward, considering that all the statements are materialized in the representation. As we have seen, in some cases, inferences must been used to generate implicit statements. Nevertheless, in other cases, KRs are defined using specific and expressive languages which imply certain considerations to be taken into account. The aim of this section is obviously not to define all rules which must be used to build a semantic graph from all languages enabling the definition of KRs, e.g. RDF, OWL. We invite the reader to the appendix 4 which discusses some mapping techniques which can be used to extract a semantic graph from some language-specific expressions of KRs, e.g. RDF, OWL.

It is therefore important to consider a distinction between the expression of a semantic graph in a particular language, which can be built on a graph-based formalism such as RDF, and the semantic graph which will be processed by SMs. As an example, OWL KR can be expressed (serialized) using RDF graph as an exchange format. However, since graph syntax based on triplets are limiting, many constructs used in OWL are encoded into a set of triplets (Horrocks & Patel-Schneider 2003). It's also important to understand that some KR defined using expressive language, such as description logic, may therefore only partially be modelled in a graph structure as expected by most SMs.

### 5.2.2.4 Semantic Graph Reductions

We denote $G(O)$, shorten $G$ if there is no ambiguity, the reduction of the KR $O$ to a semantic graph. In addition, we denote $G_{R'}(O)$, also shorten $G_{R'}$ if there is no ambiguity, the reduction of $O$ as a semantic graph only considering the relationships having as predicate $r \in R' \subseteq R$.

A common reduction of a KR as a graph is $G_{\{subclass-of\}}$, shortened by $G_T$ and named the taxonomical reduction. $G_T$ corresponds to the taxonomy $\preccurlyeq_C$, and therefore only contains classes. As we will see this reduction is widely used to compute the similarity between classes.

Graph reductions can naturally be more complex. The graph $G_{Rx}(O)$, with $Rx = \{subClassOf, isA\}$, refers to the reduction which is composed of the relationships having as predicate `subClassOf` or `isA`. We denote such a graph $G_{TI}$ ($T$ stands for *Taxonomical* and $I$ for `isA` relationship).

Studies relying on semantic graphs can be conducted taking the full semantic graph into account or focusing on a particular subgraph. Depending on the amount of information considered, some properties of the graph may change (e.g., acyclicity), along with the strategies and algorithmic treatments used for their processing. Since most SMs require the graph to fulfil specific properties, we briefly discuss the link between graph topologies and SMs.

Considering all types of semantic relationships, a semantic graph generally forms a connected directed graph which can contain cycles, i.e. path from a node to itself. The taxonomical reduction ($G_T$), also leads to a graph given that a class can inherit from multiple classes. Nevertheless, due to the transitivity of taxonomical relationships, $G_T$ is expected to be acyclic.

Taxonomic reductions composed of a unique class which subsumes the others form a Rooted Directed and Acyclic Graph ($RDAG$). $DAG$ properties enable efficient graph treatments to be performed, numerous SMs take advantage of them. The graph $G_{TI}$ is also a $RDAG$.

**Figure 8** presents some of the reductions of a semantic graph which are usually performed prior to consider SMs treatments. This example is based on the reduction of the Gene Ontology (GO) in order to extract the taxonomical knowledge which is related to a specific aspect of the GO. Such a reduction is generally performed before comparing pairs of classes. The figure shows the GO, which is composed of three subparts (sub-graphs): Molecular Function (MF), Biological Processes (BP), and Cellular Component (CC). The GO originally forms a cyclic graph composed of classes linked by various semantic relationships. The first reduction shows the isolation of the MF subpart. Only classes composing the MF subpart and the relationships involving a pair of MF classes are considered. The resulting graph can be cyclic. The final reduction only contains MF classes linked by taxonomical relationships, which corresponds to a $RDAG$.

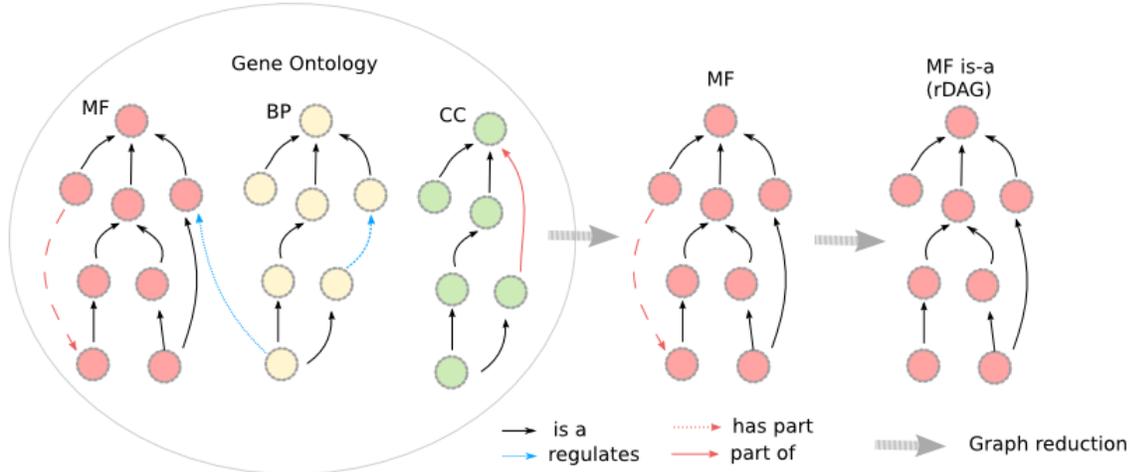

**Figure 8**: Example of a reduction of a KR and its effects on graph properties.

### 5.2.2.5  *Semantic Graph Cleaning*

The accuracy of treatments relying on SMs and the semantics of their results highly depends on the semantic graph which is processed. The better the semantic graph is, the more accurate the SMs results will be. In this context, the quality of semantic graph relies on both, the choices made to model the domain and the way this knowledge is defined. During the definition of semantic graph, relationship redundancies can be introduced. Such redundancies can impact SMs' results and thus have to be removed, e.g., documented in (Park et al. 2011).

Relationship redundancies appear when a direct semantic relationship between two elements can be inferred (explained) by an indirect one[i]. Redundancies involve transitive relationships. As an example, since the taxonomic relationship is transitive, if the semantic graph defines that (`Human, subClassOf, Mammal`) and (`Mammal, subClassOf, Animal`) a semantic reasoner can infer that (`Human, subClassOf, Animal`). In this case, a redundancy occurs when an explicit relationship (non-inferred) defines (`Human, subClassOf, Animal`). Figure 9 shows examples of such redundancies found in the GO.

Formally, a simple case of redundancy can be formalized by:

- Any relationship $(u, t, v)$ is redundant if $\{u, t, v\} \backslash \{(u, t, v)\} \neq \emptyset$.

Relationship redundancies can negatively impact results produced by SMs. **As an example, in** Figure 9, because of these redundancies, a naive SM relying on the edge counting strategy will underestimate the distance between two classes. In this case, the semantic distance between two classes is defined as a function of the length of the shortest path linking them, e.g. the distance between `GO:2000731` and `GO:0031327` will be set to 1 instead of 4.

---

[i] For those familiar to RDF(S), notice that the domain and the range (co-domain) of a predicate, if represented as a relationship, cannot induce redundancies, e.g. the triplet (*is a parent of, rdfs:domain, Human*) doesn't mean that the triplet (*Jean, is-a, Human*) is redundant considering that  (*Jean , is a parent of, Louise*) is specified in the knowledge representation.

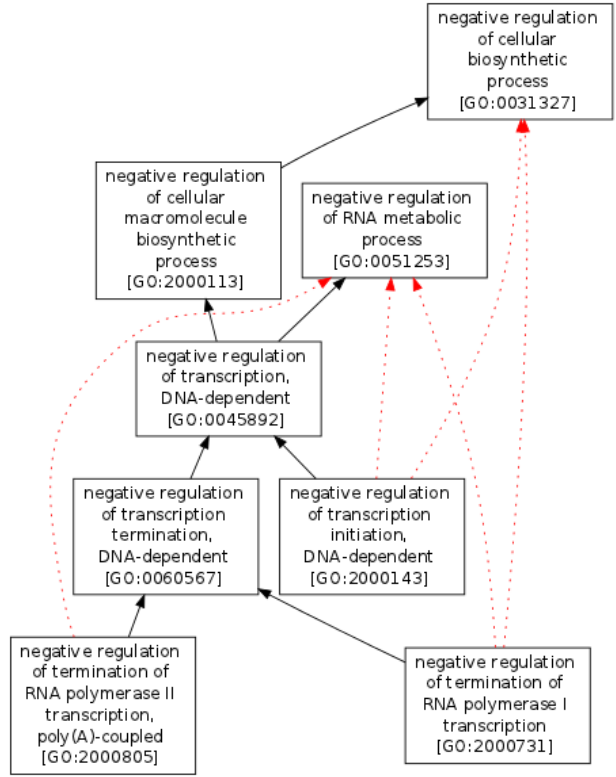

**Figure 9**: Example of redundant relationships which can be found in the Gene Ontology (GO), version 06/03/2012. The graph represents a taxonomy of GO terms. Examples of redundant taxonomical relationships are represented by red-dotted edges. Since the taxonomic relationships are transitive, redundant relationships can be removed without loss of information.

A transitive reduction of $G_T$ has to be done to remove redundant relationships. An adaptation of the algorithm proposed by (Aho et al. 1972) can be used. Moreover, if the whole KR is considered, the transitive reduction has to (i) consider all transitive relationships and (ii) take into account redundant relationships explained by the transitivity of some predicates over others predicates, e.g., `isA`, `hasPart` and `partOf` are transitive over `subClassOf`.

A specific type of such redundancies are redundancies of class membership. As an example, knowing that `(Mammal, subClassOf, Animal)`, defining that `(jean, isA, Mammal)` also implies that `(jean, isA, Animal)`. The latter statement will therefore be considered as redundant if both are defined in a KR.

In some annotation repositories, instances can be annotated by multiple classes/concepts defined in a KR[i]. However, the process of annotation may vary depending on the application context considered, i.e., conceptual annotations of gene products can be defined manually by curators from multiple evidences (e.g. literature, lab experiences) or inferred by algorithms (Hill et al. 2008).

---

[i] Some communities will better understand "Some instances are defined as members of multiple classes".

Nevertheless, usually, SMs expect instances to be characterized by a minimal set of statements. As an example, the transitivity of the taxonomical relationship must be taken into account. It must therefore be considered that an instance is indirectly annotated by (i.e. belongs to) the classes subsuming its assigned classes. This inference rule is defined as the *true path rule* in some communities, e.g. Bioinformatics (Camon et al. 2004). Therefore, if an instance is characterized by the class `OxygenBinding`, it is also characterized by the class `Binding` as the latter subsumes the former.

Due to concurrent and automatic annotations processes, redundant annotations are sometimes largely found in annotation files and large KRs. However, since some SMs can be affected by such redundancies, all inferable annotations are generally expected to be removed. This is for example the case when an instance is regarded as a set of concepts/classes. **Figure 10** shows an example of redundant annotations which have been found in UniprotKB human gene products annotations[i]. Coloured classes[ii] represent the GO annotations of a particular gene product. However, according to the *true path rule*, the red classes are redundant as they can be inferred from those coloured in blue (bold frame).

The question of statement redundancies generalizes the detection of taxonomical redundancies and can also be solved efficiently using an adaptation of a transitive reduction algorithm.

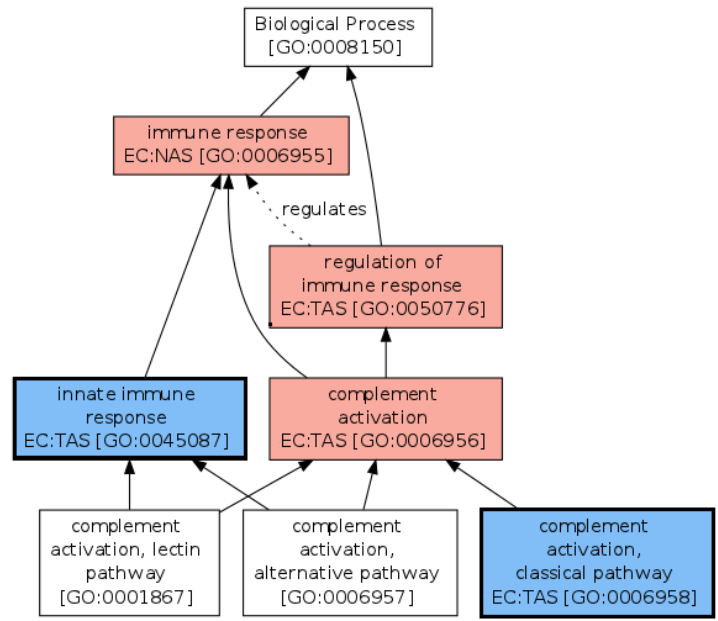

**Figure 10**: An example of redundant annotations in the Gene Ontology (GO). GO version 03/06/2012, annotations GOA human 03/06/2012. These annotations do not take the transitivity of the taxonomical relationship into account. Coloured classes represent P01773 gene GO annotations (`GO:0003823` and `GO:0005576` omitted in this graph). Red classes (normal frame) are redundant as they can be inferred from blue ones (bold frame).

---

[i] We found that 45% of the 45014 UniprotKB annotated human genes contain undesired GO annotations (representing 13% of all human GO annotations) (date 06/2012).

[ii] Here class refers to GO term.

## 5.3 Evidences of Semantics and their Interpretations

A semantic graph carries explicit semantics, e.g. through the taxonomy defining the partial ordering of the classes. It also contains implicit semantics evidences. We here consider *semantic evidences* as any information on which can be based interpretations regarding the meaning carried by the KR or the elements it defines (classes, instances, relationships).

Semantic evidences derive from the study of specific factors which can be used to discuss particular properties of the semantic graph or particular properties of its elemens. Therefore, a semantic evidence, either based on high assumptions or theoretically justified by the core semantics defined in the representation, relies on a particular interpretation of a specific property of the KR, e.g. the number of classes described in a taxonomy gives a clue on the degree of coverage of the ontology.

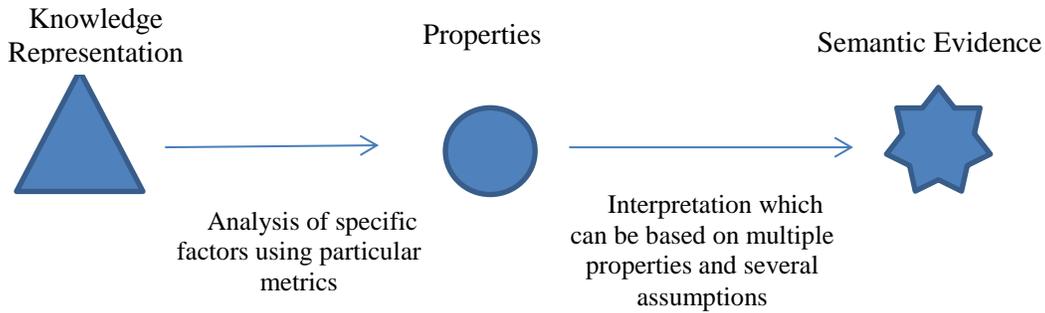

**Figure 11**: Simple process showing how semantic evidences can be obtained from the analysis of the KR.

Figure 11 schematizes the acquisition of semantic evidences which can be obtained **when** mining a semantic graph. Based on the analysis of specific factors, using particular metrics, some properties of both the semantic graph and the elements it defines can be obtained, e.g., the depth of classes and the depth of the taxonomy. Based on these properties, and sometimes considering particular assumptions, semantic evidences can be obtained. As an example one can consider that the deeper a class is regarding the depth of the taxonomy, the more expressive the class is expected to be.

As we will see, several properties are used to consider extra semantics from semantic graphs; they are especially important for designing SMs. Knowing (i) the properties which can be used, (ii) how they are computed and (iii) the assumptions on which their interpretation rely on, is essential for both SMs designers and users. Indeed, semantic evidences are the core elements of measures; they have been used for instance to: (i) normalize measures, (ii) estimate the specificity of classes and to (iii) weight the relationships defined in the graph, that is to say, to estimate the *strength of connotation* between classes/instances.

Most of the properties which are used to obtain semantic evidences correspond to well-known graph properties defined by graph theory. In this section, we only introduce the main properties which are based on the study of the taxonomy $G_T$. We next introduced two applications of these properties: the estimation of the specificity of classes and the estimation of the strength of connotation between classes.

### 5.3.1 Semantics Evidences

In this section we mainly focus on taxonomical evidences. Two kinds of semantic evidences can be distinguished:

- *Intentional evidences* also called intrinsic evidences. They are based on the analysis of properties associated to the topology of the graph, and mainly rely on the analysis of the topology of the taxonomy.

- *Extensional evidences*. They are based on the analysis of both the topology of the graph and distribution of the usage of classes (instance memberships)[i], i.e., the number of instances associated to classes. In short, interpretations regarding the semantics of the classes can be driven by the study of their usage.

Among these evidences, we can distinguish those which are related to the properties of the full KR, i.e., based on global properties, from those which are related to specific elements (classes, instances, relationships) and rely on local properties analysis.

### *5.3.1.1 Intentional Evidences*

#### *5.3.1.1.1 Global Properties*

- Depth of the taxonomy / maximal number of super classes.

The depth of the taxonomy corresponds to the maximal depth in $G_T$. It informs on the degree of expressiveness/granularity of the taxonomy. As an example, the deeper $G_T$, the more detailed the taxonomy is expected to be.

The maximal number of super classes of a class has is also used as an estimator of the upper bound of the maximal degree of expressivity of a class. Inversely, the number of classes defined in the taxonomy (i.e., the number of subclasses of the root) can also be used as an upper bound of the maximal degree of generality of a class defined in the taxonomy.

- Width of the taxonomy.

The width of taxonomy corresponds to the length of the longest shortest path which links two classes in $G_T$. It informs on the degree of coverage of the taxonomy. Generally the taxonomy is assumed to better cover a domain the more important its width is.

---

[i] Notice that we don't consider semantic evidences only based on the usage of classes, i.e., without taking into account the taxonomy. Indeed, in most cases, to be meaningful, the distribution of the usage of classes must be evaluated considering the transitivity of the taxonomic relationship. If this is not the case, incoherent results could be obtained, e.g., that a class contains more instances than one of its superclass.

##### 5.3.1.1.2 *Local Properties*

- Local density.

It can be considered that relationships in dense part of a taxonomy represent smaller taxonomical distances. Metrics such as *compactness* can be used to characterize local density (Botafogo et al. 1992)[i]. Other metrics such as the (in/out)-branching factor of a class, i.e., the number of neighbours of a given class[ii], can also be used (Sussna 1993). It is generally assumed that more important the number of subclasses of a class is, the more general the class is.

- Number of super classes of class / depth / number of subsumed classes / number of subsumed leaves / distance to leaves.

The number of super classes of a class is often considered to be directly proportional to its degree of expressiveness. The more a class is subsumed, the more detailed/restrictive the class is expected to be. The number of superclasses can also be interpreted with regard to the maximal number of superclasses a class of the taxonomy can have.

The depth of a class is also expected to be directly proportional to its degree of expressiveness. The more the depth of a class (according to the maximal depth), the more detailed/restrictive the class is regarded[iii]. The depth of a class can also be evaluated according to the depth of the branch in which it is defined.

In a similar fashion, in some cases, the distance of a class to the leaves it subsumes or the number of leaves it subsumes will be considered as an estimator of expressiveness, the more the distance/number the less expressive the class is considered.

#### 5.3.1.2 *Extensional Evidences*

##### 5.3.1.2.1 *Global Properties*

- Distribution of the instances among the classes.

The distribution of the instances among the classes can be used to evaluate the balance of the distribution and to design local correction factors, e.g. to correct the expressiveness of a class.

##### 5.3.1.2.2 *Local Properties*

- Number of instances associated to a class.

---

[i] (Botafogo et al. 1992) also introduces interesting factors for graph-based analysis; the depth of a node is also introduced.
[ii] Called (in/out) degree of a node in graph theory.
[iii] Note that the depth of a class as an estimator of its degree of expressiveness can be seen as an inverse function of the notion of status already introduced by (Harary & Norman 1965) to analyze status phenomena in organization.

The number of instances of a class is expected to be inversely proportional to its expressiveness, the less a class has instances, the more specific the class is expected to be.

These semantic evidences and their interpretations have been used to characterize notions extensively used by SMs. They are indeed used to estimate the specificity of classes as well as the strength of connotations between classes.

### 5.3.2 Estimation of Class Specificity

Not all classes have the same degree of informativeness, specificity. Indeed, most people will commonly agree that the class *Dog* is a more specific description for a *Living Being* than the class *Animal*.

The notion of specificity can be associated to the concept of salience defined by Tversky to characterize a stimulus according to its '*intensity, frequency, familiarity, good form, and informational content*". In (Bell et al. 1988) it is also specified that "*salience is a joint function of intensity and what Tversky calls diagnosticity, which is related to the variability of a feature in a particular set* [universe, collection of instances]". The idea is to capture the amount of information carried by a class which is expected to be directly proportional to its degree of specificity and proportional to its degree of generality.

The notion of specificity of classes is not completely artificial and can be explained by the root of the taxonomic organization of knowledge. Indeed, the transitivity of the taxonomical relationship specifies that not all classes have the same degree of specificity or detail. The ordering of two classes defines that the class which subsumes the other has to be considered as the more abstract one (less specific). In fact, the taxonomy explicitly defines that if a class $v$ subsumes another class $u$, all the instances of $u$ are also instances of $v$:

$$u \preccurlyeq v \Rightarrow I(u) \subseteq I(v)$$

This expression is represented by **Figure 12** in which we can see that the more a class is subsumed by numerous classes: (A) the number of properties which characterize the class increases, and (B) the number of instances associated to the class decreases.

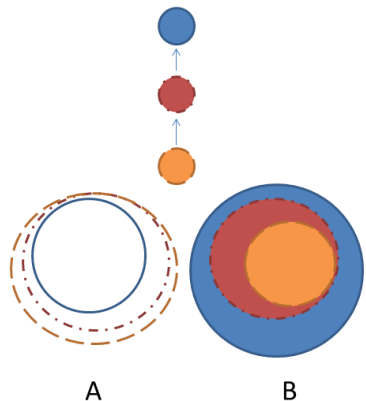

**Figure 12**: **S**et-based representations of ordered classes according to (A) their intentional expressions in term of properties characterizing the classes, and (B), in term of their extensional expressions, i.e., the set of instances which compose the classes. Figure based on (Blanchard 2008).

Therefore, another way to compare the specificity of two ordered classes is to study their usage, analysing a collection of instances. Indeed, in a total order of classes[i], the comparison of the degree of specificity of two classes can be made regarding their number of instances. The class which contains the highest number of instances will be the less specific (its universe of interpretation is larger). In this case, it is therefore possible to assess the specificity of ordered classes either studying the topology of their ordering or the set of instances associated to these classes.

Nevertheless, in taxonomies, classes are generally only partially ordered. This means that the evidences used to compare the specificity of two classes without assumption cannot be used anymore, i.e., classes which are not ordered are in some sense not comparable. This aspect is underlined by Figure 13. It's impossible to compare, in an exact manner, the specificity of two non-ordered classes. This is due to the fact that the amount of shared and distinct properties can only be estimated regarding the properties characterizing the common class they derive from. However this estimation can only be a lower bound since extra properties shared by the two instances may not be carried by such a common class.

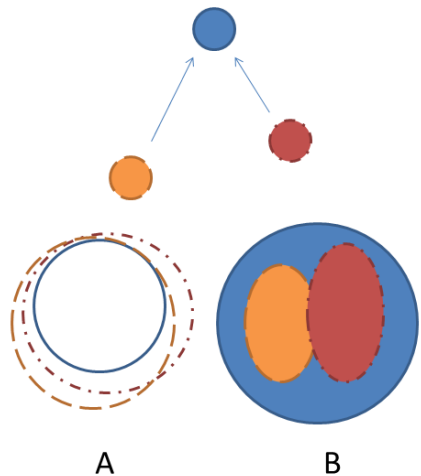

**Figure 13:** Potential set-based representations of ordered classes according to (A) their intentional expressions in term of properties characterizing the classes, and (B), in term of their extensional expressions, i.e., the set of instances which compose the classes. Figure based on (Blanchard 2008).

However, the appreciation of the degree of specificity of classes is of major importance in the design of SMs. Therefore, given that discrete levels of class specificity are not explicitly expressed in a taxonomy, various approaches have been explored to define a function $\theta: C \rightarrow \mathbb{R}^+$ in the aim to evaluate the degree of specificity of classes. The function $\theta$ relies on the intrinsic and extrinsic properties presented above.

---

[i] for any pair of classes $u, v$ either $u \preccurlyeq v$ or $v \preccurlyeq u$.

The function evaluating the specificity of classes must be in agreement with the taxonomical representation which defines that classes are always semantically broader than their specializations[i]. Thus, the function which estimates the specificity of classes must monotonically decrease from the leaves (classes without subclasses) to the root of the ontology, i.e.:

$$x \preccurlyeq y \Rightarrow \theta(x) \geq \theta(y)$$

We present the main approaches defined in the literature to express such a function $\theta$.

### 5.3.2.1 Basic intrinsic estimators of class specificity

The specificity of classes can be estimated considering the location of its corresponding node in the graph. A naive approach will define the specificity of the class $c$, $\theta(c)$, as a function of some simple properties related to $c$, e.g., $\theta(c) = f(depth(c))$, $\theta(c) = f(A(c))$ or $\theta(c) = f(D(c))$ with $A(c)$ and $D(c)$ the ancestors and descendants of $c$.

The main drawback of simple specificity estimators is that classes with similar depth or equal number of superclasses/subclasses will have similar specificities, which is not always true. In fact, two classes can be described with various degrees of detail independently of their depth, e.g., (Yu, Jansen & Gerstein 2007). More refined $\theta$ functions have been proposed to address this limitation.

### 5.3.2.2 Extrinsic Information Content

Another strategy explored by SMs designers has been to characterize the specificity of classes according to the well-known theoretical framework established in computer-sciences, namely Shannon's Information Theory. The specificity of a class will further be regarded as the amount of information the class conveys, its Information Content (IC). The IC of a class can for example be estimated as a function of the size of the universe of interpretations associated to it. The IC is a common expression of the $\theta$ function and was originally defined by (Resnik 1995) to assess the informativeness of concepts.

The IC of the class $c$ is defined as inversely proportional to $p(c)$, the probability to encounter an instance of $c$ in a collection of instances (negative entropy). The original definition of the IC was proposed to estimate the informativeness of a concept as a function of its number of occurrences in a corpus of texts.

We denote $eIC$ any IC which relies on extensional information, i.e., a corpus or collection of instances. We consider the formulation of $eIC$ originally defined by (Resnik 1995):

$$p(c) = \frac{|I(c)|}{|I|}$$

---

[i] This explains that the specificity of classes cannot be estimated only considering extrinsic information such as the number of instances directly characterized by a class (without inference). Indeed, the partial ordering of classes also needs to be taken into account when the specificity of classes is estimated. If the transitivity of the taxonomic relationship is not considered to propagate class usage/instance membership, the instance distribution can be incoherent with regard to the partial order defined in the underlying taxonomy, i.e., a class can have less instances than one of its superclass.

$$eIC_{Resnik}(c) = -\log\big(p(c)\big)$$

$$eIC_{Resnik}(c) = \log(|I|) - \log(|I(c)|)$$

with $I(c)$ the set of instances of the class $c$, e.g., occurrences of a concept in a corpus. The suitability of the log function can be supported by the work of (Shepard 1987)[i]. Notice also the link with Inverse Document Frequency (IDF) commonly used in information retrieval (Jones 1972):

$$IDF(c) = \log(|I|/|I(c)|)$$

$$IDF(c) = \log(|I|) - \log(|I(c)|)$$

The main drawback of $\theta$ functions based on extrinsic information is that they highly depend on the usage of the classes and will therefore automatically reflect its biases[ii]. In some cases, such a strong dependence between class usage and the estimation of its specificity is desired as all classes which are highly represented will be considered as less informative, even the classes which will be considered to be specific regarding intrinsic factors (e.g., depth of classes). However, in some cases, biases in class usage can badly affect the estimation of the IC and may not be adapted. In addition, the IC computation based on text analysis can be time consuming and challenging given that, in order to be accurate, complex disambiguation techniques have to be used to detect to which concept/class refers an occurrence of a word.

### 5.3.2.3 Intrinsic Information Content

In order to avoid the dependency of $eIC$ calculus to statistics related to class usages, various intrinsic IC formulations ($iIC$) have been proposed. They can be used to define $\theta$ functions by only considering structural information extracted from the KR, e.g., the intrinsic factors we presented in the previous section 5.3.1.1. This IC formulation extends the basic specificity estimators presented above.

Multiple topological characteristics can be used to express $iIC$, e.g., number of descendants, ancestors, depth, etc. (Seco et al. 2004; Schickel-Zuber & Faltings 2007; Zhou et al. 2008; Sánchez et al. 2011). The formulation proposed by (Zhou et al. 2008) is presented. It enables to refine the contribution of both depth and number of subclasses ($D(c)$) to compute class specificity.

$$iIC_{Zhou}(c) = k\left(1 - \frac{\log(|D(c)|)}{\log(|C|)}\right) + (1-k)\left(\frac{\log\big(depth(c)\big)}{\log(\max\_depth)}\right)$$

With $|C|$ the number of classes defined in the taxonomy, $depth(c)$ the depth of the class $c$, $max\_depth$ the maximal depth of the ontology and $k \in [0,1]$.

---

[i] Shepard derived his universal law of stimulus generalization based on the consideration that logarithm functions are suited to approximate semantic distance (Al-Mubaid & Nguyen 2006), please refer to section 2.2.1 for more details.

[ii] As an example, this can be problematic for GO-based studies as some genes are more studied and annotated (e.g. drug related genes) and annotation distribution patterns among species reflect abnormal distortions, e.g. human/mouse (Thomas et al. 2012).

$iICs$ are of particular interest as only the topology of the ontology is considered. They prevent errors related to biases on class usage. However, the relevance of $iIC$ relies on the assumption that the ontology expresses enough knowledge to rigorously evaluate the specificities of classes. Therefore, as a counterpart, $iICs$ are sensitive to structural biases in the taxonomy and are therefore sensible to unbalanced taxonomy, degree of completeness, homogeneity and coverage of the taxonomy (Batet, Sánchez & Valls 2010).

### 5.3.2.4 Non-taxonomical Information Content

Both introduced $iIC$ and $eIC$ only take taxonomical relationships into account. (Pirró & Euzenat 2010a) proposed the *extended IC* ($extIC$) in order to take advantage of all predicated and semantic relationships.

$$extIC(c) = \alpha EIC(c) + \beta IC(c)$$

$$EIC(c) = \sum_{r \in R} \frac{\sum_{u \in C(c,r,both)} iIC(u)}{|C_{in/out}(u,r)|}$$

With $C_{in/out}(u,r)$ the set of classes linked to the class $c$ by any relationship of type $r \in R$.

In this formula, the contribution of the various relationships of the same predicate is averaged. However, the more a class establishes relationships of different predicates, the more its $extIC$ will be high. We thus propose to average the $extIC$ (by $|R|$) or to weight the contribution of the various types of relationships.

**Table 3** lists the various expressions of the function $\theta$ which have been expressed in the literature.

| Names and References | Use class Extension | Co-domain | Comment |
|---|---|---|---|
| Depth | No | [0, 1] | Normalized depth or max depth can be used. In a graph, considering the minimal depth of a class doesn't ensure that the specificity increases according to the partial ordering (due to multi-inheritance). |
| Depth non-linear (Seco 2005) | No | [0, 1] | Use log to introduce non-linear estimation. |
| IDF (Chabalier et al. 2007) | Yes | [0,inf[ | Inverse Document Frequency (IDF) obtained by dividing the number of instances by the number of instances of the class, i.e. $IDF(c) = \log(|I|/|I(c)|)$. |
| IC Resnik (Resnik 1995) | Yes | [0,inf[ , [0, 1] | IC depends on class usage. $IC(c) = -\log(|I(c)|/|I|)$. Normalized version have also been proposed, e.g., (Sheehan et al. 2008). |
| IC Resnik intrinsic (Resnik 1995) | No | [0, 1] | Resnik's IC with $\forall c \in C, |I^-(c)| = 1$. |
| IC Seco (Seco 2005) | No | [0, 1] | IC estimated from the number of descendants/subclasses. |
| IC Zhou (Zhou et al. 2008) | No | [0,1] | Parametric hybrid iIC mixing Seco's IC and nonlinear depth using a contribution factor k (originally set at 0.6). |
| IC Sanchez et al (A) (Sánchez et al. 2011) | No | [0,inf[ | Consider the number of leaves contained in D(c), the more the number of leaves is high, the less specific c is considered. |
| IC Sanchez et al. (B) (Sánchez et al. 2011) | No | [0,inf[ | Refined version of version A (see above) exploiting the number of subsumed classes (descendants) of a class. |
| Yu et al. (TAM) (Yu, Jansen & Gerstein 2007) | Yes | [0,inf[ | The probability p(c) associated to a class is computed as the number of pairs of instances which are members of c divided by total the number of pairs. |
| extIC (Pirró & Euzenat 2010a) | No | [0, 1] | iIC based on all predicates. |
| APS (Schickel-Zuber & Faltings 2007) | No | [0,1/2] | iIC based on the number of descendants of a class. |

**Table 3**: Selection of $\theta$ functions which can be used to estimate the specificity of a class defined in a taxonomy. The estimation can be based on an intrinsic strategy, i.e. only evaluating the topology of the taxonomy, or taking advantage of the extensional information associated to the classes.

We have presented various strategies which can be used to estimate the specificities of classes defined in a partially ordered set. It's important to understand that these estimators are based on assumptions regarding the representation of the knowledge.

### 5.3.3 Estimation of Strength of Connotation between Classes

A notion strongly linked to the specificity of classes is the estimation of the strength of connotation between instances or classes, i.e., the strength of the relationship linking two instances or two classes.

Considering taxonomical relationships, it is generally considered that the strength of connotation between classes is stronger the deeper two classes are in the taxonomy. As an example, the strength of the taxonomical relationship linking `SiberianTiger` to `Tiger` will generally considered to be more important than the one linking `Animal` to `LivingBeing`. Such a notion is quite intuitive and has for instance been studied by Quillian and Collins in the early studies of semantic networks (Collins & Quillian 1969) - Hierarchical-Network model were built according to mental activations evaluated based on the time people took to correctly response to sentences linking two classes, e.g., *a* `Canary` *is an* `Animal` - *a* `Canary` *is a* `Bird` − *a* `Canary` *is a* `Canary`. Showing the variation of the response time to correctly response to sentences involving two ordered classes (`Canary / Animal`), the authors highlighted the variation of the strength of connotation and the link with the notion of specificity of classes.

It's worth to note that the estimation of the strength of connotation of two linked classes is in some sort a measure of the semantic similarity or taxonomical distance between the two ordered classes. The aim of the model proposed to define the strengths of connotation between classes is generally based on the assumption that the taxonomical distance conveyed by a taxonomical relationship *shrinks* with the depth of the two linked classes (Richardson et al. 1994). Given that the strength of connotation between classes is not explicitly expressed in a taxonomy, it has been proposed to consider several intrinsic factors to refine its estimation, e.g., (Young Whan & Kim 1990; Sussna 1993; Richardson et al. 1994).

A taxonomy *only* explicitly defines the partial ordering of its classes, which means that if a class $v$ subsumes another class $u$, all the instances of $u$ are also instances of $v$, i.e., $u \preccurlyeq v \Rightarrow I(u) \subseteq I(v)$. Nevertheless, non-uniform strength of connotation aim to consider that all taxonomic relationships do not convey the same semantics.

Strictly speaking, taxonomic relationships *only* define class ordering and class inclusion. Therefore, according to the extensional interpretation of a class ordering, the universe of interpretation of a class, i.e., the set of possible instances of the class regarding the whole set of instances, must reduce the more a class is specialized[i]. This reduction of the universe of eligible interpretations corresponds to a specific understanding of the semantics of non-uniform strengths of connotation. Alternative explanations which convey the same semantics can also be expressed according to the insights of the various cognitive models which have been introduced in section 2.2:

- Spatial/Geometric model, it states that the distance between classes is a non-linear function which must take into account class saliency.

---

[i] We here consider a finite universe

- Feature model (which states that a class is represented by a set of properties). It can be seen as the difficulty to further distinguish a class which is meaningful to characterize the set of instances of a domain.
- Alignment and Transformational models: the effort of specialization which must be done to extend a class increases the more a class has been specialized.

All this interpretations state the same central notion - the strength of connotation linking two classes is a function of two factors: (i) the specificities of the linked classes and (ii) the variation of specificity between the two compared classes. The several semantic evidences introduced in the previous section, as well as the notion of information content of classes, can be used to assess the strength of connotation between two classes.

As an example, a simple model for the definition of the strength of connotation $w$ associated to a taxonomical relationship linking two classes $u, v$ with $u \preccurlyeq v$ can be defined as a function of the information content of $u$ and $v$ (Jiang & Conrath 1997).

$$w(u, v) = IC(u) - IC(v)$$

It's important to stress that supporting the estimation of the strength of connotations according to the density of classes, the branching factor, the maximal depth or the width of the taxonomy is based on assumptions regarding the definition of the KR (once again, refer to the section 5.3.1.1 which presents the semantic evidences and the assumptions on which they are based on).

## 5.4 Types of Semantic Measures and Graph Properties

Depending on the properties of the semantic graph the SMs evaluate, two main groups of measures can be distinguished:

- Measure adapted to semantic graphs composed of (multiple) predicate(s) potentially inducing cycles.
- Measure adapted to acyclic semantic graphs composed of a unique predicate inducing transitivity.

### 5.4.1 Semantic Measures on Cyclic Semantic Graphs

As we have seen, considering all predicates defined in $G$ potentially leads to a cyclic graph. Only few SMs framed in the relational setting are designed to deal with cycles. Since these measures take advantage of all predicates, they are generally used to evaluate the semantic relatedness and not the semantic similarity. Notice that they can be used to compare concepts and instances. Two types of measures can be further distinguished:

- *Graph-Traversal measures,* pure graph-based measures. These measures have initially been proposed to study node interactions in a graph and essentially derive from graph theory contributions. They can be used to estimate the relatedness of nodes considering that the more two nodes interact, directly or indirectly, the more related they are. These measures are not SMs *per se* but graph measures used to compare nodes. However, they can be used on semantic graphs and can also be adapted in order to take into account evidences of semantics defined in the graph.

- *Graph property model measures* designed to compare elements described using the graph property model. These measures consider classes or instances as a set of properties expressed in the graph.

### 5.4.1.1 Graph-Traversal Measures

Measures based on graph traversals can be used to compare any pair of classes or instances, represented as a node. These measures rely on algorithms designed for graph analysis which are generally used in a straightforward manner. Nevertheless, some adaptations have also been proposed in order to take into account the semantics defined in the graph. Among the large diversity of measures and metrics which can be used to estimate the relatedness of two nodes in a graph, we distinguish:

- Shortest path approaches.
- Random-walk techniques.
- Other interconnection measures.

The main advantage of these measures is their unsupervised nature. Their main drawback is the absence of extensive control over the semantics taken into account, which generates difficulties in justifying and explaining the resulting scores. However, in some cases, these measures enables fine-grain control over the predicates considered. Indeed, approaches have naturally proposed to tune the contribution of each relationship or predicate in the estimation of the relatedness.

### 5.4.1.1.1 <u>Shortest path Approaches</u>

The shortest path problem is one of the most ancient problems of graph theory. The intuitive edge-counting strategy can be used on any graph. It can be applied to compare both pairs of instances and classes, considering their relatedness as a function of the distance between the nodes corresponding to the two compared elements. More generally, the relatedness is estimated as a function of the weight of the shortest path linking them. Classical algorithms proposed by graph theory can be used; the algorithm to use depends on specific properties of the graph (e.g., does it contains cycles? Are their nonnegative weights associated to relationships? Is the graph considered to be oriented?).

(Rada et al. 1989) were among the first to use the shortest path technique to compare two classes defined in a semantic graph. This approach is sometimes denoted as the edge-counting strategy in the literature – edge here refers to relationship. Because the shortest path can contain relationships of any predicate we call it *unconstrained shortest path* (*usp*).

One of the drawbacks of usp-based techniques is that the meaning of the relationships from which derive the relatedness is not taken into account. In fact, complex semantic paths involving multiple predicates and those only composed of taxonomic relationships are for instance considered equally. Therefore, propositions to penalize *usp* reflecting complex semantic relationships have been proposed (Hirst & St-Onge 1998; Bulskov et al. 2002) . Approaches have also been proposed to consider particular predicates in a specific manner. To this end, a weighting scheme can also be applied to $G$ in order to tune the contribution of each relationship or predicate in the computation of the final score.

These techniques are based on a Markov chain model of random walk. The random walk is defined through a transition probability associated to each relationship. The walker can therefore walk from node to node, i.e., each node represents a state of the Markov chain. Based on random walk techniques, several measures can be designed to compare two nodes $u$ and $v$. A selection of measures introduced in (Fouss et al. 2007) is listed:

- The average first-passage time, i.e. the average number of steps needed by the walker starting to $u$ to reach $v$.
- The average commute time, Euclidean Commute Time Distance.
- The average first passage cost.
- The pseudo inverse of the Laplacian matrix.

These approaches are closely related to spectral-clustering and spectral-embedding techniques (Saerens et al. 2004). Examples of measures based on random Walk techniques defined in the literature are (Hughes & Ramage 2007; Fouss et al. 2007; Ramage et al. 2009; Alvarez & Yan 2011; Garla & Brandt 2012). Approaches based on graph-kernel can also be used to estimate the *similarity* of two nodes of a graph (Kondor & Lafferty 2002) and have been used to design SMs (Guo et al. 2006).

As an example, the Hitting time $H(u, v)$ of two nodes $u, v$ is defined as the expected number of steps a random walker starting from $u$ will do before $v$ is reached. The hitting time can be defined recursively by:

$$H(u, v) = 1 + \sum_{k \in N_{out}(u)} p(u, k) \, H(k, v)$$

With $N_{out}(u)$ the set of nodes which are linked to $u$ by an outgoing relationship starting from $u$ and $p(u, k)$ the transition probability of the Markov Chain, i.e., $p(u, k) = \frac{w(u,k)}{\sum_{i \in N_{out}(u)} w(u,i)}$ with $w(u, k)$ the weight of the relationship linking $u$ to $k$.

The commute distance corresponds to $C(u, v) = H(u, v) + H(v, u)$, the expected time a random walker travel from $u$ to $v$ and back to $u$. Therefore, the more paths connect $u$ and $v$, the smaller their commute distance becomes. Critics of classical approach to evaluate hitting and commute approaches, as well as extensions, have been formulated in the literature, please refer to (Sarkar et al. 2008; von Luxburg et al. 2010).

These measures take advantage of second-order information which are generally hard to interpret in semantic terms.

### *5.4.1.1.3 Other Measures based on Interaction Analysis*

Several approaches exploiting graph structure analysis can be used to estimate the similarity of two nodes of a graph through their interconnections. Such approaches estimate the proximity between two elements without explicitly taking into account the semantics carried by the graph. Consequently, the more the elements are interconnected, either directly or indirectly, the more related they will be assumed.

(Chebotarev & Shamis 2006a; Chebotarev & Shamis 2006b) proposed to take into account indirect path linking two nodes using the matrix-forest theorem. SimRank, proposed by (Jeh & Widom 2002), is an example of such a measure. Considering $N$ as the set of nodes of the graph, $N_{in}(n)$ the nodes linked to the node $n$ by a single relationship ending to $n$ (i.e., in-neighbors), and $N_{out}(n)$ the nodes linked to $n$ by a single relationship starting from $n$ (i.e., out-neighbors), SimRank similarity is defined by:

$$sim_{SimRank}(u,v) = \frac{|N|}{|N_{in}(u)||N_{in}(v)|} \sum_{i=1}^{N_{in}(u)} \sum_{j=1}^{N_{in}(v)} s\left(C_{in}^{i}(u), C_{in}^{j}(v)\right)$$

SimRank is a normalized function. An adaptation of this measure have been proposed for semantic graph built from linked data (Olsson et al. 2011).

### *5.4.1.2 Semantic Measures for the Graph Property Model*

The second approach which can be used to compare pair of instances and concepts defined in a (potentially) cyclic semantic graph are measures associated to the graph property model.

These measures take advantage of semantic graphs encompassing expressive definitions of classes and instances through properties. The properties may sometimes refer to specific data types, e.g. strings. Therefore, the nodes composing the semantic graph can be data value, classes or instances. The semantic graphs generally correspond to RDF graphs or labelled graphs. Note that in RDF, a property corresponds to a specific type of relationship, i.e. predicate. Therefore, a direct property $p$ of an element $u$ corresponds to the set of values associated to $u$ through the relationships of predicate $p$.

SMs based on property analysis can be used to consider specific properties of compared elements. These measures are particularly useful to compare objects defined in (relational) databases, to design ontology mapping algorithms and to perform instance matching treatments in heterogeneous knowledge bases. We present three approaches which can be used to compare elements through this notion of property.

*5.4.1.2.1 Elements Represented as a List of Direct Property*

An element can be evaluated by studying its direct properties, i.e., the set of values[i] associated to the element according to the study of a particular predicate. As an example, two types of predicate can be associated to an instance:

- Taxonomic relationships, i.e., the relationship links the instance to a class.
- Non-taxonomical relationships:
    - The property links the instance to another instance[ii].
    - The property links the instance to a specific data value[iii].

Two elements will be compared with regard to the values associated to each property considered. Each property is therefore associated to a specific measure which is used to compare the values taken by this property.

Properties which link an instance to other instances are in most occasions compared using set-based measures, which for example will evaluate the quantity of instances of shared sets (e.g., the number of music genres two groups have in common). Taxonomical properties are evaluated using SMs adapted to the class comparison. Properties associated to data values can be compared using measures adapted to the type of data considered, e.g., in using a measure to compare dates if the corresponding property is a date.

The scores produced by the various measures associated to the various properties are aggregated in order to obtain a global score of relatedness for two instances (Euzenat & Shvaiko 2007). Such a representation has been formalized in the framework proposed by (Ehrig et al. 2004). This is a strategy commonly adopted in ontology alignment, instance matching or link discovery between instances; SemMF (Oldakowski & Bizer 2005), SERIMI (Araujo et al. 2011) and SILK (Volz et al. 2009) are all based on this approach.

*5.4.1.2.2 Elements Represented as An Extended List of Property*

Several contributions underscore the relevance of indirect properties in comparing entities represented through graphs, especially in object models (Bisson 1995). Referring to the KR proposed in **Figure 14**, such a representation might be used to consider the characteristics (properties) of the music genres for the purpose of comparing two music bands.

---

[i] Note that we here do not refer to the specific data value as the value can be a class or another instance.
[ii] Object properties in OWL.
[iii] Datatype properties in OWL.

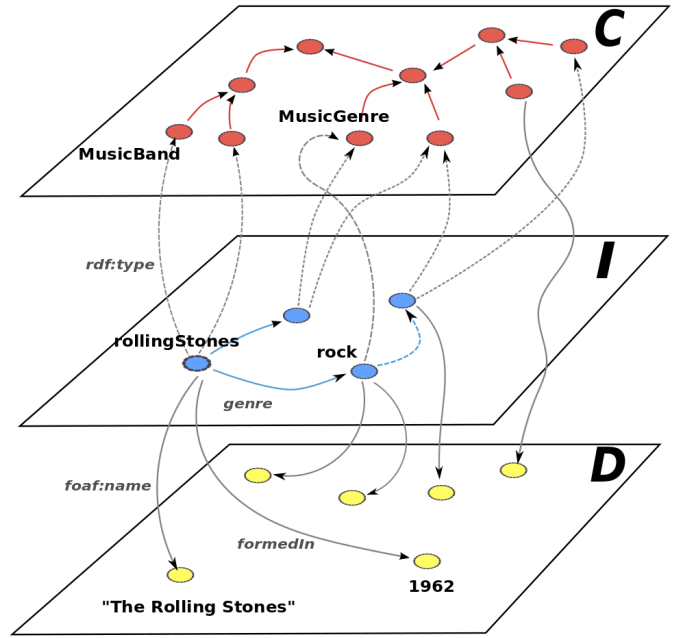

**Figure 14**: Example of a semantic graph involving classes, instances and data values (Harispe, Ranwez, et al. 2013a).

This approach relies on a representation of the compared elements which is an extension of the canonical form to represent an element as a list of properties. This approach can be implemented to take into account indirect properties of compared elements, e.g., properties induced by the elements associated to the element we want to characterize.

A formal framework, enhancing the one proposed in (Ehrig et al. 2004), has thus been proposed to capture some of the indirect properties (Albertoni & De Martino 2006). This framework is dedicated to the comparison of instances. It formally defines an indirect property of an instance along a path in the graph. The indirect properties to be taken into account are defined for a class and depend on a specific context, e.g. application context.

From a different perspective, Andrejko and Bieliková (Andrejko & Bieliková 2013) suggested an unsupervised approach for comparing a pair of instances by considering their indirect properties. Each direct property shared between the compared instances plays a role in computing the global relatedness. When the property links one instance to another, the approach combines a taxonomical measure with a recursive process to take into account the properties of instances associated with the instance being processed.

Lastly, in estimating the similarity between two instances, the measure aggregates the scores obtained during the recursive process. The authors have also proposed to weight the contributions of the various properties so as to define a personalized information retrieval approach.

*5.4.1.2.3* <u>*Elements Represented Through Projections*</u>

The framework proposed by (Harispe, Ranwez, et al. 2013a) enables to compare elements through different projections characterizing the properties of interest.

The approach has initially been defined to compare two instances, but can also be used to compare classes. We here present the perspectives it opens for the comparison of instances. In addition to the formal characterization of an instance through an extended list of properties, this framework also considers complex properties of instances, i.e., properties that rely in combining various properties.

Indeed, using the other approaches, considering that the weight and the size of several persons have been specified in the graph, it is impossible to compare two persons by taking into account their body mass index, a metric which can be computed from the weight and the height. Therefore according to the following statements:

```
luc asWeightInKG 70 .
luc asHeightInM 1,75 .

marc asWeightInKG 85 .
marc asHeightInM 1,80 .

steve asWeightInKG 120 .
steve asHeightInM 1,70
```

Luc and Marc will be regarded as more similar than Luc/Steve and Marc/Steve according to their body mass index. Therefore the main idea is to enable the definition of complex property reflecting properties of the compared elements which are not materialized into the graph. The framework proposed by (Harispe, Ranwez, et al. 2013a) can be used to consider such properties.

Finally, the comparison of two instances is made through the aggregation of the value of *similarity* which is associated to the compared projections. As an example, the authors used a simple weighted sum in their experiments.

### 5.4.2 Semantic Measures on Acyclic Graphs

Note that all the measures which can be used on the whole semantic graph G can also be used for any acyclic reduction $G_R \subseteq G$. Nevertheless, numerous SMs have been defined to work on a reduction of $G$. Depending on the topological properties of the reduction, two cases can be distinguished:

1. The reduction $G_R$ leads to a cyclic graph. Adapted SMs are therefore those previously presented for cyclic graphs in section 5.4.1.

2. $G_R$ is acyclic, then particular techniques and algorithms can be used. Most SMs defined for acyclic graphs focus on taxonomical relationships defined in $G_R$ and consider the reduction to be the taxonomy of classes $G_T$. However, some measures consider a specific subset of $R$, e.g., $R = \{isA, partOf\}$, which also produces in some cases an acyclic graph (Wang et al. 2007). The measures which can be used in this case are usually designed from a generalization of semantic similarity measures, i.e. measures only considering $G_T$.

SMs applied to graph-based KRs were originally designed for taxonomies, i.e. , $G_T$. Since most KRs are usually mainly composed of taxonomic relationships or poset structures, substantial literature is dedicated to semantic similarity measures. Thus, most SMs designed for semantic graphs focus on $G_T$ and have been defined for the comparison of pairs of classes.

## 5.5 Semantic Similarity between Pairs of Classes

The majority of SMs framed in the relational setting have been proposed to assess the semantic similarity or taxonomical distance of a pair of classes defined in a taxonomy. Given that they are designed to compare two classes, these measures are denoted *pairwise measures* in some communities, e.g., in bioinformatics (Pesquita, Faria, et al. 2009). As we will see, an extensive literature is dedicated to these measures as they can be used to compare any pairs of nodes expressed in a graph which defines a (partial) ordering, that is to say, any graph structured by relationships which are transitive, reflexive and antisymmetric (e.g., `isA`, `partOf`).

In section 3.4.1, we already distinguished the main approaches used for the definition of knowledge-based SMs framed in the relational setting. Considering the measures which can be applied to acyclic graphs, we distinguished:

- **Measures based on graph structure analysis**. They estimate the similarity as a function of the degree of interconnection between classes. They are generally regarded as measures framed in the spatial model – the similarity of two classes is estimated as a function of their distance in the graph, e.g., based on the analysis of the lengths of the paths linking the classes. These measures can also be considered to be framed in the transformational model, considering them as functions which estimate the similarity of two classes regarding the *difficulty* to transform a class to another.

- **Measures based on class features analysis**. This approach uses the graph to extract features of classes. These features will next be analysed to estimate the similarity as a function of shared and distinct features of the compared classes. This approach is conceptually framed in the feature model. The diversity of feature-based measures relies on the fact that various strategies proposed to characterize class features and to take advantage of them to assess the similarity.

- **Measures based on Information Theory**. Based on a function used to estimate the amount of information carried by a class, i.e. its information content (IC), these measures assess the similarity according to the evaluation of the amount of information which is shared and distinct between the compared classes. This approach is framed in information theory; it can however be seen as a derivative of the feature-based approach in which features are not appreciated using a binary feature-matching evaluation (shared/not shared), but incorporate also their saliency, i.e. the degree of informativeness.

- **Hybrid measures**. Measures which are based on multiple paradigms.

The broad classification of measures we propose is interesting to introduce the basic approaches defined to assess the similarity of two classes, and to put them in perspectives with the models of similarity proposed by cognitive sciences. It's however challenging to constraint the diversity of measures to this broad classification. It's important to understand that these three main approaches are highly interlinked and cannot be seen as disjoint categories. As an example, all measures rely on the analysis of the structure of the ontology as they all take advantage of the partial ordering defined by the (structure of the) taxonomy. These categories must be seen as devices used by SMs designers to introduce approaches and highlight relationships between several proposals. Indeed, as we will see, numerous approaches can be regarded as hybrid measures which take advantage of techniques and paradigms used to characterize measures of a specific approach. Therefore, the affiliation of a specific measure to a particular category is often subject to debate, e.g., (Batet 2011a). This can be partially explained by the fact that several measures can be redefined or approximated using reformulations, in a way that further challenge the classification. Indeed, the more you analyse SMs, harder it is to constraint them to specific boxes; the analogy can be made with the relationship between the cognitive models of similarity[i].

Several classifications of measures have been proposed. The most common one is to distinguish measures according to the elements of the graph they take into account (Pesquita, Faria, et al. 2009). This classification distinguishes three approaches: (i) edge-based, measures focusing on relationship analysis, (ii) node-based, measures based on node analysis and (iii) hybrid measures, measures which mix both approaches. In the literature, edge-based measures often refer to structural measures, node-based measures refer to measures framed in the feature-model and those based on information theory. Hybrid measures are those which implicitly or explicitly mix several paradigms.

Another interesting way to classify measures is to study if they are (i) intentional, i.e. based on the explicit definition of the classes expressed by the taxonomy, (ii) extensional, i.e., based on the analysis of the realizations of the classes (i.e., instances), or (iii) hybrid, measures which mix both intentional and extensional information about classes[ii].

In some cases, authors will mix several types of classifications to present measures. In this section we will introduce the measures according to the four approaches presented above: (i) structural, (ii) feature-based, (iii) framed in information theory and (iv) hybrid. We will also specify the extensional, intentional, or hybrid nature of the vision adopted in the design of the measures.

Numerous class-to-class measures have been defined for trees, i.e. special graphs without multiple inheritances. In the literature, these measures are generally considered to be applied out-of-the-box on graphs. However, in the context of graphs, some adaptations deserve to be made and several components of the measures generally need to be redefined in order to avoid ambiguity, e.g., to be implemented in a computer software. For the sake of clarity, we first highlight the diversity of proposals by introducing the most representative measures defined according to the different approaches. Measures will most of the time be presented according to their original definitions. When the measures have been defined for trees, we will not stress the modifications which must be taken into account for them to be used on graphs. These modifications will be discussed after the introduction of the diversity of measures.

---

[i] We invite the reader to refer to the dedicated section 2.2.5 and more particularly to the efforts made for the unification of the various models.

[ii] French reader can refer to (Gandon et al. 2005; Aimé 2011) for examples of such classifications.

### 5.5.1 Structural Approach

Structural measures rely on the graph-traversal approaches presented in subsection 5.4.1.1 (e.g., shortest path techniques, random walk approaches). They focus on the analysis of the interconnection between classes to capture their similarity. However, they most of the time consider specific tuning in order to take into account specific properties and interpretations induced by the transitivity of the taxonomical relationships. In this context, some authors, e.g., (Hliaoutakis 2005), have linked this approach to *spreading activation theory* (Collins & Loftus 1975). The similarity is in this case seen as a function of propagation between classes through the graph.

Back in the eighties, (Rada et al. 1989) expressed the taxonomical distance of two classes defined in a taxonomic tree as a function of the shortest path linking them[i]. We denote $sp(u, isa^*, v)$ the shortest path between two classes $u$ and $v$, i.e., the path of minimal length in $\{u, isa^*, v\}$. Remind that the length of a path has been defined as the sum of the weights associated to the edges which compose the path. When the edges are not weighted we refer to the edge-counting strategy and the length of the shortest path is the number of edges it contains. The taxonomical distance is therefore defined by[ii]:

$$dist_{Rada}(u, v) = sp(u, subof^*, v)$$

with $subof$ the name given to the predicate `subClassOf`.

In a tree, the shortest path $sp(u, isa^*, v)$ contains a unique common ancestor of $u$ and $v$. This common ancestor is the Least Common Ancestor (LCA)[iii] of the two classes according to any function $\theta$ (since the $\theta$ function is monotonically decreasing)[iv].

Note that distance-to-similarity conversions can also be applied to express a similarity from a distance (see Appendix 5). A semantic similarity can therefore be defined in a straightforward manner:

$$sim_{Rada}(u, v) = \frac{1}{dist_{Rada}(u, v) + 1}$$

Notice the importance to consider the transitive reduction of the tree/graph to obtain coherent results using shortest path based measures – in the following presentation, we consider that the taxonomy do not contains redundant relationships.

Several critics of the shortest path techniques have been formulated. The edge-counting strategy or more generally any shortest path approach with uniform edge weight, have been criticized for the fact that the distance represented by an edge linking two classes do not take into account of class specificities/salience[v]. Several modifications have therefore been proposed to break this constraining uniform appreciation of edges induced by the edge-counting strategy. Implicit or

---

[i] It's worth to note that they didn't invented the notion of shortest path in a graph. In addition, in (Foo et al. 1992), the authors refers to a measure proposed by Gardner and Tsui (1987) to compare concepts defined in a conceptual graph using the shortest path technique.

[ii] In this section, equations named *dist* refer to taxonomical distances.

[iii] The Least Common Ancestor is also denoted the Last Common Ancestor (LCA), the Least Common Subsumer/Superconcept (LCS) or Lowest SUPER-ordinate (LSuper).

[iv] Here rely the importance of applying the transitive reduction of the taxonomical graph/tree, redundant taxonomic relationships can challenge this statement and therefore heavily impact the semantics of the results.

[v] As an example, (Foo et al. 1992) quotes remarks made in Sowa personal communication.

explicit models defining non-uniform strength of association between classes, have therefore been introduced e.g., (Young Whan & Kim 1990; Sussna 1993; Richardson et al. 1994) – they have been introduced in section 5.3.3.

One of the main challenge of SM designers over the years have therefore been to implicitly or explicitly take advantage of semantic evidences regarding the expressiveness of classes and the strength of connotation between classes in the design of measures. The different strategies and factors used to appreciate class specificity as well as strength of connotations have already been introduced in section 5.3. Another use of the various semantic evidences which can be extracted from $G_T$ has been to normalize the measures. As an example, (Resnik 1995) proposed to consider the maximal depth of the taxonomy to bound the edge-counting strategy:

$$sim_{Resnik-eb}(u,v) = 2*max\_depth - sp(u, subof, LCA(u,v)) - sp(v, subof, LCA(u,v))$$

To simulate non uniform edge weighting (Leacock & Chodorow 1998)[i] introduced a log trans-form consideration of the edge counting strategy:

$$sim_{LC}(u,v) = -\log\left(\frac{N}{2*\max\_depth}\right) = \log(2*\max\_depth) - \log(N)$$

with $N$ the cardinality of the union of the sets of nodes involved in the paths $sp(u, subof, LCA(u,v))$ and $sp(v, subof, LCA(u,v))$.

Authors have also proposed to take into account the specificity of compared classes, e.g., (Mao & Chu 2002), sometimes as a function of the depth of their LCA, e.g., (Wu & Palmer 1994; Pekar & Staab 2002; J. Wang et al. 2012).

As an example, the strategy proposed by (Wu & Palmer 1994) was to express the similarity of two classes as a ratio taking into account the shortest path linking the classes as well as the depth of their LCA.

$$sim_{WP}(u,v) = \frac{2\,depth(LCA(u,v))}{2\,depth(LCA(u,v)) + sp(u, subof, LCA(u,v)) + sp(v, subof, LCA(u,v))}$$

This function is of the form:

$$f(x,y,z) = \frac{x}{x + (y+z)/2}$$

with $x$ the depth of the LCA of the two classes $u, v$ and $y + z$ the length of the shortest path linking $u, v$, it is easy to see that for any given non-null length of the shortest path, this function is increasing with $x$; otherwise stated, to a given shortest path length, the similarity of $u, v$ increases with the depth of their LCA. In addition, as expected, for a given depth of the LCA, the more the length of the shortest path linking $u, v$ increases, less similar they will be considered.

Based on a specific expression of the notion of depth, a parameterized expression of $sim_{WP}$ has been proposed in (Wang & Hirst 2011). A variation has also been proposed by (Pekar & Staab 2002):

---

[i] Note that according to (Resnik 1995), this approach was already proposed in an 1994 unpublished paper of the same authors (Leacock & Chodorow 1994).

$$sim_{PS}(u,v) = \frac{depth(LCA(u,v))}{sp(u, subof, LCA(u,v)) + sp(v, subof, LCA(u,v)) + depth(LCA(u,v))}$$

(Zhong et al. 2002) also proposed to compare classes taking into account the notion of depth:

$$dist_{Zhong}(u,v) = 2 \frac{1}{2k^{depth(LCA(u,v))}} - \frac{1}{2k^{depth(u)}} - \frac{1}{2k^{depth(v)}}$$

with $k > 1$ a factor defining the contribution of the depth.

In a similar fashion, (Li et al. 2003; Li et al. 2006) defined a parametric function in which both the maximal depth and the length of the shortest path are taken into account:

$$sim_{LB}(u,v) = e^{-\alpha\, dist_{Rada}(u,v)} \times df(u,v)$$

with,

$$df(u,v) = \frac{e^{\beta h} - e^{-\beta h}}{e^{\beta h} + e^{-\beta h}}$$

The parameter $h$ corresponds to the depth of the LCA of the compared classes, i.e. $h = depth(LCA(u,v))$. The parameter $\beta > 0$ is used to tune the depth factor (df) and to set the importance to give to the degree of specificity. The function used to express df corresponds to the hyperbolic tangent which is normalized between 0 and 1. It defines the degree of nonlinearity to associate to the depth of the LCA. In addition $\alpha \geq 0$ controls the importance of the taxonomic distance expressed as a function of the length of the shortest path linking the two classes.

Approaches have also been proposed to modify specific measures in order to obtain particular properties. As an example, (Slimani et al. 2006) proposed an adaptation of Wu & Palmer measure to avoid the fact that in some cases neighbour classes can be estimated as more similar than classes defined in the same hierarchy. To this end, the authors introduced $sim_{tbk}$ which is based on a factor used to penalized classes defined in the neighbourhood:

$$sim_{tbk}(u,v) = sim_{WP}(u,v) \times pf(u,v)$$

with,

$$pf(u,v) = (1 - \lambda)\ (\min(depth(u), depth(v)) - \max\_depth)$$
$$+ \lambda(depth(u) + depth(v) + 1)^{-1}$$

In the same vain (Ganesan et al. 2012; Shenoy et al. 2012) recently proposed alternative measures answering the same problem. The approach proposed by (Shenoy et al. 2012) is presented[i]:

$$sim_{Shenoy}(u,v) = \frac{2 \max\_depth\ e^{-\lambda L/\max\_depth}}{depth(u) + depth(v)}$$

---

[i] Note that we assume that the paper contains an error in the equation defining the measure. The formula is considered to be X / (Y+Z), not X/Y +Z as written in the paper.

with $L$ the weight of the shortest path computed by penalizing paths with multiple changes of type of relationships, e.g. a path following the pattern $<isa, isa^-, isa, ... >$. Note that the penalization of paths inducing complex semantics, e.g. which involves multiple type of relationships, was already introduced in (Hirst & St-Onge 1998; Bulskov et al. 2002).

Several approaches have also been proposed to consider density of classes, e.g., through analysis of cluster of classes (Al-Mubaid & Nguyen 2006). Other adaptations also proposed to take into account classes distance to leaves (Wu et al. 2006) and variable strengths of connotation considering particular strategies (Lee et al. 1993; Zhong et al. 2002), e.g. using IC variability among two linked classes or multiple topological criteria (Jiang & Conrath 1997; Couto et al. 2003; M. Alvarez et al. 2011).

In the vein of the spreading activation theory, measures have also been defined as a function of transfer between the compared classes (Schickel-Zuber & Faltings 2007). (Wang et al. 2007) use a similar approach based on a specific definition of the strength of connotation. Finally, pure graph-based approaches defined for the comparison of nodes can also be used to compare classes defined in a taxonomy (refer to section 3.4.1.1). As an example, (Garla & Brandt 2012; Yang et al. 2012) used random walk techniques such as the personalized page rank approach to define semantic similarity measures.

As we have seen, most structural semantic similarity measures are extensions or refinements of the intuitive shortest path distance considering intrinsic factors to consider both the specificity of classes and variable strengths of connotations. Nevertheless, the algorithmic complexity of the shortest path algorithms hampers the suitability of these measures for large semantic graphs. To remedy this problem, we have seen that shortest path computation is replaced by approximation based on the depth of the LCA of the compared classes[i] and that several measures proposed by graph theory can be used instead.

### 5.5.1.1 Towards Other Estimators of Semantic Similarity

Most of critics relative to the initial edge-counting approach were related to the uniform consideration of edge weights. As we have seen, several authors proposed to consider various semantic evidences to differentiate strengths of connotation between classes.

One of the central findings conveyed by the early developments in structure-based measures is that the similarity function can be decomposed in several components, in particular those distinguished by the feature model: commonality and difference. Indeed, the shortest path between two classes can be seen as the difference between the two classes, considering that all specialization add properties to a class. More particularly, in trees, or under specific constraints in graphs, we have seen that the shortest path linking two classes contains their LCA. The shortest path can therefore be break down in two parts corresponding to the shortest path linking the compared classes to their LCA. In this case, the LCA can thus be seen as a proxy who partially[ii] summarizes

---

[i] The algorithmic complexity of the LCA computation is significantly lower than the computation of the shortest path.

[ii] The LCA can indeed only be an upper-bound of the commonality since highly similar classes (`Man`, `Women`) may have for LCA a general class which only encompasses a limited amount of their commonalities (e.g., `LivingBeing`). Please refer to section 5.3

the commonality of the compared classes. The *distance* between the compared classes and their LCA can therefore be used to estimate the differences between the classes.

The fact that measures can be break down in specific components evaluating commonalities and differences is central in the design of the approaches we will further introduce:

- *The Feature-based strategy*.
- *The Information Theoretical strategy*.

These approaches will focus on characterizing the compared classes in order to express measures as a function of their commonalities and differences.

### 5.5.2 Feature-based Approach

*The Feature-based* approach generally refers to measures relying on a taxonomical interpretation of the feature model proposed by Tversky (Tversky 1977). However, as we will see, contrary to the original definition of the feature model, this approach is not necessarily framed in set theory[i]. The main idea is to represent classes as collections of features, i.e., characteristics describing the classes, to further express measures based on the analysis of the common and distinct features of the compared classes. The score of the measures will only be influenced by the strategy adopted to choose the features of the classes[ii] and the strategy adopted to compare them.

As we will see, the reduction of classes to collections of features makes it possible to set the semantic similarity back in the context of classical binary similarity or distance measures (e.g., set-based measures).

An approach commonly used to represent the features of a class is to consider its ancestors as features[iii]. We denote $A(u)$ the set of ancestors of the class $u$. Since Jaccard index, that was proposed 100 years ago, numerous binary measures have been defined in various fields. A survey of these measures distinguishes 76 of them (Choi et al. 2010). Considering that the features of a class $u$ are defined by $A(u)$, an example of semantic similarity measure expressed from the Jaccard index was proposed in (Maedche & Staab 2001)[iv]:

$$sim_{CMatch}(u,v) = \frac{|A(u) \cap A(v)|}{|A(u) \cup A(v)|}$$

---

[i] Recall that the feature matching function on which is based the feature model relies on binary evaluations of the features "*In the present theory, the assessment of similarity is described as a feature-matching process. It is formulated, therefore, in terms of the set-theoretical notion of a matching function rather than in terms of the geometric concept of distance*" (Tversky & Itamar 1978).

[ii] As stressed in (Schickel-Zuber & Faltings 2007), there is a narrow link with the multi-attribute utility theory (Keeney 1993) in which the utility of an item is a function of the preference on the attributes of the item.

[iii] Implicit senses if we consider that compared elements are not classes but concepts.

[iv] This is actually a component of a more refined measure.

Another example of a set-based expression of a feature-based approach is the measure defined by (Bulskov et al. 2002):

$$Sim_{Bulskov}(u,v) = \alpha \frac{|A(u) \cup A(v)|}{|A(u)|} + (1-\alpha) \frac{|A(u) \cup A(v)|}{|A(v)|}$$

with $\alpha \in [0,1]$ a parameter used to tune the symmetry of the measure.

(Rodríguez & Egenhofer 2003) proposed a formulation derived from de ratio model defined by Tversky (introduced in section 2.2.2):

$$sim_{RE}(u,v) = \frac{|A(u) \cap A(v)|}{\gamma \cdot |A(u) \backslash A(v)| + (1-\gamma) \cdot |A(v) \backslash A(u)| + |A(u) \cap A(v)|}$$

with $\gamma \in [0,1]$, a parameter that enables to tune the symmetry of the measure.

(Sánchez, Batet, et al. 2012) define the taxonomic distance of two classes as a function of the ratio between their distinct and shared features:

$$dist_{Sanchez}(u,v) = \log_2 \left( 1 + \frac{|A(u) \backslash A(v)| + |A(v) \backslash A(u)|}{|A(u) \backslash A(v)| + |A(v) \backslash A(u)| + |A(u) \cap A(v)|} \right)$$

Various refinements of these measures have been proposed to enrich class features taking into account their descendants, e.g., (Ranwez et al. 2006).

The feature-based measures have not to be intentional, i.e., they are not expected to solely rely on the knowledge defined in the taxonomy. When the instances of the classes are known, the feature of a class can also be seen by extension and be defined on the basis of the instances associated to the classes. As an example, the Jaccard index could also be used to compare two classes which are ordered according to their shared and distinct features, here characterized by extension:

$$sim_{Jaccard}(u,v) = \frac{|I(u) \cap I(v)|}{|I(u) \cup I(v)|}$$

with $I(u) \subseteq I$, the set of instances of the class $u$. Note that this approach makes no sense if the will is to compare classes not ordered since the set $\{I(u) \cap I(v)\}$ will tend to be empty.

(D'Amato et al. 2008b) also define an extensional measures considering:

$$sim_{d'Amato}(u,v) = \frac{\min(|I(u)|,|I(v)|)}{|I(LCA(u,v))|} \left(1 - \frac{|I(LCA(u,v))|}{|I|}\right) \left(1 - \frac{\min(|I(u)|,|I(v)|)}{|I(LCA(u,v))|}\right)$$

The measures presented above summarize the features of a class through a set representation corresponding to the set of classes or instances. However, alternative approaches can also be explored. Therefore, even if to our knowledge such approaches have not been explored, the features of a class could be represented as a set of relationships, as a subgraph, etc.

In addition, regardless of the strategy adopted to characterize the features of a class (other classes, relationships, instances), the comparison of the features is not necessarily driven by a set-based measure. Indeed, the collections of features can also be seen as vectors. As an example, a class $u$ can be represented by a vector $U$ in a chosen real space of dimension $|C|$, e.g., considering that each dimension associated to an ancestor of $u$ is set to 1. Vector-based measures will evaluate the distance of two classes by studying the coordinates of their respective projections.

(Bodenreider et al. 2005) proposed to compare two classes according to their representation through the Vector Space Model (VSM). Considering a class to instance matrix, a weight corresponding to the inverse document frequency is associated to the cell $(u, i)$ of the matrix if the instance $i$ is associated to the class $u$, i.e., $i \in I(u)$. The vectors representing two classes are next compared using the dot product.

### 5.5.3 Information Theoretical Approach

*The Information Theoretical* approach relies on Shannon's Information theory. Like for the feature-based strategy, Information Theoretical measure relies on the comparison of two classes according to their commonalities and differences, here defined in terms of information. This approach formally introduces the notion of salience of classes through the definition of their informativeness – Information Content (IC)[i].

(Resnik 1995) defines the similarity of a couple of classes as the IC of their common ancestor maximizing an IC function (originally *eIC*), i.e., their most informative common ancestor (MICA).

$$sim_{Resnik}(u, v) = IC(MICA(u, v))$$

Resnik's measure doesn't explicitly capture the specificities of compared classes. Indeed, couples of classes with an equivalent MICA will have the same semantic similarity, whatever their respective ICs are. To correct this limitation, several authors refined the measure proposed by Resnik to incorporate the specificities of compared classes; we here present the measures proposed by (Lin 1998)[ii] - $sim_{Lin}$ , (Jiang & Conrath 1997) - $dist_{JC}$, (Mazandu & Mulder 2013) - $sim_{Nunivers}$, (Pirró & Seco 2008; Pirró 2009) - $sim_{PSec}$ and (Pirró & Euzenat 2010a) - $sim_{Faith}$:

$$sim_{Lin}(u, v) = \frac{2\,IC(MICA(u, v))}{IC(u) + IC(v)}$$

$$sim_{Nunivers}(u, v) = \frac{IC(MICA(u, v))}{\max(IC(u), IC(v))}$$

$$dist_{JC}(u, v) = IC(u) + IC(v) - 2\,IC(MICA(u, v))$$

---

[i] Section 5.3.1.2 introduces the notion of information content.

[ii] The measure proposed is a redefinition commonly admitted of the original measure: $sim_{Lin}(u, v) = \frac{2 \times \log(p(MICA(u,v)))}{\log(p(u))\log(p(v))}$

$$sim_{PSec}(u,v) = 3IC\big(MICA(u,v)\big) - IC(u) - IC(v)$$

$$sim_{Faith}(u,v) = \frac{IC(MICA(u,v))}{IC(u) + IC(v) - IC(MICA(u,v))}$$

Taking into account specificities of compared classes can lead to high similarities (low distances) when comparing general classes. As an example, when comparing general classes ing $sim_{Lin}$, the maximal similarity will be obtained comparing a (general) class to itself. In fact, the identity of the indiscernible is generally ensured (except for the root which generally has an IC equal to 0). However, some treatments require this property not to be respected. Authors have therefore proposed to lower the similarity of two classes according to the specificity of their MICA, e.g. (Schlicker et al. 2006; Li et al. 2010). The measure proposed by (Schlicker et al. 2006) is presented:

$$sim_{Rel}(u,v) = Sim_{Lin}(u,v) \times (1 - p(MICA(u,v)))$$

with $p(MICA(u,v))$ the probability of occurrence of the MICA. An alternative approach proposed by (Li et al. 2010) relies on the IC of the MICA and can therefore be used without extensional information of the classes using an intrinsic expression of the IC.

Authors have also proposed to characterize the information carried by a class summing the IC of their super classes (Gaston K. & Nicola J. 2011; Cross & Yu 2011):

$$sim_{Mazandu}(u,v) = \frac{2 \sum_{c \in A(u) \cap A(v)} IC(c)}{\sum_{c \in A(u)} IC(c) + \sum_{c \in A(v)} IC(c)}$$

$$sim_{JacAnc}(u,v) = \frac{\sum_{c \in A(u) \cap A(v)} IC(c)}{\sum_{c \in A(u) \cup A(v)} IC(c)}$$

These measures can also be considered as hybrid strategies between the feature-based and the information theory approaches. One can consider that these measures rely on a redefinition of the way to characterize the information conveyed by a class (by summing the IC of the ancestors). Other interpretations can simply consider that features are weighted. Thus, following the set-based representations of features, authors have also studied these measures as fuzzy measures (Cross 2004; Cross 2006; Cross & Sun 2007; Cross & Yu 2010; Cross & Yu 2011), e.g. defining the membership function of a feature corresponding to a class as a function of its IC.

Finally, other measures based on information theory have also been proposed, e.g., (Maguitman & Menczer 2005; Maguitman et al. 2006; Cazzanti & Gupta 2006). As an example in (Maguitman & Menczer 2005) the similarity is estimated as a function of prior and posterior probability regarding instances and class membership.

### 5.5.4 Hybrid Approach

Other techniques take advantage of the various paradigms introduced in the previous section. Among the numerous proposals, (Jiang & Conrath 1997; Bin et al. 2009) defined measures in which density, depth, strength of connotation and information content of classes are taken into account. We present the measure proposed by (Jiang & Conrath 1997) [i], which considers the strength of association defined as follow:

$$w(u,v) = \left( \beta + (1-\beta)\, \frac{\overline{dens}}{|children(v)|} \right) \left( \frac{depth(v)+1}{depth(v)} \right)^{\alpha} (IC(u) - IC(v))\, T(u,v)$$

The factor $\overline{dens}$ refers to the average density of the whole taxonomy (see publication for details). The factors $\alpha \geq 0$ and $\beta \in [0;1]$ control the importance of the density factor and the depth respectively. $T(u,v)$ defines the weight associated to the predicates. Finally, the similarity is estimated as the weight of the path which links the compared classes which are constrained by their LCA:

$$dist_{JC-Hybrid}(u,v) = \sum_{(s,p,o) \in sp(u,subOf,LCA(u,v)) \cup sp(v,subOf,LCA(u,v))} w(s,o)$$

Note that defining $\alpha = 0$, $\beta = 1$ and $T(u,v) = 1$, we obtain the information theoretical measure defined by the same authors:

$$dist_{JC}(u,v) = IC(u) + IC(v) - 2\, IC(MICA(u,v))$$

(Singh et al. 2013) proposed a mixing strategy based on (Jiang & Conrath 1997) IC-based measure and the consideration of transition probability between classes relying on a depth-based estimation of the strength of connotation.

(Rodríguez & Egenhofer 2003) also proposed to mix a feature-based approach also considering structural properties such as the depth of the classes. (Paul et al. 2012) also proposed multiple measures based on a mixing strategy by taking advantage of several exiting measures.

---

[i] This measure is a parametric distance. (Couto et al. 2003) discuss the implementation, (Othman et al. 2008) propose a genetic algorithm which can be used to tune the parameters and (Wang & Hirst 2011) propose a redefinition of the notion of depth and density initially proposed.

### 5.5.5 Considerations for the Comparison of Classes defined in Semantic Graphs

Several measures introduced in the previous sections were initially defined to compare classes expressed in a tree. However, several considerations must be taken into account in order to estimate the similarity of classes defined in a semantic graph. The content of this section is quite technical and may therefore not be suited for all publics – please refer to the notations introduced in section 5.2.

#### 5.5.5.1 Shortest path

A tree is a specific type of graph in which multi-inheritances cannot be encountered, which therefore implies that two classes which are not ordered will have no common subclasses, i.e., $G_u^- \cap G_v^- = \emptyset$. Therefore, if there is no redundant taxonomical relationship, then, the shortest path linking two classes defined in a tree always contains a single common ancestor of the two classes. However, in a graph, since two non-ordered classes can have common subclasses, i.e., $G_u^- \cap G_v^- \neq \emptyset$, the shortest path linking two classes can in some cases not contain one of their common ancestor. Figure 15 illustrates the modifications induced by multi-inheritances.

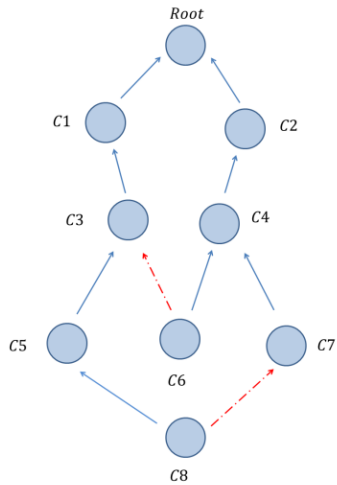

Figure 15 The graph composed of the plain (blue) edges is a taxonomic tree, i.e., it doesn't contain classes with multi-inheritances. If the (red) dotted relationships are also considered, the graph is a directed acyclic graph (e.g., a taxonomical graph).

In Figure 15, the shortest path linking the two non-ordered classes C5 and C7 in the tree (i.e. without considering the red edges) is [C5-C3-C1-root-C2-C4-C7]. However, if we consider multiple inheritances (red edges), it's possible to link the two classes through paths which do not contain one of their common superclass, e.g., [C5-C3-C6-C4-C7] or even [C5-C8-C7]. Therefore, the shortest path containing a common ancestor of the compared classes is defined in the search space $G_u^+ \cup G_v^+$. In practice, despite the fact that in most graphs $G_u^- \cap G_v^- \neq \emptyset$ for two non-ordered classes, it is commonly admitted that the shortest path must contain a single superclass of the two compared classes. Given this constraint, the edge-counting taxonomical distance of $u$ and $v$ in $G_u^+ \cup G_v^+$ is generally (implicitly[i]) defined by: $distTax_{SP}(u,v) = sp(u, subof, LCA(u,v)) + sp(v, subof, LCA(u,v))$.

Note that when disjoint common ancestors are shared between compared classes, the ancestor which maximizes the similarity is expected to be considered. Depending on the $\theta$ function which is used, the shortest path doesn't necessarily involve the class of the NCCAs which maximize $\theta$, e.g. the deeper. As an example, to distinguish the DCA to consider, (Schickel-Zuber & Faltings 2007) took into account a mix between depth and reinforcement (number of different paths leading from a concept to another).

---

[i] Generalization of measures defined from trees to graph is poorly documented in the literature.

The shortest path techniques can also be relaxed to consider paths which do not involve common ancestors or which involve multiple common ancestors:

$$sim_{SP-R}(u,v) = \frac{1}{sp(u, subof^*, v) + 1}$$

### 5.5.5.2 Notion of Depth

The definition of the notion of depth must also be reconsidered when the taxonomy forms a graph. Recall that, in a tree without redundancies, the depth of a class has been defined as the length of the shortest path linking the class to the root. The depth of a class is a simple example of estimator of its specificity. In a tree, this estimator makes perfect sense since the depth of a class is directly correlated to the number of ancestors it has, as $depth(c) = |A(c)| - 1$.

In a graph or in a tree with redundant taxonomical relationships, we must ensure that the depth is monotonically decreasing according to the ordering of the classes defined by the taxonomy. As an example, to apply depth-based measures to graphs, we must generally ensure that $depth(LCA(u,v))$ is lower or equal to both $depth(u)$ and $depth(v)$. To this end, the maximal depth of a class must be used, i.e., the length of the longest path in $\{u, subof, LCA(u,v)\}$, denoted $lp(u, subof, LCA(u,v))$.

As an example, the measure proposed by (Pekar & Staab 2002) is therefore implicitly generalized to:

$$sim_{WP}(u,v) = \frac{depth(LCA(u,v))}{lp(u, subof, LCA(u,v)) + lp(v, subof, LCA(u,v)) + depth(LCA(u,v))}$$

### 5.5.5.1 Notion of Least Common Ancestors

Most measures which have been presented take advantage of the notions of LCA or MICA of the compared classes. However, these measures do not consider disjoint semantic contributions, i.e., the set of common ancestors of the compared classes - $\Omega(u,v)$ for the classes $u$ and $v$. To remedy this, several authors have proposed to consider the whole set of DCAs in the design of measures. (Couto et al. 2005; Couto & Silva 2011) proposed GraSM and DiShIn strategies to consider the set of DCAs.

(Couto et al. 2005) proposed to modify information theoretical measures based on the notion of MICA. The authors recommended replacing the IC of the MICA by the average of the information contents of all the classes which compose the set of DCAs of the compared classes. A redefinition of the measure proposed by Lin is presented:

$$sim_{Lin-GrasM}(u,v) = \frac{\frac{\sum_{c \in \Omega(u,v)} IC(c)}{|\Omega(u,v)|}}{IC(u) + IC(v)}$$

$$sim_{Lin-GrasM}(u,v) = \frac{\sum_{c \in \Omega(u,v)} IC(c)}{|\Omega(u,v)| \times (IC(u) + IC(v))}$$

(J. Wang et al. 2012) also proposed to average the similarity between the classes according to their multiple DCAs:

$$sim_{Wang}(u,v) = \frac{\sum_{a \in \Omega(u,v)} \frac{2\, depth(a)^2}{d_a(root,u) \times d_a(root,v)}}{|\Omega(u,v)|}$$

With $d_a(root,u)$ the average length of the set of paths which contain the class $a$ and which link the class $u$ to the root of the taxonomy and $\Omega(u,v)$ the set of NCCAs of the classes $u$ and $v$.

As we have underlined, numerous approaches have been defined to compare pairs of classes defined in a taxonomy, these measures can be used to compare any pair of nodes defined in a partially ordered set. **Table 4** to **Table 7** present some properties of a selection of measures defined to compare classes.

### 5.5.6 List of Pairwise Semantic Similarity Measures

Several SMs which can be used to compare classes defined in a taxonomy of classes or any pair of elements defined in a poset. Measures are ordered according to their date of publication. Other contributions studying some properties of pairwise measures can be found in (X. Yu 2010; Slimani 2013). IOI: Identify of the Indiscernibles.

| Structural Measures | | | | | |
|---|---|---|---|---|---|
| Name | Type | KR const. | Range | IOI | Comment |
| Shortest Path | Sim / Rel | None | $\mathbb{R}^+$ | Yes | Weight of the shortest path (sp) linking the compared classes. Several modifications can be considered in graphs depending on the strategy adopted, e.g., weighting of the relationships, predicates, constraints on the inclusion of a common super class of the compared classes, etc. |
| (Rada et al. 1989) | Dist (ISA) | DAG | $\mathbb{R}^+$ | Yes | Specific shortest path strategy with uniform weight and the shortest path is constrained to contain the LCA of the compared classes. |
| (Young Whan & Kim 1990) | | | | | |
| (Lee et al. 1993) | | | | | |
| (Sussna 1993) | Dist | RDAG | $\mathbb{R}^+$ | Yes | Originally defined as a parametric semantic relatedness. Under specific constraints, this measure can be used as a semantic similarity. Shortest path technique taking into account non-uniform strength of connotation tuned according to the depth of the compared classes and specific weight associated to predicates. |
| (Richardson et al. 1994) | X | X | X | X | Propose to integrate several intrinsic metrics (e.g., depth, density) to weight the relationships and define hybrid measures mixing the structural and information theoretical approach. No measure explicitly defined. |
| (Wu & Palmer 1994) | Sim | RDAG | [0,1] | Yes | Similarity assessed according to the depth of the compared classes and the depth of their LCA. |
| (Leacock & Chodorow 1994; Leacock & Chodorow 1998) | Sim | RDAG | $\mathbb{R}^+$ | Yes | (Rada et al. 1989) formulation penalizing long shortest path between the compared classes according to the depth of the taxonomy. |
| (Resnik 1995) | Sim | RDAG | [0,2$D$] | Yes | Similarity based on the shortest path technique which has been bounded by (twice) the max depth of the taxonomy (D). |

| (Hirst & St-Onge 1998) | Sim / Rel | None | $\mathbb{R}^+$ | Yes | Shortest path penalizing multiple changes of predicate. Can be used as a similarity or relatedness measure depending on the relationships considered. |
|---|---|---|---|---|---|
| (Zhong et al. 2002) | Dist | RDAG | [0,max[ | Yes | Taxonomical distance taking into account the depth of the compared classes. With max defined as $max = 2\frac{1}{2k^{depth(LCA(u,v))}}$ with k a given constant. |
| (Pekar & Staab 2002) | Sim | RDAG | [0; 1] | Yes | Shortest path technique which take into account the depth of the LCA of the compared classes. |
| (Mao & Chu 2002) | Sim | DAG | $\mathbb{R}^+$ | No | Modification of Rada's measure taking into account class specificity as a non-linear function of the number of descendants a class has. |
| (Li et al. 2003) (Li et al. 2006) | Sim | RDAG | $\mathbb{R}^+$ | No | Measure considering both the length of the shortest path linking the compared classes and their depth. |
| (Ganesan et al. 2003) | Sim | | | | Refer to leafsim |
| (Yu et al. 2005) | Sim | RDAG | [0; 1] | Yes | Measure allowing non-null similarity only to ordered pair of classes. |
| (Wu et al. 2006) | Sim | RDAG | [0; 1] | No | Take into account compared classes (i) commonality (length of the longest shared path from the classes to the root), (ii) specificity (defined as a function of the shortest path from the class to leaves it subsumes) and (iii) local distance (Rada distance). |
| (Slimani et al. 2006) | Sim | RDAG | [0; 1] | Yes | Modification of the Wu and Palmer measure to avoid neighbours classes to have higher similarity values that classes which are ordered. |
| (Blanchard et al. 2006) | Sim | RDAG | [0,1] | No | Expression based on a specific expression of an abstract formulation of the Dice coefficient defined for trees and extended for DAG in (Blanchard 2008)[i] |
| (Nagar & Al-Mubaid 2008) | Sim | DAG | $\mathbb{R}^+$ | Yes | Use a modification of shortest path constrained by the LCA. |
| (Cho et al. 2003) | Sim | DAG | $\mathbb{R}^+$ | No | Multiple factors are considered to take into account the specificity of the compared classes. |
| (Alvarez & Yan 2011) | Sim / Rel | RDAG | [0; 1] | No | SSA, exploits three components evaluating the shortest path linking the compared classes (a weighting scheme is applied to the graph), their LCA, and their literal definitions. |
| (J. Wang et al. 2012) | Sim | RDAG | [0; 1] | Yes | Approach taking into account the depth of the compared classes, as well as the depth of all their DCAs. |
| (Shenoy et al. 2012) | Sim | RDAG | | | Modification of the Wu and Palmer measure to avoid neighbours classes to have higher similarity values that classes which are ordered. |
| (Ganesan et al. 2012) | Sim | RDAG | | | Modification of the Wu and Palmer measure to avoid neighbours classes to have higher similarity values that classes which are ordered. |

[i] In french

**Table 4**: Semantic similarity measures or taxonomical distances defined using an structural approach. These measures can be used to compare a pair of classes defined in a taxonomy or any pair of elements defined in a partial ordered set.

| Information Theoretical Approach | | | | | |
|---|---|---|---|---|---|
| Name | Type | KR const. | Range | IOI | Comment |
| (Resnik 1995) | Sim | DAG | $[0;1]$ $[0;inf]$ | No | Similarity considered as the IC of concepts' MICA. The range depends on the IC. |
| (Jiang & Conrath 1997) | Dist | DAG | $[0;1]$ | Yes | Taxonomic distance computed as a function of the IC of the compared classes and their MICA. |
| (Lin 1998) | Sim | DAG | $[0;1]$ | Yes | Similarity computed as a ratio between the IC of the MICA of the compared classes and the sum of the ICs of the compared classes. |
| (Schlicker et al. 2006) (Li et al. 2010) | Sim | DAG | $[0;1]$ | No | Lin measure modified to take specificity into account, i.e. to avoid high score of similarity comparing two general classes. |
| (Couto et al. 2007) | Sim | DAG | $[0;1]$ | No | Lin measure in which all the DCAs of the compared classes are taken into account. |
| (Yu, Jansen, Stolovitzky, et al. 2007) | Sim | DAG | $[0;inf]$ | No | Total Ancestry Measure (TAM) considering the Yu el al. definition of the LCA. |
| (Pirró & Seco 2008; Pirró 2009) | Sim | DAG | $[0;x]$ | No | With $x$ the maximal class IC. |
| (Pirró & Euzenat 2010a) | Sim | DAG | $[0;1]$ | Yes | Information Theoretical expression of the Jaccard Index. |
| (Jain & Bader 2010) | Sim | DAG | $[0;1]$ | No | Build a meta-graph reducing the original ontology into cluster of related concepts. Similarity is assessed through a specific function evaluating LCA information content. |
| (Gaston K. & Nicola J. 2011) simDIC | Sim | DAG | $[0;1]$ | Yes | Specific formulation of a set-based measure considering classes as their sets of ancestors. |
| (Mazandu & Mulder 2013) Sim Nuniver | Sim | DAG | $[0;1]$ | Yes | IC of the MICA of the compared classes divided by the maximal IC of the compared classes. |

**Table 5**: Semantic similarity measures or taxonomical distances defined using an information theoretical approach. These measures can be used to compare a pair of classes defined in a taxonomy or any pair of elements defined in a partial ordered set.

| Feature-based Approach | | | | | |
|---|---|---|---|---|---|
| Name | Type | KR const. | Range | IOI | Comment |
| (Stojanovic et al. 2001) (Maedche & Staab 2001) | Sim | DAG | [0; 1] | Yes | Feature-based expression relying on the Jaccard index. |
| (Bodenreider et al. 2005) | Sim | DAG | DAG | [0,1] | Cosine similarity on a vector-based representation of the classes. The vector representation is built according to the set of instances of the classes. |
| (Ranwez et al. 2006) | Dist | DAG | $\mathbb{R}^+$ | Yes | The distance is a function of the number of descendants of the LCA of the compared concepts. Distance properties have been proved (positivity, symmetry, triangle inequality). |
| (Jain & Bader 2010) | Sim | DAG | [0; 1] | No | Build a meta-graph reducing the original ontology into cluster of related concepts. Similarity is assessed through a specific function evaluating LCA information content. |
| (Batet, Sánchez & Valls 2010) | Sim | DAG | $\mathbb{R}^+$ | Yes | Comparison of the classes according to their ancestors. Formulation expressed as a distance converted to a similarity using negative log. |

**Table 6**: Semantic similarity measures or taxonomical distances designed using a feature-based approach. These measures can be used to compare a pair of classes defined in a taxonomy or any pair of elements defined in a partial ordered set.

| Hybrid Approach | | | | | |
|---|---|---|---|---|---|
| Name | Type | G constraint | Range | IOI | Comment |
| (Jiang & Conrath 1997) (Couto et al. 2003) (Othman et al. 2008) | Sim Dist | RDAG | [0; 1] | Var. | Strategy based on the shortest path constrained by the LCA of the compared classes. Relationships are weighted according to the difference of IC of the classes they link. |
| (Al-Mubaid & Nguyen 2006) | Sim | DAG | [0; ∞] | Yes | Assigns cluster(s) to classes. The similarity is computed considering multiple |
| | | | | | metrics. |
| (Alvarez & Yan 2011) | Sim / Rel | RDAG | [0; 1] | No | Exploits three components evaluating classes, their shortest path (a weighting scheme is applied to the graph), their LCA, and their literal definitions. |
| (Wang et al. 2007) | Sim/Rel | RDAG | [0; 1] | Yes | Originally defined as a semantic relatedness, it can be used to compute semantic similarity. Define a non-linear approach to characterize the strength of connotation and the notion of S-value to characterize the informativeness of a class. |
| (Paul et al. 2012) | Sim | | | | Multiple approaches are mixed |

**Table 7**: Semantic similarity measures or taxonomical distances designed using a hybrid approach. These measures can be used to compare a pair of classes defined in a taxonomy or any pair of elements defined in a partial ordered set.

# 5.6 Semantic Similarity between Groups of Classes

Two main approaches are commonly distinguished to introduce semantic similarity measures designed for the comparison of two sets of classes, i.e., *groupwise measures*:

- *Direct approach*, the measures which can be used to directly compare the sets of classes according to information characterizing the sets with regard to the information defined in the graph.
- *Indirect approach* corresponds to the measures which assess the similarity of two sets of classes using a pairwise measure, i.e. a measure designed for the comparison of pairs of classes. They are generally simple aggregations of the scores of similarities associated to the pairs of classes defined in the Cartesian product of the two compared sets.

Once again, a large diversity of measures have been proposed, we distinguish some of them.

### 5.6.1 Direct Approach

The direct approach corresponds to a generalization of the approaches defined for the comparison of pairs of classes in order to compare two sets of classes.

It is worth noting that two sets of classes can be compared using classical set-based approaches. They can also be compared through their vector representations, e.g., using the cosine similarity measure. Nevertheless, these are in most cases not meaningful since these measures do not take into account of the similarity of the elements composing the compared sets[i].

#### 5.6.1.1 Structural approach

Considering $G_X^+$ as the graph induced by the union of the ancestors of the classes which compose the set $X$, (Gentleman 2007) defined the similarity of two sets of classes $(U, V)$ according to the length of the longest $sp(c, subClassOf, root)$ which links the class $c$ to the $root$ in $G_U^+ \cup G_V^+$.

#### 5.6.1.2 Feature-based Approach

The feature-based measures are characterized by the approach adopted to express the features of a set of classes.

Several measures have been proposed from set-based measures. Considering $C(G_X^+)$ as the set of classes contained in $G_X^+$, we introduce SimUI (Gentleman 2007) [ii], and the Normalized Term Overlap measure (NTO):

---

[i] These simple approaches are generally used when the compared sets contain redundant elements (refer to section 5.2.2.5).
[ii] Also published through the name Term Overlap in (Mistry & Pavlidis 2008).

$$sim_{UI}(U,V) = \frac{|C(G_U^+) \cap C(G_V^+)|}{|C(G_U^+) \cup C(G_V^+)|}$$

$$sim_{NTO}(U,V) = \frac{|C(G_U^+) \cap C(G_V^+)|}{\min(|C(G_U^+)|, |C(G_V^+)|)}$$

### 5.6.1.3  Information theoretical measures

Among other, (Pesquita et al. 2007) proposed to consider the information content of the classes (originally $eIC$):

$$Sim_{GIC}(U,V) = \frac{\sum_{u \in C(G_U^+) \cap C(G_V^+)} IC(u)}{\sum_{u \in C(G_U^+) \cup C(G_V^+)} IC(u)}$$

## 5.6.2 Indirect Approach

Section 5.5 presents numerous pairwise measures for the comparison of pairs of classes. They can be used to drive the comparison of sets of classes.

### 5.6.2.1  Improving the Direct Approaches for the Comparison of Sets of Classes

One of the main drawbacks of basic vector-based measures is that they consider dimensions as mutually orthogonal and do not exploit class relationships. In order to remedy this, vector-based measures have been formulated to:

- Weight dimensions considering class specificity evaluations (e.g., IC) (Huang et al. 2007; Chabalier et al. 2007; Benabderrahmane, Smail-Tabbone, et al. 2010).
- Exploit an existing pairwise measure to perform base vectors' products (Ganesan et al. 2003; Benabderrahmane, Smail-Tabbone, et al. 2010).

Therefore, pairwise measures can be used to refine the measures proposed to compare sets of classes using a direct approach.

### 5.6.2.2  Aggregation Strategies

A two-step indirect strategy can also be adopted in order to take advantage of pairwise measures to compare sets of concepts:

1. The similarity of pairs of classes obtained from the Cartesian product of the two compared sets has to be computed.
2. Pairwise scores are next aggregated using an aggregation strategy, also called *mixing strategy*.

Classical aggregation strategies can be applied (e.g. max, min, average); more refined strategies have also been proposed. Among the most used we present: Max average (AVGMAX), Best Match Max – BMM (Schlicker et al. 2006) and Best Match Average – BMA (Pesquita et al. 2008):

$$sim_{\text{AVG}}(U,V) = \frac{\sum_{u \in U} \sum_{v \in V} sim(u,v)}{|U| \times |V|}$$

$$sim_{\text{AVGMAX}}(U,V) = \frac{1}{|U|} \sum_{u \in U} max_{v \in V} sim(u,v)$$

$$sim_{BMM}(U,V) = \max(sim_{AVG-dir}(U,V), sim_{AVG-dir}(V,U))$$

$$sim_{BMA}(U,V) = \frac{sim_{AVG-dir}(U,V) + sim_{AVG-dir}(V,U)}{2}$$

### 5.6.3 List of Groupwise Semantic Similarity Measures

| Direct Groupwise Measures | | | | | | |
|---|---|---|---|---|---|---|
| Name | Type | Approach | KR const. | Range | IOI | Comment |
| (Ganesan et al. 2003) Optimistic Genealogy Measure | Sim | Hybrid | RDAG | [0,1] | Yes | Feature-based approach taking into consideration structural properties during the comparison |
| (Popescu et al. 2006) A | Sim | Feature-based | DAG | [0,1] | yes | Weighted Jaccard |
| (Popescu et al. 2006) B | Sim | Feature-based | DAG | [0,1] | | Fuzzy Measure |
| (Chabalier et al. 2007) | Sim | Feature-based (Vector) | RDAG | [0,1] | Yes | Groups of classes are represented using the Vector Space Model and compared using the cosine similarity. |
| (Gentleman 2007) SimLP | Sim | Structural | RDAG | [0,1] | Yes | Similarity as a function of the longest-path shared found in the graph induced by the sets of classes characterizing the compared instances. |
| (Cho et al. 2007) | Sim | Feature-based | RDAG | $\mathbb{R}^+$ | No | Feature-based measures taking into account the specificity of the compared classes. |
| (Pesquita et al. 2008) SimGIC | Sim | Feature-based | DAG | [0,1] | Yes | Jaccard measure in which a set of classes is represented by the classes contained in the graph it induces. |
| (Sheehan et al. 2008) SSA | Sim | Feature-based | RDAG | [0,1] | Yes | Extend the notion of MICA to pair of genes then redefine the Dice coefficient. |
| (Ali & Deane 2009) | Sim | Feature-based | DAG | [0,1] | No | Commonality is assessed considering shared nodes in the graph induced by the ancestors of the compared sets of clas- |

| | | | | | | ses. |
|---|---|---|---|---|---|---|
| (Jain & Bader 2010) TCSS | Sim | Feature-based | RDAG | [0,1] | No | Max Strategy Considering (Jain & Bader 2010) pairwise measure |
| (Diaz-Diaz & Aguilar-Ruiz 2011) | Sim | Structural | RDAG | [0,1] | Yes | Distance taking into account the shortest path between the classes and the depths of the compared classes. |
| (Alvarez & Yan 2011) | Sim/Rel | Structural | None | $\mathbb{R}^+$ | Yes | Structural measure |
| (M. A. Alvarez et al. 2011) SPGK | Sim | Structural | None | | Yes | The set of classes is represented by its induced subgraph. A similarity measure is used to compare the two graphs. |
| (Teng et al. 2013) | Sim | | | | | |

| **Indirect Groupwise Measures** Based on a Direct Approach | | | | | | |
|---|---|---|---|---|---|---|
| Name | Type | Approach | KR const. | Range | IOI | Comment |
| (Ganesan et al. 2003) GCSM | Sim | Feature-based | RDAG | [0,1] | Yes | GCSM: Generalized Cosine-Similarity Measure. Groups of classes are represented using the Vector Space Model. The dimensions are not considered independent, i.e. the similarity of two dimensions is computed using an approach similar to the one proposed by Wu and Palmer. The similarity of the vector representation of the instances is estimated using to the cosine similarity. |
| (Huang et al. 2007) | | | RDAG | [0,1] | Yes | |
| (Benabderrahmane, Smail-Tabbone, et al. 2010) Intelligo | Sim | Feature-based (Vector) | RDAG | [0,1] | Yes | Groups of classes are represented using the Vector Space Model - Also consider (Benabderrahmane, Devignes, et al. 2010). The dimensions are not considered to be independent. |

| Indirect Groupwise Measures<br>Mixing strategies | | | |
|---|---|---|---|
| Name | Range | IOI | Comment |
| Classic approaches<br>Max/Min/AVG, etc.<br>(Azuaje et al. 2005)<br><br>Best Match Max (BMM)<br>Best Match Average<br>(Azuaje et al. 2005) | depends | depends | Properties depend on the measures used to compute the pairwise scores which will be aggregated. In (Couto et al. 2007) the max was used. |

## 5.7 Unification of Similarity Measures for the Comparison of Classes

This section presents the works related to the unification of knowledge-based semantic measures dedicated to the comparison of classes.

### 5.7.1 Similitude between Semantic Measures

Several similitudes have been observed between SMs. As an example, in a tree, the edge-counting strategy defined by (Rada et al. 1989) can also be expressed as a function of the depths of the compared classes and the depth of their LCA (Blanchard 2008):

$$dist_{Rada}(u,v) = depth(u) + depth(v) - 2\, depth(LCA(u,v))$$

Therefore the depth can be seen as a simple expression of the function $\theta$ used to estimate the specificity of a particular class. The edge-counting strategy can thus be defined through the abstract expression:

$$dist_{abstract}(u,v) = \theta(u) + \theta(v) - 2\, \theta(LCA(u,v))$$

We can see that this expression generalize the information theoretical distance proposed by (Jiang & Conrath 1997):

$$dist_{JC}(u,v) = IC(u) + IC(v) - 2\, IC(MICA(u,v))$$

In the same manner, it has also been stressed by several authors, e.g., (Blanchard et al. 2008), that, in a tree[i], the measure proposed by Wu and Palmer can be reformulated by:

$$sim_{WP}(u,v) = \frac{2\, depth(LCA(u,v))}{depth(u) + depth(v)}$$

Therefore, once again, this expression can be generalized by an abstract similarity measure:

$$sim_{Dice-abstract}(u,v) = \frac{2\, \theta(LCA(u,v))}{\theta(u) + \theta(v)}$$

Such an abstract expression of a similarity measure highlights the relationship between the structural measure proposed by Wu and Palmer and the information theoretical measure proposed by Lin:

---

[i] In which a transitive reduction has been performed.

$$sim_{Lin}(u,v) = \frac{2\,IC(MICA(u,v))}{IC(u) + IC(v)}$$

A similar approach can be adopted to underline the relationship between some feature-based measures and information theoretical measures. Indeed, under specific tuning, comparing two classes using a feature-based measure, i.e., according to their shared and distinct features, can be equivalent to considering a particular expression of an information theoretical measure.

As an example, defining $A(u)$ the features of the class $u$, and using a SM based on the Dice index, we obtain the following feature-based measure:

$$sim_{Dice-FB}(u,v) = \frac{2\,|A(u) \cap A(v)|}{|A(v)| + |A(u)|}$$

Since in a tree, two classes have a unique LCA, this feature-based expression can be reformulated as:

$$sim_{Dice-FB}(u,v) = \frac{2\,|A(LCA(u,v))|}{|A(v)| + |A(u)|}$$

Thus, this expression is a specific expression of the abstract formulation of Dice formula $sim_{Dice-abstract}$ presented above, defining $\theta(u) = |A(u)|$.

Using a similar reformulation of the measure proposed by (Stojanovic et al. 2001), (Blanchard 2008; Blanchard et al. 2008) also underlined that, in trees, feature-based expressions can be reformulated using depth estimator (since $A(c) = depth(c) + 1$). Therefore, in a tree, we obtain:

$$sim_{CMatch}(u,v) = \frac{|A(u) \cap A(v)|}{|A(u) \cap A(v)|} = \frac{depth(LCA(u,v)) + 1}{depth(u) + depth(v) + 2 - (depth(LCA(u,v)) + 1)}$$

### 5.7.2 Framework for the Expression of Semantic Measures

The feature model proposed by Tversky was the first formulation of a framework from which several similarity measures can be derived through parametric formulation of measures (Tversky 1977). The feature model proposes to compare objects represented through sets of features. It therefore requires the features of the elements we want to compare to be specified.

For the comparison of classes, this model requires the definition of a function characterizing the features of a class. The similarity is next intuitively defined based on the common and distinctive features of the compared classes. This approach is used for a long time to compare sets according to the study of their shared and distinct elements (e.g., Jaccard Index, Dice coefficient). As we have seen in section 2.2.2, Tversky defined the contrast model and the ratio model as func-tions which can be used to compare objects represented as sets of features. Below we recall the formulation of the ratio model:

$$sim_{RM}(u,v) = \frac{f(U \cap V)}{\alpha\, f(U \backslash V) + \beta\, f(V \backslash U) + f(U \cap V)}$$

Such a general parameterized formulation of a similarity measure can be used to derive a large number of concrete measure. As an example, considering the salience of a set of feature (i.e., the function $f$) as the cardinality of the set, and $\alpha = \beta = 1$, the ratio model leads to the original definition of the Jaccard index. Setting $\alpha = \beta = 0.5$ leads to the Dice coefficient.

A large diversity of set-based measures can expressed from specific instances of such parameterized functions. In other words, such general measures are abstract similarity measures which can be used to instantiate concrete similarity measures through the definition of a limited set of parameters.

The framework proposed by Tversky constrains the compared objects to be represented as sets of features and the similarity to be assessed as a function of the commonalities and differences of the two sets. By definition the contrast model and the ratio model are therefore constrained to set-based formulations of measures. These models are more particularly constrained to fuzzy set theory, since, originally, Tversky defined the commonalities and differences of two objects as a function of the salience of their shared and distinct features.

Most set-based measures can be expressed using Caillez and Kuntz $\sigma_\alpha$ formulation, and Gower and Legendre $\sigma_\beta$ formulation (Blanchard et al. 2008). Since set-based measures can be used to design semantic measures, $\sigma_\alpha$ and $\sigma_\beta$ can be generalized in a straightforward manner according to the Tversky feature approach:

$$\sigma_\alpha(u,v) = \frac{f(U \cap V)}{\left(\frac{f(U)^\alpha + f(V)^\alpha}{2}\right)^{1/2}}$$

$$\sigma_\beta(u,v) = \frac{\beta\, f(U \cap V)}{f(U) + f(V) + (\beta - 2)f(U \cap V)}$$

Therefore defining the function $f(X)$ as the cardinality of the set of features $X$, the abstract formulation $\sigma_\alpha$ can be used to derive the Simpson ($\alpha = -\infty$) and Ochaiai ($\alpha = 0$) coefficients, to cite a few (Choi et al. 2010). The $\sigma_\beta$ reformulation can also be used to express other numerous measures, e.g. Sokal and Sneath ($\beta = 0.5$), and Jaccard index ($\beta = 1$) and Dice coefficient ($\beta = 2$) (Blanchard et al. 2008; Choi et al. 2010).

In (Roddick et al. 2003) the authors propose a model of semantic distance relying on a graph-based approach which quantifies the distance between data values as a function of graph traversal. However, Blanchard and collaborators, were the first to take advantage, in an explicit manner, of abstract definition of SMs for the comparison of pairs of classes defined in KRs (Blanchard et al. 2008). In their studies, the authors focused on an information theoretical expression of semantic measures to highlight relationships between several measures proposed in the literature. As we have seen based on the intuitive notion of commonalities and differences, and based on a particular expression of the notion of specificity, the authors underlined that the expressions proposed by Wu and Palmer and Lin can both be derived from an abstract expression of the Dice Index.

$$sim_{WP}(u,v) = \frac{2\,depth(LCA(u,v))}{depth(u) + depth(v)}$$

$$sim_{Lin}(u,v) = \frac{2\,IC(MICA(u,v))}{IC(u) + IC(v)}$$

Therefore both Wu and Palmer and Lin measures rely on a general expression of the Dice coefficient, here named $(sim_{G-DICE})$ which corresponds to the ratio model defining $\alpha = \beta = 0.5$, they can also be seen as particular expression of $\sigma_\beta$ with $= 2$ :

$$sim_{G-DICE}(u,v) = \frac{2\,f(U \cap V)}{f(U) + f(V)}$$

Indeed, defining $U \cap V$ as the properties of the least common ancestors of the compared concepts which maximizes a function $\theta$ – with $\theta(c) = depth(c)$ for the measure proposed by Wu and Palmer and $\theta(c) = IC(c)$ for the measure proposed by Lin, and considering $f(U) = \theta(u)$ with the respective $\theta$ function selected to distinguish the LCA in the two measures, we can derive both measures from the general expression $sim_{G-DICE}$. Another expression derived from such a general expression of the Dice coefficient have been proposed by the authors in (Blanchard et al. 2006). Several other abstract expressions of measures can be found in (Blanchard et al. 2008; Blanchard 2008).

In their studies, summarized in the PhD thesis of Blanchard (Blanchard 2008)[i] and in (Blanchard & Harzallah 2005; Blanchard et al. 2008), the authors stressed the suitability of the decomposition of SMs through abstract expressions to further characterize their properties and to study groups of measures.

Other authors have also demonstrated relationships between different similarity measures and took further advantage of abstract frameworks to design new measures or to study existing ones (Cross 2006; Pirró & Euzenat 2010a; Cross & Yu 2010; Sánchez & Batet 2011; Cross et al. 2013). These contributions mainly focused on establishing local relationships between set-based measures and measures framed in Information Theory. (Pirró & Euzenat 2010a) present an Information Theoretical expression of the component distinguished by the feature model (commonalities and difference) and therefore enable the expression of numerous measures based on the ratio model or the contrast model. Table 8 presents the mapping between feature-based and information theoretical similarity models proposed by the authors.

---

[i] In french

| Description | Feature-based model | Information-theoretic model |
|---|---|---|
| Salience of Common Features | $f(U \cap V)$ | $IC(MICA(u,v))$ |
| Salience of the features of $u$ not shared with the features of $v$ | $f(U \backslash V)$ | $IC(u) - IC(MICA(u,v))$ |
| Salience of the features of $v$ not shared with the features of $u$ | $f(V \backslash U)$ | $IC(v) - IC(MICA(u,v))$ |

**Table 8** : Mapping proposed by (Pirró & Euzenat 2010a) between the feature model and the information theoretic approach (reproduction with some modifications in order to be in accordance with the notions and notations introduced).

Setting $\alpha = \beta = 1$, the authors proposed the definition of a new measure which correspond to a particular expression of an abstract form of the Jaccard coefficient:

$$sim_{Faith}(u,v) = \frac{IC(MICA(u,v))}{IC(u) + IC(v) - IC(MICA(u,v))}$$

(Sánchez & Batet 2011) also proposed a framework, grounded in information theory, which allows several measures (i.e., edge-counting and set-based coefficients) to be uniformly redefined according to the notion of IC. They explicitly defined a mapping table to take advantage of set-based measures for the expression of measures framed in information theory.

| Expressions found in set-based similarity coefficients | Approximation in terms of IC |
|---|---|
| $|U|$ | $IC(u)$ |
| $|V|$ | $IC(v)$ |
| $|U \cap V|$ | $IC(MICA(u,v))$ |
| $|U \backslash V| = |U| - |U \cap V|$ | $IC(v) - IC(MICA(u,v))$ |
| $|V \backslash U| = |V| - |U \cap V|$ | $IC(v) - IC(MICA(u,v))$ |
| $|U \cup V| = |U| + |V| - |U \cap V|$ | $IC(u) + IC(v) - IC(MICA(u,v))$ |
| $|U| + |V|$ | $IC(u) + IC(v)$ |

**Table 9**: Mapping proposed by (Sánchez & Batet 2011) between expressions found in set-based similarity measures and the information theoretic approach (reproduction with some modifications in order to be in accordance with the notations introduced).

Based on the correspondences defined in Table 9, the authors derived several SMs from set-based measures. They also proposed several redefinitions of structural measures using the notion of information content. As an example they underlined the link between the edge-counting strategy and the information theoretical measure defined by Jiang and Conrath[i]:

---

[i] This correspondence have also been underlined in (Blanchard 2008).

$$sp(u,v) \simeq depth(u) + depth(v) - depth(LCA(u,v))$$
$$sp(u,v) \simeq IC(u) - IC(v) + IC(v) - IC(u)$$
$$sp(u,v) \simeq IC(u) + IC(v) - 2IC(MICA(u,v))$$

In the same vain (Cross 2004; Cross 2006; Cross et al. 2013) proposed a similar contribution in which feature-based approaches and measures based on Information Theory are expressed through the frame of the fuzzy set theory.

In (Mazandu & Mulder 2013), the authors propose another general framework and unified description of measures relying on the notion of information content for the comparison of pairs of classes. Like (Blanchard et al. 2008), the authors focused in an information theoretical definition of measures to underline similarities between existing measures.

Despite the suitability of these frameworks for studying some properties of SMs, only a few works rely on them to express measures (Sánchez & Batet 2011; Cross et al. 2013). Moreover, current frameworks only focus on a specific paradigm (e.g., feature-based strategy), to express measures. In fact, most existing frameworks only encompass a limited number of measures and were not defined in the purpose of unifying measures expressed using the variety of paradigms reviewed in section 3.4.1.

The main limitation of these frameworks rely on the fact that they derive from the feature model or an information theoretical expression of the feature model and are therefore by definition limited to these paradigms. To overcome this limitation (Harispe, Sánchez, et al. 2013) recently proposed a framework framed in the strategy adopted to characterize the representation of the compared elements. This framework has its roots in the teaching of cognitive sciences regarding the central role played by the representation adopted to characterize compared elements. Therefore, contrary to the other frameworks, this proposal is not limited to specific approaches constrained by a particular representation of the compared elements (feature-based, structural, information theoretical). Indeed, this framework defines the possibility to explicitly express the strategy adopted to characterize the representation of a class (set-based representation, information-theoretical, graph-based, etc.).The framework further distinguishes the primitive functions commonly found in SM expressions (e.g., functions used to characterize the commonality and the differences of the compared representations, the saliency of a representation).

## 5.8 Semantic Relatedness between Two Classes

The semantic measures which can be used to assess the semantic relatedness of a pair of classes generalize those defined for the estimation of the semantic similarity. These measures take advantage of all predicates defined in the semantic graph. Generally, these measures are specific expressions of the structural measures presented for the estimation of the semantic similarity of two classes. Refer to the contributions of (Sussna 1993; Wang et al. 2007).

## 5.9 Semantic Relatedness between Two Instances

This subsection presents the various approaches which can be used to compare a pair of instances[i].

Evaluating the proximity between instances requires defining a representation (or canonical form) to characterize an instance. Four approaches can be distinguished depending on the canonical form adopted:

- *Instances represented as graph nodes.*
- *Instances represented as sets of classes.*
- *Instances represented as sets of properties.*
- *Hybrid techniques*

Most of the measures used to compare instances have already been introduced in section 5.4.1. We briefly recall the various strategies which can be adopted according to the representation of an instance, and we particularly focus on the comparison of instances through the notions of projections.

### 5.9.1 Comparison of Instances Using Graph Structure Analysis

Two instances can be compared using their interconnections in the graph of relationships defined in the KR. In this case, structural measures introduced in section 5.4.1 can be used in a straightforward manner, e.g., shortest path techniques, random walk approaches, SimRank (Jeh & Widom 2002).

### 5.9.2 Instances as Sets of Classes

The semantic relatedness of two instances can be evaluated regarding reductions of the compared instances as sets of classes. In this case, the approaches defined to estimate the semantic similarity of two sets of classes are used (refer to section 5.6). Such an approach is commonly used to compare instances characterized by classes or concepts structured in a KR, e.g. gene products annotated by Gene Ontology terms, documents annotated by MeSH descriptors, etc.

### 5.9.3 Instances as a Set of Properties

The comparison of instances is most of the time driven by the comparison of their representation through sets of properties. The SMs which can be used to compare such representations of instances are the measures introduced for the graph property model in section 5.4.1.2.

The following presentation focuses on the comparison of instances characterized through the notion of projection. These approach has been defined in (Harispe, Ranwez, et al. 2013a) and generalizes the comparison of instances represented through sets of properties.

*5.9.3.1 Characterization of Properties through Projections*

---

[i] This is an extended version of the state of the art presented in (Harispe, Ranwez, et al. 2013a)

A direct or indirect property of an instance $i$ corresponds to a partial representation of $i$. In Figure 16 for example, the *rollingStones* instance can be represented by its name or music genres. A *simple property* of an instance is therefore expressed through resources linked to it. Representing an instance through its labels is therefore the same as considering all the $l$ labels for which a path links $i$ to $l$ through the relationship `rdf:label`. In other words, it correspond to considering all the labels for which a triplet (`i,rdf:label,l`) exists.

In a general manner, the path linking two resources is characterized by path pattern $< r_0, \dots, r_n >$, with $r_i \in R$, the set of predicates defined in the KR. A path is therefore associated in this manner with a range defined by the type of resources specified by the range of $r_n$, the last predicate composing the path pattern. The definition of path pattern thus enables to characterize some of the properties of instances through a path $p: I \to K'$, with $K'$ the range of path pattern $p$, a set of values that may be included in $C, I$ or composed of values of the type `rdfs:Datatype`, e.g., String.

Let's distinguish three types of paths pattern depending on the range of their last predicate $r_n$:

- *Data*: the range of $r_n$ is a set of data values, e.g. Strings, Dates (Figure 16, case 2).
- *Instances*: the range of $r_n$ is a set of instances (Figure 16, case 1).
- *Classes*: the range of $r_n$ is a set of classes (Figure 16, case 3).

A path pattern may be used to characterize simple (either direct or indirect) properties of an instance. Complex properties however require several paths in order to be expressed. As an example, the comparison of two music bands through the Euclidian distance between their places of origin does indeed involve defining a complex property encompassing the latitude and longitude of a place that requires two paths $<hometown, geo:lat>$ and $<hometown, geo:long>$ (Figure 16, case 4). In other words, the information characterizing a music band via a property defining its place of origin corresponds to the projection of the instance onto two specific resources capable of being reached through paths in the semantic graph. In order to characterize all properties of an instance, the notion of path pattern can thus be generalized by introducing the notion of projection.

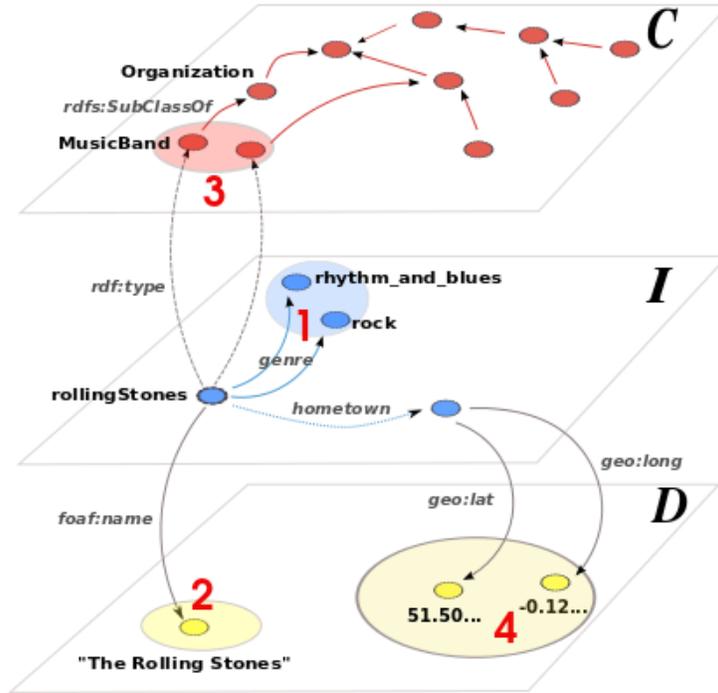

**Figure 16**: Examples of properties associated with the class `MusicBand`.

### 5.9.3.2 Definition of a Projection

A projection refers to projecting a mathematical structure from one space to another. In formal terms, a projection $P$ is composed of a set of paths and is defined by $P: I \rightarrow K$, with $K$ being the set defining the types of projection $k \in K$, onto which an instance can be *projected*.

The projection type corresponds to the range associated with the projection, i.e., the type of values potentially used to characterize the instance. When simple projections are used, i.e., when the projection is composed of a single path pattern, then the projection range is defined by the path range, i.e., $K = K'$. Yet when complex projections involving multiple paths are used, other types of projections can be defined, in yielding $K = K' \cup K''$ with $K''$ being a set indicating the complex objects available for use in representing the complex properties of an instance. Let's note that complex objects are used to represent properties not explicitly expressed in the knowledge base, e.g. geographic location (latitude, longitude).

Four types of projections can therefore be distinguished: the three capable of being associated with a single path pattern (*Data, Instances, Classes*), and the *Complex* type used to represent an instance by means of a set of complex objects combining various (simple) properties. Let's denote $P^k$ the projection of range $k \in K$ and $P^k(i)$ the type $k$ projection of instance $i$.

### 5.9.3.3 Characterization of Instances through Projections

Any instance can be represented by a set of projections. We therefore define a set of projections as a *context of projection* which can be associated to any set of instances, e.g., class, SPARQL query.

Let's consider the definition of a context of projection defined in order to characterize the instances of a specific class. We denote $CP^c$ the context of projection associated to class $c$. This context of projection defines the approach adopted to represent an instance of class $c$, by distinguishing the various properties of interest when characterizing an instance of $c$. The proximity of two instances of $c$ will next be computed regarding the projections defined in $CP^c$.

The SM used to assess the proximity takes into account all projections composing the context of projection which has been defined. Therefore, this SM requires a method to compare two instances considering a specific projection.

### 5.9.3.4 Comparison of Two Projections

Each projection is associated with a measure $\sigma^k$ that enables comparing a pair of instance projections of type $k$, where $\sigma^k: k \times k \rightarrow [0,1]$. Recall the types of projections which have been defined: *Data, Instances, Classes*, and *Complex*.

- *Classes*: Two projections of the *Classes* type can be compared using a SM adapted to a comparison of classes.

- *Data*: A comparison of *Data* type projection requires defining a measure adapted to the type of values constituting the sets of data values produced by the given projection. As an example, two strings may be compared using the Levenshtein distance (Levenshtein 1966).

- *Instances* type projections, the projection is associated to a set of instances, they can be compared using set-based measures, e.g., in order to evaluate the size of the intersection of the two sets.

- *Complex* projections require defining a measure to enable comparing two complex objects. Let's note that in some cases, complex objects or compared values will require some data pre-processing prior to use of the proximity function; as an example, such a pre-processing step could consist of computing the body mass index from the size and weight of an instance of a class *Person*.

As previously observed, a projection defines a set of resources that characterize a specific property of an instance. To estimate the similarity of two instances relative to a specific projection, a measure $\sigma^k$ must be specified so as to compare two sets (sometimes singletons) of resources. Various approaches are available for evaluating these two sets, namely:

- *Cardinality*: The measure evaluates the cardinality of both sets, e.g., by comparing two instances of a class *parent* with respect to the number of children they have.

- *Direct method*: A measure adapted for a set comparison is to be used (e.g. Jaccard index); one example herein would be to compare instances relative to the number of overlapping resources, e.g. the number of common friends. Vector representations can also be used.

- *Indirect method:* This method relies on evaluating the proximity of the pair of resources able to be built by considering the compared sets (a Cartesian product of sets), e.g. couples of strings. In this case, an aggregation strategy must be defined to aggregate the proximity scores obtained for all resource pairs built from the Cartesian product of the two compared sets. Classical operators such as Min, Max, Average or more refined approaches may be used to aggregate the scores (refer to section 5.6.2.2).

As pointed out above, when an indirect method is used to compare two projections, a measure enabling the comparison of two sets of resources needs to be defined. Several approaches are available for comparing sets of classes, strings or numerical values. Note that the relevance of a measure is once again defined by both the application context and the semantics the similarity scores are required to carry.

Two groups of instances can be compared by using a direct or an indirect approach. When an indirect approach is selected, a strategy to enable comparing a couple of instances must be determined. It is therefore possible to use the context of projection defined for the class of the two instances under comparison. This context of projection actually defines the properties that must be taken into account when comparing two instances of this specific type. Applying such a strategy potentially corresponds to a recursive treatment, for which a stopping condition is required. In all cases, computing the proximity of two projections should not imply use of the context of projection containing both projections. A proximity measure can thus be represented through an execution graph highlighting the dependencies occurring between contexts of projection. Consequently, this execution graph must be analysed to detect cycles, for the purpose of ensuring computational feasibility. If a cycle is detected, the measure will not be computable.

### 5.9.3.1 Comparison of Two Instances through Their Projections

Once a measure has been chosen to compare each projection, then a general SM $\sigma_c$ can be aggregated between two instances $u$ and $v$ of the set of instances $c$, e.g:

$$\sigma_c(u, v) = \sum_{P_i^k \in CP^c, \exists P_i^k(u) \wedge \exists P_i^k(v)} w_i \times \sigma^k\big(P_i^k(u), P_i^k(v)\big)$$

where $w_i$ is the weight associated to the projection $P_i^k$ and the sum of weights equals 1. This measure exploits each projection shared between the compared instances. In other words, the instances of the class are compared based on a specific characterization of all relevant properties that must be taken into account in order to rigorously conduct the comparison.

### 5.9.1 Hybrid Techniques

(Pirró 2012) proposes an hybrid techniques taking into consideration the direct properties characterizing the compared instances as well as the shortest path linking the two instances.

# 6 Challenges

At the light of the state-of-the-art of the large diversity of SMs presented in this survey, this section highlights some of the challenges offered to the communities involved in SM study.

## 6.1 Better Characterize Semantic Measures and their Semantics

All along this paper we have stressed the importance to control the semantics associated to SMs, i.e., the meaning of the scores produced by the measures. This particular aspect is of major importance since the semantics of measures must be explicitly understood by users of SMs as it conditions the relevance to use a specific measure in a particular context. Indeed, the semantics of a SM is generally not discussed in proposals (expect some broad distinction between similarity and relatedness). However, for instance, measures only based on taxonomical analysis (knowledge-based semantic similarity measures) can have different meanings depending on the assumptions on which they rely. In this paper, we have underlined that the semantics associated to SMs can only be understood with regard to the semantic proxy used to support the comparison, the mathematical properties associated to the measures and the assumptions on which the measures are based.

The semantics of the measures can therefore only be captured if a deep characterization of SMs is provided. In the last decades, researchers have mainly focused in the design of SMs and despite the central role of the semantic of SMs, only few contributions focused on this specific aspect. As we have seen, this *disinterest* to the study of the semantics of the measures can be partially explained by the fact that numerous SMs have been designed in order to better mimic human appreciation of semantic similarity/relatedness. In this case, the semantics to be carried by the measures is expected to be implicitly constrained by the benchmarks used to evaluate measures' accuracy. Nevertheless, despite evaluation protocols based on *ad-hoc* benchmarks are meaningful to compare SMs in particular contexts of use, they do not give access to a deep understanding of measures and therefore lack to provide the information needed to take advantage of SMs in other contexts of use.

The implications of a better characterization of semantic measures are numerous. We already stressed its importance for the selection of SMs in specific contexts of use. Such a characterization could also benefit cognitive sciences. Indeed, as we have seen in section 2.2, the proposals of cognitive models aiming to explain human appreciation of similarity have been supported by the study of properties expected by the measures. As an example, recall that the spatial models have been challenged according to the fact that human appreciation of similarity has proven not to be in accordance with the axioms of distance. Therefore, characterizing: (i) which SMs best performed according to human expectations of semantic similarity/relatedness and (ii) the properties satisfied by these measures, could help cognitive scientists to improve existing models of similarity or derive more accurate ones.

In this paper, we have proposed an overview of the various SMs which have been proposed to compare units of language, classes or instances semantically characterized. In section 3.1, we distinguished various aspects of SMs which must be taken into account for their broad classification:

- The types of elements which can be compared.
- The semantic proxies used to extract semantic evidences on which will be based the measures.
- The canonical form adopted to represent the compared elements and therefore enable the design of algorithms for their comparison.

In section 2.3.1, we also recalled some of the mathematical properties which can be used to further characterize SMs. In section 2.1.3, based on the several notions introduced in the literature, we proposed a characterization of the general semantics which can be associated to SMs (e.g., similarity, relatedness, distance, taxonomical distance). Finally, all along this paper, and particularly in section 5.3, we distinguished several semantic evidences on which can be based SMs and we underlined the assumptions associated to their consideration.

We encourage SM designer to provide an in-depth characterization of the measures they propose. To this end, they can use the various aspects and properties of the measures distinguished in this paper. We also encourage the communities involved in the study of SMs to better define what a *good* semantic measure is and what makes a measure better than another. In this aim, the study of the role of contexts seems to be of major importance. Indeed, as we have seen in section 4.2, accuracy of measures can only be discussed with regard to specific expectations of measures. Several other properties of measures could also be taken into account and further investigated:

- Algorithmic complexity.
- Degree of control on the semantics of the scores produced by the measures.
- The confidence which can be associated to a score.
- The *robustness* of a measure, i.e., the capacity for a measure to produce robust scores considering the uncertainty associated to expected scores or perturbations of the semantic proxies on which rely the measure (modification of the KRs, corpus modifications).
- The discriminative power of the measure, i.e., the distribution of the scores produced by a measure.

## 6.2 Provide Tools for the Study of Semantic Measures

The communities studying and using SMs require software solutions, benchmarks, and theoretical tools to compute, compare and analyse SMs.

### 6.2.1 Develop benchmarks

As we have seen in section 4.2.2 several benchmarks exist to evaluate semantic similarity and relatedness. Most of these benchmarks aim at evaluating SMs' accuracy according to human appreciation of similarity. For the most part they are composed of a reduced number of entries, e.g., pairs of words/concepts, and have been computed using a reduced pool of subjects.

Initiative for the development of benchmarks must be encouraged in order to obtain larger benchmarks in various domains of study. Word-to-word benchmarks must be conceptualized as much as possible in order for them to be used to evaluate knowledge-based SMs[i]. It is also important to propose benchmarks which are not based on human appreciation of similarity, i.e. benchmarks relying on an indirect evaluation strategy.

### 6.2.2 Develop Generic Open-source Software Solutions for Semantic Measures

In section 4.1, we proposed an overview of the main software solutions dedicated to SMs. They are of major importance to: (i) ease the use of the theoretical contributions related to SMs, (ii) support large scale comparisons of measures and therefore better understand the measures, (iii) develop new proposals.

Software solutions dedicated to distributional measures are generally developed without being restricted to a specific corpus of texts. They can therefore be used in a large diversity of contexts of use as long as the semantic proxy considered corresponds to a corpus of texts.

Software solutions dedicated to knowledge-based SMs are most of the time developed for a specific domain (e.g., refer to the large number of solutions developed for the Gene Ontology alone). Such a diversity of software is limiting for SMs designers since implementations made for a specific KR cannot be reused in applications relying on others KRs. In addition, it hampers the reproducibility of results since some of our experiments have shown that specific implementations tend to produce different results[ii]. In this context, we encourage the development of generic open-source software solutions not restricted to specific KRs. This is challenging since the formalism used to express KRs is not always the same and specificities of particular KRs sometimes deserve to be taken into account to develop SMs. However, there are several cases in which generic software can be developed. As an example, numerous knowledge-based SMs rely on data structures corresponding to partial ordered set or more generally semantic graphs. Other measures are designed to take advantage of KRs expressed in standardized languages such as RDF(S), OWL. Generic software solutions can be developed to encompass these cases. The development of the Semantic Measures Library is an example of such an initiative (Harispe, Ranwez, et al. 2013b). Reaching such a goal could open interesting perspectives. Indeed, based on such generic and robust software supported by several communities, domain specific tools and various programming language interfaces can next be developed to support specific use cases and KRs.

The diversity of software solutions also has benefits as it generally stimulates the development of robust solutions. Therefore, another interesting initiative, complementary to the former, could be to provide generic and domain specific tests to facilitate both the development and the evaluation of the software solutions. Such tests could for instance be the expected scores to be produced by specific SMs according to a reduced example of a corpus/KR. This specific aspect is important in order to standardize the software solutions dedicated to SMs and to ensure users of specific solutions that the score produced by the measures are in accordance with the original definitions of the measures.

---

[i] It is quite common to find papers describing knowledge-based SM evaluation using word-to-word benchmarks without giving access to the concepts associated to each words and the strategy adopted when multiple concepts could be associated to a word.

[ii] This can be explained by bugs or particular interpretations on the definitions of measures or on the way to handle KRs. Refer to https://github.com/sharispe/sm-tools-evaluation for an example in the biomedical domain.

As we have seen in section 4.2, evaluation of SMs is mainly governed by empirical studies used to assess their accuracy according to expected scores/behaviours of the measures. Therefore, the lack of open-source software solutions implementing a large diversity of measures hampers the studies of SMs. It explains for instance that evaluations of measures available in the literature only involve the comparison of a subset of measures which is not representative of the diversity of SMs today available. Initiatives aiming at developing robust open-source software solutions giving access to a large catalogue of measures must therefore be encouraged. It's worth to note the importance of these solutions to be open-source. Our communities also lack open-source software dedicated to SMs evaluation. Indeed, despite some initiative in specific domains[i], evaluations are not made through a common framework like it's done by most communities, e.g. information retrieval, ontology alignment.

### 6.2.3 Develop Theoretical Tools for Semantic Measures

The large amount of SMs which have been proposed is hard to study, e.g. deriving interesting properties of measures require the analysis of each measure. However, as we have seen in section 5.7, several initiatives have proposed theoretical tools to ease the characterization of measures; they open interesting perspectives to study groups of measures. Such theoretical frameworks have proven to be essential to better understand the limitation of existing measures and the benefits of new proposals. They are also critical to distinguish the main components on which the measures rely. Characterizing such components open interesting perspectives to improve families of SMs based on the components, e.g., the definition of the GrasM strategy to better characterize the commonality of two classes is an example of the redefinition of a component used by several measures.

## 6.3 Standardize Knowledge Representation Handling

In section 5.2.2, we discussed the process required to transform a KR to a data structure which can be processed by the measures. Such a process is actually too much subject to interpretations and deserves to be carefully discussed and formalised. Indeed, as an example, we stressed that numerous measures consider KRs as semantic graphs despite the fact that the formalism on which KRs rely cannot be mapped to semantic graphs without some loss of knowledge. The impact of such a reduction of KRs is of major importance since it can highly impact the results produced by the measures[ii]. The treatment performed to map a KR to a semantic graph is generally not documented which explains some of the difficulties encountered to reproduce experiments.

## 6.4 Promote Interdisciplinarity

From cognitive sciences to biomedical informatics, the study of SMs involves numerous communities. Efforts have to be made to promote interdisciplinary studies and to federate the contributions made in the various fields. We briefly provide a non-exhaustive list of the main commu-

---

[i] E.g., CESSM to evaluate SMs designed for the Gene Ontology. Note that this solution is not open-source, it can therefore not be used to support large scale evaluations and it's impossible to reproduce the experiments and the conclusion derived from them…

[ii] Consider for instance the simple case of a taxonomy corresponding to a semantic graph in which redundant relationships have been defined.

nities involved in the studies of SMs as well as communities which could contribute to their studies or field of studies which must be relevant to solicit to further study SMS. The list is alphabetically ordered:

- **Biomedical Informatics and Bioinformatics**: active in the definition and study of SMs. These communities are also active users of SMs.

- **Cognitive Sciences**: propose cognitive models of similarity and mental representations which can be used to improve the design of SMs and better understand human expectations regarding similarity/relatedness. These communities can also use empirical studies made for the evaluation of SMs to discuss the cognitive models they propose.

- **Complexity Theory**: study of the complexity of SMs.

- **Geoinformatics**: Definition and study of SMs. They are also active users of SMs.

- **Graph Theory**: several major contributions relative to graph processing. Essential for the optimization of measures based on graph-based KRs. This community will play an important role in the near future of knowledge-based SMs since large semantic graphs composed of billions of relationships are today available. Processing such semantic graphs require optimization techniques to be developed.

- **Information Retrieval**: define and study SMs taking advantage of corpus of texts or KRs.

- **Information Theory**: play an important role to better understand the notion of information and to define metrics which can be used to capture the amount of information conveyed, shared and distinct between the compared elements, e.g., notion of information content.

- **Knowledge Engineering**: study of KRs and define KRs which will further be used by some SMs. This community could for instance play an important role to characterize the assumptions made by the measures.

- **Logic**: define formal method to express and take advantage of knowledge. This community can play an important role to characterize the complexity of knowledge-based semantic measures for instance.

- **Machine Learning**: play an important role for the definition of techniques and parameterized functions which can be used for the definition and tuning of SMs.

- **Measure Theory**: define a mathematical framework for the study of the notion of measure. Essential to derive properties of measures, better characterize SMs and to take advantage of theoretical contributions proposed by this community.

- **Metrology**: study both theoretical and practical aspects of measurements.

- **Natural Language Processing**: actively involved in the definition of distributional measures. They propose models to characterize corpus-based semantic proxies and to define measures for the comparison of units of language.

- **Optimization area**: important contributions which can be used to optimize measures, to study their complexity and to improve their tuning.

- **Philosophy**: play an important role in the definition of essential concepts on which SMs rely on, e.g., definitions of the notions of Meaning, Context.

- **Semantic Web and Linked Data**: define standards (e.g., languages, protocols) and process to take advantage of KR. The problematic of ontology alignment and instance matching are actively involved in the definition of (semantic) measures based on KRs. This community is active in the definition of measures.

- **Statistics and Data Mining**: Important contributions which can be used to characterize large collection of data. Major contributions in clustering can for instance be used to better understand SMs.

## 6.5 Study the Algorithmic Complexity of Semantic Measures

As we have seen all along this survey, most contributions have focused on the definition of SMs. Their algorithmic complexity is however near inexistent despite the fact that this aspect is essential for practical applications. Therefore, to date, no comparative studies can be made to discuss the benefits of using computationally expensive measures. This aspects is however essential to compare SMs. Indeed, in most application contexts, users will prefer to reduce measure accuracy for a significant reduction of the computational time and resources required to use a measure. To this end, SM designers must, as much as possible, provide algorithmic complexity of their proposals. In addition, as the theoretical complexity and the practical efficiency of an implementation are different, developers of software tools must provide metrics to discuss and compare measures' implementation efficiency.

## 6.6 Support Context-Specific Selection of Semantic Measures

Both theoretical and software tools must be proposed to orient end-users of SMs in the selection of measures according to the needs defined by their application contexts. Indeed, despite most people only (blindly) consider benchmark results to select a measure, efforts have to be made in order to orient end-users in the selection of the best suited approach according to their usage context, understanding the implications (if any) to use an approach compared to another.

The several properties of measures we have presented to characterize the measures can be used to guide the selection of SMs. Nevertheless, numerous large-scale comparative studies have to be performed to better understand the benefits to select a specific SM in a particular context of use.

# 7 Conclusions

In this paper, we have introduced the large diversity of semantic measures (SMs) which can be used to compare various types of elements, i.e., units of language, concepts or instances, based on texts and knowledge representation analysis. These measures have been proved to be essential tools to drive the comparison of such elements by taking advantage of semantic evidences which formally or implicitly support their meaning or describe their nature. From Natural Language Processing to Biomedical Informatics, SMs are used in a broad field of applications and have become a cornerstone for designing intelligent agents which will for instance use semantic analysis to mimic human ability to compare *things*.

SMs, through the diverse notions presented in this paper (e.g., semantic similarity/relatedness and distance), have been actively studied by several communities over the last decades. However, as we have seen, the meaning of the large terminology related to SMs was not clearly defined and misuses are frequent in the literature. Based on the commonly admitted definitions and new proposals, this paper presents a classification and clear distinctions between the semantics carried by the numerous notions related to SMs.

The extensive survey presented in this paper offers an overview of the main contributions related to the broad subject of SMs. It also underlines interesting aspects regarding the interdisciplinary nature of this field of study. Indeed, as we have seen, the design of SMs is (implicitly or explicitly) based on models of mental representations proposed by cognitive sciences. These models are further expressed mathematically according to specific canonical forms adopted to represent elements and functions designed to compare these representations – this whole process enables computer algorithms to compare units of language, concepts or instances, taking into account of their semantics.

Our analysis of existing contributions underlines the lack of an extensive characterization of measures and provides several aspects and properties of SMs which can be used to this end. We also stressed the importance for our communities to better capture the semantics associated to SMs, i.e., to control the meaning which can be associated to a score produced by a SM. Our analysis helped us to distinguish three main characteristics which can be used to characterize this semantics: (i) the semantic evidences which are used to drive the comparison, (ii) the mathematical properties of the measure and (iii) the assumption on which is based the measure.

Finally, at the light of the state-of-the-art of the analysis of the large diversity of contribution related to SMs presented in this paper, we stressed the importance: (i) to better characterize SMs, (ii) to develop both software and theoretical tools dedicated to their analysis and computations, (iii) to standardize and to formalize some treatments performed by SMs which are subject to interpretations, (iv) to facilitate the selection and comparison of measures (e.g., by exploring new properties of measures, by defining new domain-specific benchmarks), and (v) to promote interdisciplinarity to federate the efforts made by the several communities involved in SMs study.

## Contributions

This paper summarizes the state-of-the-art related to semantic measures which have been made by Sébastien Harispe during his PhD thesis. Sébastien Harispe wrote the paper; co-authors (PhD supervisors) supervised the project and provided corrections and advices.

# Appendix

## 1. Categorization of Contributions Related to Semantic Measures

This appendix proposes a categorization of some contributions related to semantic measures. In each category reference ordering is made by date.

- Seminal works
  Contributions commonly admitted as seminal contributions by the different communities involved in the study of semantic measures:
  (Tversky 1977; Lin 1998; Deza & Deza 2013)

- Surveys
  Different surveys relative to the notion of semantic measures (most of them focus on a specific type of measure and/or focus on a specific application content (e.g. NLP, Bioinformatics):

  o Natural Language Processing (S. M. Mohammad & Hirst 2012),
  o Biomedical studies (Pesquita, Faria, et al. 2009; Jiang et al. 2013)
  o Geo-Spatial data (Schwering 2008),
  o General survey: (Slimani 2013)

- Influence of context on Judgment of semantic similarity
  (Mcdonald & Ramscar 2000)

- Graph-theoretic framework for semantic distance
  (Tsang & Stevenson 2010)

- Semantic relatedness studying object properties in ontologies
  (D'Amato et al. 2005a; D'Amato et al. 2005b; D'Amato et al. 2006; D'Amato et al. 2009)

- Semantic measures and description logics in ontologies
  (D'Amato et al. 2005a; Fanizzi & D'Amato 2006; D'Amato et al. 2009)

- Semantic measures applied to Wikipedia
  (Strube & Ponzetto 2006; Gabrilovich & Markovitch 2009) (Yazdani & Popescu-Belis 2013). Refer to (Yazdani & Popescu-Belis 2013) for more measures using Wikipedia.

- Fuzzy semantic similarity between ontological concepts
  (Cross 2004; Song et al. 2007)

- Semantic measures and Wordnet

(Richardson et al. 1994; Richardson & Smeaton 1995; McHale 1998; Budanitsky 2001; Lewis 2002; Kim & Zhang 2002; Zavaracky 2003; Seco et al. 2004; Pedersen et al. 2004; Yang & Powers 2005; Varelas et al. 2005; Budanitsky & Hirst 2006; Patwardhan & Pedersen 2006; Hoa A. Nguyen 2006; Angryk 2007; Liu et al. 2007; Zhou et al. 2008; Pirró & Seco 2008; Xia et al. 2009; Qin et al. 2009; Agirre et al. 2009; Wang & Hirst 2011; H. Li et al. 2011) (Pirró & Euzenat 2010b)

- Semantic similarity and MeSH
  (Hliaoutakis 2005; Zhu et al. 2009)

- Semantic similarity applied to the Biomedical domain
  (Caviedes & Cimino 2004; Spasić & Ananiadou 2005; Al-Mubaid & Nguyen 2006; Steichen et al. 2006; Hoa A. Nguyen 2006; Pedersen et al. 2007; Yoo et al. 2007; Faria et al. 2007; Zhang et al. 2007; Köhler et al. 2009; McInnes et al. 2009; Batet, Sánchez, Valls, et al. 2010; Batet, Sánchez & Valls 2010; Ferreira & Couto 2010; Saruladha, Aghila & Bhuvaneswary 2010; Grego et al. 2010; Pakhomov et al. 2010; Wang et al. 2010; Sánchez & Batet 2011; Pakhomov et al. 2011; M. H. Schulz et al. 2011; J. Li et al. 2011; Sánchez, Solé-Ribalta, et al. 2012; Paul et al. 2012; Garla & Brandt 2012; Liu & McInnes 2012)
  - see also the section dedicated to the Gene Ontology

- Semantic measures and the Gene Ontology
  (Lord et al. 2003; Couto et al. 2003; Cao et al. 2004; Couto et al. 2005; Sevilla et al. 2005; Yu et al. 2005; Wu et al. 2005; Guo et al. 2006; Othman et al. 2006; Wu et al. 2006; Schlicker et al. 2006; Pesquita et al. 2007; Couto et al. 2007; Pesquita 2007; Wang et al. 2007; Guo 2008; Sheehan et al. 2008; Pesquita et al. 2008; Othman et al. 2008; Mistry & Pavlidis 2008; Nagar & Al-Mubaid 2008; Schlicker & Albrecht 2008; Al-Mubaid & Nagar 2008; Xu et al. 2008; Fontana et al. 2009; Pesquita, Pessoa, et al. 2009; Pesquita, Faria, et al. 2009; Xu et al. 2009; Z. Du et al. 2009; P. Du et al. 2009; Ruths et al. 2009; G. Yu 2010; Tedder et al. 2010; Benabderrahmane, Devignes, et al. 2010; Jain & Bader 2010; Yu et al. 2010; Benabderrahmane, Smail-Tabbone, et al. 2010; Li et al. 2010; Richards et al. 2010; Alvarez & Yan 2011; Park et al. 2011; Mukhopadhyay & De 2011; M. Alvarez et al. 2011; Gaston K. & Nicola J. 2011; Kozielski & Gruca 2011; Mina & Guzzi 2011; Couto & Silva 2011; Cho et al. 2011; Baralis et al. 2011; J. Wang et al. 2012; Yang et al. 2012; Guzzi et al. 2012; Sy et al. 2012; Mazandu & Mulder 2012; Xu et al. 2013)

- Application to image caption retrieval
  (Smeaton & Quigley 1996)

- Applications to GeoInformatics
  (Andrea Rodríguez & Egenhofer 2004; Schwering & Raubal 2005; Akoka et al. 2005; Rodríguez et al. 2005; Janowicz 2006; Keßler et al. 2007; Schwering 2008; Janowicz et al. 2008; Formica & Pourabbas 2008; Su et al. 2010; Janowicz et al. 2011; Ballatore et al. 2012)

- Semantic measures and algorithmic
  (Maguitman & Menczer 2005; Maguitman et al. 2006)

- Construction of semantic network of related concepts
  (Szumlanski & Gomez 2010)

- Ontology-based clustering (and classification)
  (Sanderson & Croft 1999; Maedche & Zacharias 2002; Pekar & Staab 2002; Pantel & Lin
  2002; Nauer & Napoli 2006; Al-Mubaid & Nguyen 2006; Yoo et al. 2007; Ginter et al.
  2007; Zhang et al. 2007; Fanizzi et al. 2008b; Batet et al. 2008; Esposito et al. 2008;
  Esposito et al. 2010; M. Schulz et al. 2011; Batet 2011b)

- Non English Language Processing using semantic measures
  (Wu & Palmer 1994; Osathanunkul et al. 2011)

- General studies of ontology-based semantic measures
  (Blanchard & Harzallah 2005; Blanchard et al. 2008)

- Semantic measures and social networks
  (Waltinger et al. 2009; Capocci et al. 2010)

- Semantic measures using multiple ontologies
  (Rodríguez & Egenhofer 2003; Xiao & Cruz 2005; Petrakis et al. 2006; Kai et al. 2007;
  M.C. Lange, D.G. Lemay 2007; Al-mubaid & Nguyen 2009; Batet, Valls, et al. 2010;
  Saruladha, Aghila & Bhuvaneswary 2010; Coates et al. 2010; Saruladha & Aghila 2011;
  Chiabrando et al. 2011; Saruladha 2011; Sánchez, Solé-Ribalta, et al. 2012; Batet et al.
  2013)

- Unclassified Semantic measures between pairs of classes defined in a knowledge repre-
  sentation.
  (Schickel-Zuber & Faltings 2007; Mazuel & Sabouret 2008; Bin et al. 2009; Cai et al.
  2010; Xiquan et al. 2010; Li & Xia 2011; Spagnola & Lagoze 2011; Ye et al. 2011;
  Sánchez, Batet, et al. 2012)

- Information Retrieval and Semantic measures
  (Lee et al. 1993; Richardson & Smeaton 1995; Hliaoutakis 2005; Hliaoutakis et al. 2006;
  Knappe 2006; Baziz et al. 2007; Saruladha, Aghila & Raj 2010a; Saruladha, Aghila &
  Raj 2010b; Delbru 2011; Sy et al. 2012) (Varelas et al. 2005)

- Semantic measures and Web of Documents
  (Bollegala 2007a; Bollegala et al. 2007; Bollegala 2007b; Bollegala et al. 2009; Sánchez
  et al. 2009; Al-mubaid & Nguyen 2009; Batet, Sánchez & Valls 2010; Bollegala et al.
  2010; Iosif & Potamianos 2010; Jian 2011)

- Semantic Measures and Semantic Web
  (Delbru 2011)
  - See section related to ontology mapping/matching instance matching and Linked data

- Semantic Measures and Ontology matching and mapping
  (Euzenat & Shvaiko 2007; Ge & Qiu 2008; Wang & Jou 2011)

- Semantic Measures and Instance Matching
  (Castano et al. 2011)

- Semantic Measures and reasoning
  (Huang & Harmelen 2008)

- Semantic Measures and ontology development
  (Ramezani 2011)

- Semantic Measures and Linked Data
  (Sheng et al. 2010; Passant 2010; Olsson et al. 2011; Baumann & Schirru 2012)

- Semantic Measures between RDF instances (or entities characterized in a graph)
   (Bisson 1995; Oldakowski & Bizer 2005; Albertoni & De Martino 2006; Volz et al. 2009; Araujo et al. 2011; Harispe, Ranwez, et al. 2013a; Andrejko & Bieliková 2013) (Pirró 2012)

- Semantic relatedness and conceptual network
   (Gurevych 2005)

- Semantic measures between texts
   (Mihalcea et al. 2005)

- Semantic measures for question answering
   (C. Wang et al. 2012)

- Semantic measures and machine translation
   (Wu & Palmer 1994)

- Semantic measures and paraphrase detection
   (Iordanskaja et al. 1991; Fernando & Stevenson 2008)

- Information Content
  (Resnik 1995; Seddiqui 2010; Sánchez et al. 2011; Sánchez & Batet 2012)

- Semantic measures and Cognitive Sciences
  (Tversky 1977; Miller & Charles 1991; Landauer & Dumais 1997; Schwering 2005; Martin 2007; Binder & Desai 2011; Aimé 2011; Decock & Douven 2011)

- Semantic measures and Disambiguation
  (Resnik 1999) (Sussna 1993)

- Unification of measures
  (Cross et al. 2013) (Mazandu & Mulder 2013) (Blanchard & Harzallah 2005; Blanchard et al. 2008; Blanchard 2008) (Cross 2006; Pirró & Euzenat 2010a; Cross & Yu 2010;

Sánchez & Batet 2011; Cross et al. 2013) (Harispe, Sánchez, et al. 2013) (Roddick et al. 2003)

- Graph-based measures (can be applied to deal with semantic graphs)
  (Entringer et al. 1976; Chartrand et al. 1998; Fogaras & Rácz 2005; Fouss et al. 2007; Goddard & Oellermann 2011)

- Software solutions and tools for semantic measures
  (Pedersen et al. 2004; Oldakowski & Bizer 2005; Bernstein et al. 2005; Janowicz et al. 2007; McInnes et al. 2009; Rus et al. 2013) (Pesquita, Pessoa, et al. 2009)

- Other contributions (unclassified)
  (Tsang 2008) (Banerjee & Pedersen 2003) (Haase et al. 2004) (Budanitsky 1999) (S. M. Mohammad & Hirst 2012) (Salahli 2009) (Jarmasz & Szpakowicz 2002) (Jiang & Conrath 1997) (Raftopoulou & Petrakis 2005) (Rada et al. 1989) (Hall 2006) (Wu et al. 2009) (Zhao et al. 2009) (Lee et al. 2008) (Shi & Setchi 2010) (Tudhope & Taylor 1997) (Hawalah & Fasli 2011) (Pirró 2009) (Cross 2006) (Haralambous & Klyuev 2011) (Danowski 2010) (Patwardhan 2003) (Hughes & Ramage 2007) (Roddick et al. 2003) (Ramage et al. 2009) (Fanizzi et al. 2008a) (Stuckenschmidt 2009a) (Pirró & Euzenat 2010a) (Mario Jarmasz 2003) (Lemaire & Denhière 2008) (Li et al. 2003) (Resnik 1995) (Tsang & Stevenson 2006) (Iosif & Potamianos 2010) (Stevenson M 2005) (Cheng et al. 2009) (Ranwez et al. 2006) (Cho et al. 2003) (Anna 2008) (Van Buggenhout & Ceusters 2005) (Tadrat et al. 2011) (Agirre et al. 2010) (Liu 2011) (Halawi et al. 2012) (Cordi et al. 2005) (Ramage et al. 2009) (Maedche & Staab 2001) (Alqadah & Bhatnagar 2011)

- Related work
  (Orozco & Belanche n.d.) (Bisson 1995) (Kovács 2010)

## 2. Semantic measures for Text Segments Comparison

Most of the work related to text segment comparison has focus on vector model as well as n-gram language and topic models. These models follow a direct approach. A second approach, defined as indirect, regroup the models which assess the semantic similarity/relatedness of texts according to the similarity of their words. Both approaches are briefly detailed in this appendix.

### Direct Approaches

*Lexical overlap*

The similarity of two texts can be assessed based on the number of words or n-gram the two texts have in common. Note that this type of measures is not based on the semantic analysis of texts. A large diversity of combination of parameters can be used (Rus et al. 2013):

- Pre-processing: Collocation detection, punctuation stop, word removal.
- Filtering option: all words, content of words.

- Weighting schemes.

Measures based on extensions of the LSA word-to-word measures have also been proposed for text-to-text comparison (Lintean et al. 2010). They characterize a text through a LSA vector build summing individual words' vectors (vectors are therefore compared using classical techniques, e.g., cosine similarity).

Other approaches based on Latent Dirichlet Allocation (Blei et al. 2003) can be used to compare texts adopting a topic model approach. Texts are considered as random mixture over latent topics and each topic is associated to a distribution over words, i.e., has a probability to generate specific words (the probabilities are defined through supervised labelling or co-occurrence analysis). The distributions over topics associated to each document and the similarity of topics (the distribution over words) can next be used to compare two texts.

### Indirect Approaches – Word-to-Word Aggregation

Indirect approaches model the semantic similarity/relatedness of texts according to the similarity and the specificity of the words composing them.

In (Mihalcea et al. 2006), the authors proposed to sum the maximal similarity between the Cartesian product of the two sets of words composing the two texts. The specificity of words is also taken into account using the Inverse Document Frequency (IDF) (Jones 1972). Any word-to-word similarity measures can be used (a word disambiguation step may be needed). In (Rus et al. 2013), the authors proposed the optimal lexical matching strategy, an approach derived from the assignment problem[i]. In (Fernando & Stevenson 2008) all similarities of the pairs of words contained in the Cartesian product were considered. Measures based on the aggregation or extensions of the LSA word-to-word measures have also been proposed for text-to-text comparison (Lintean et al. 2010).

## 3. Corpora of Texts for Distributional Approaches

List of some resources commonly used to compute the similarity of words from corpora analysis:

- British National Corpus (BNC)
- Associated Press news history
- Wall street journal
- Tipster corpus
- Brown corpus

---

[i] Famous combinatorial optimization problem.

# 4. How to Map a Knowledge Representation to a Semantic Graph

### RDF(S) Graphs and Semantic Measures

This note discusses some technical aspects relative to the computation of SMs on RDF(S) graphs. It mainly presents the considerations to be taken into account to map an RDF(S) graph to the kind of semantic graph generally expected by SMs.

The simple graph data model, considered for algorithmic studies and definitions of SMs, differs from the RDF graph specification in multiple ways, e.g. no blank nodes or literals in some cases. These differences do not prevent the use of SMs on RDF graphs. Nevertheless, to ensure both coherency and reproducibility of results, guidelines regarding the use of these measures on those graphs have to be rigorously defined. Handling of RDF graphs which take advantage of precise formal vocabularies such as RDFS must also be clearly defined. To our knowledge, required pre-processing enabling the use of SMs on RDF(S) graphs has not been previously discussed in literature. We will refer to RDFS entailment rules, please consider the W3C specification for rule numbering[i].

On RDF(S) graphs, instances and classes are not clearly separated as "*A class may be a member of its own class extension and may be an instance of itself*"[ii] (e.g. using punning techniques, meta-class). Such cases are not limiting to our work as in general practice, instances and classes can easily be distinguished using basic rules and restrictions. Moreover, the separation of instance data and the taxonomy of classes is considered as a fundamental aspect of knowledge representation modelling which is therefore usually respected. The definition of a function enabling two distinguish the type to associate to a node, i.e., Class (C), Instance (I), Predicate (P) or Data value (D) is therefore not considered to be limiting[iii].

SMs algorithms heavily rely on graph traversals. In order for the measures scores to be accurate and reproducible, the graph must first be entailed according to RDFS entailment rules. The graph needs to be reduced for some properties required by the measures to be respected. Both graph entailment and reduction required prior to SMs computation are detailed.

Since most treatments associated to SMs are expressed in terms of graph traversals, a RDFS reasoner must be used to infer all implicit relationships based on RDFS entailment rules 3, e.g. `rdf:type` inference according to the domain/range associated to a property (predicate). To reduce the complexity of the entailment, only the RDFS entailment according to rules 2, 3 and 5, 7 must be applied. Rules 2 and 3 are respectively related to `rdfs:domain` and `rdfs:range` type inference. Rules 5 and 7 are related to sub-property relationships and are therefore important in order to infer new statements according to rules 2 and 3. Other entailment rules have no direct implications on the topology of the resulting graph as inferred relationships will not be considered by SMs algorithms.

---

[i] RDF Semantics, http://www.w3.org/TR/rdf-mt
[ii] RDF Schema, http://www.w3.org/TR/rdf-schema
[iii] Note that in OWL-DL the sets of classes, predicates and instances (concepts, roles and individuals) must be disjoint (Horrocks & Patel-Schneider 2003).

The graph must respect certain properties. Not all inferable relationships, according to the transitivity of taxonomic relationship or chain of transitive relationships (i.e. transitivity over `rdfs:subClassOf`), are to be considered. This treatment can be carried out through an efficient transitive reduction algorithm (see section 5.2.2.5). Furthermore, some classes associated to RDFS Vocabulary and/or other classes not explicitly defined in the ontology must be ignored prior to the treatment, e.g. `rdfs:Property`, `rdfs:Class`. Triplets associated to these excluded classes and RDFS axiomatic triples[i] must also be ignored. Such pre-processing is important to ensure coherency of both SMs and particular metrics (e.g., information content). As an example, an RDFS reasoner will infer that all classes are subclasses of `rdfs:Resource` and create the corresponding triplets. Thus, considering the edge-counting measure, the maximal distance between two concepts will be set to 2, which is not the expected behaviour for most usage contexts.

In RDF graphs, blank nodes or reification techniques can be used to model specific information into the graph. We consider that any blank node is associated to a class, a predicate, an instance of a class or a specific relationship. As an example, consider the set of RDF statements:

```
_r1    rdf:subject      ex:luc
_r1    rdf:predicate    foaf:knows
_r1    rdf:object       ex:louise
_r1    ex:Degree        ex:High
```

This set of statements can be graphically represented by the graph (A) in Figure 17.

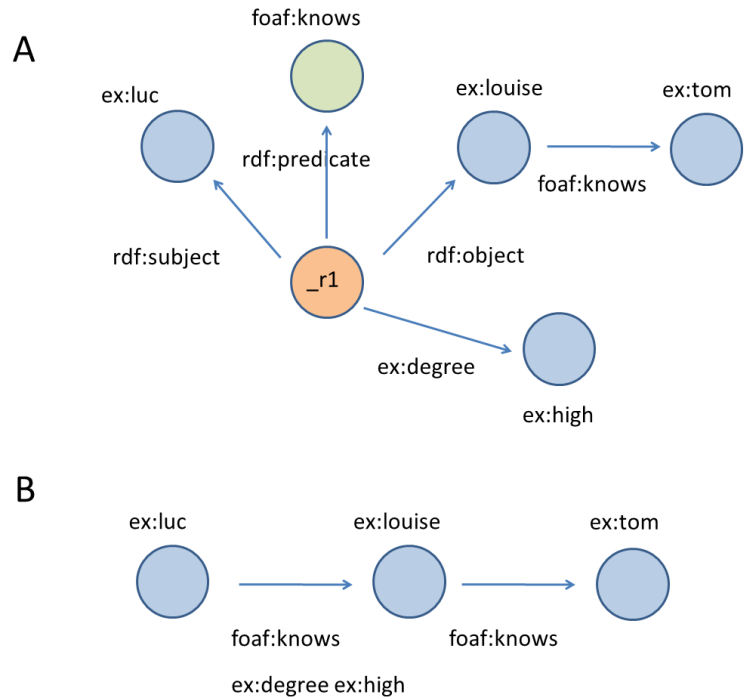

Figure 17: Example of mapping between (A) an RDF graph modelling a specific knowledge using particular design pattern, and (B) a semantic graph representation of this knowledge as it is expected by most SMs.

In (A), the node `_r1` corresponds to a blank node used to express properties on a specific relationship (reification). The representation expected by SMs is generally based on the classical graph property model.

Nevertheless, most SMs expect such a knowledge to be expressed as the graph presented in Figure 17 (B). As an example, the path between `ex:luc` and `ex:tom` is expected to be:

$$[ex:luc, \ ex:louise, \ ex:tom]^{foaf:knows}$$

However, in the graph, the shortest path between the two instances is:

```
[
(ex:luc,rdf:subject˜,_r1),
(_r1,rdf:object,ex:louise),
(ex:louise, foaf:knows,ex:tom)
]
```

### From OWL to Semantic Graphs (TODO)

We describe some modifications which can be performed to map a KR expressed in OWL into a semantic graph which can be processed by most SMs framed in the relational setting.

➜ Reference (Horrocks & Patel-Schneider 2003)

## 5. From Distance To Similarity and *vice-versa*

A similarity $s$ (bounded by 1) can be transformed to a distance $d$ considering multiple approaches (Deza & Deza 2013). A distance can also be converted to a similarity. Some of the approaches used for the transformations are presented above.

### Similarity to distance

If *sim* is normalized:

$$dist = 1 - sim$$

$$dist = \frac{1 - sim}{sim}$$

$$dist = -\ln sim$$

**Distance to Similarity**

$$sim = \frac{1}{dist + 1}$$